\documentclass[conference]{IEEEtran}
\IEEEoverridecommandlockouts
\def\BibTeX{{\rm B\kern-.05em{\sc i\kern-.025em b}\kern-.08em
		T\kern-.1667em\lower.7ex\hbox{E}\kern-.125emX}}
\usepackage[font=small,labelfont=bf,tableposition=top]{caption}

\usepackage{blindtext}

\usepackage{times}

\usepackage[numbers]{natbib}
\usepackage{multicol}
\usepackage[bookmarks=true]{hyperref}

\usepackage{amsmath,amsfonts,bm}

\def\eqref#1{equation~\ref{#1}}

\def\1{\bm{1}}

\def\ra{{\textnormal{a}}}

\def\rs{{\textnormal{s}}}

\def\va{{\bm{a}}}

\def\vh{{\bm{h}}}

\def\vn{{\bm{n}}}

\def\vp{{\bm{p}}}
\def\vq{{\bm{q}}}

\def\vt{{\bm{t}}}

\def\vv{{\bm{v}}}

\def\vx{{\bm{x}}}
\def\vy{{\bm{y}}}
\def\vz{{\bm{z}}}

\def\mK{{\bm{K}}}

\def\mR{{\bm{R}}}

\def\mT{{\bm{T}}}

\DeclareMathAlphabet{\mathsfit}{\encodingdefault}{\sfdefault}{m}{sl}
\SetMathAlphabet{\mathsfit}{bold}{\encodingdefault}{\sfdefault}{bx}{n}

\def\sH{{\mathbb{H}}}

\newcommand{\R}{\mathbb{R}}

\DeclareMathOperator*{\argmin}{arg\,min}

\usepackage{multirow}
\usepackage{hyperref}
\usepackage{url}
\usepackage{graphicx,subfigure} 
\usepackage{wrapfig}

\usepackage{floatrow}
\newfloatcommand{capbtabbox}{table}[][\FBwidth]

\usepackage[dvipsnames]{xcolor}
\definecolor{blau}{RGB}{0, 69, 134}
\definecolor{orangeR}{RGB}{255,140,0}
\definecolor{gruen}{RGB}{0,150,0}
\definecolor{lila}{rgb}{0.49400,0.18400,0.55600}
\definecolor{rot}{rgb}{1,0.12500,0.009800}
\definecolor{hellblau}{RGB}{23, 190, 207}
\definecolor{rosa}{RGB}{207, 23, 190}

\usepackage{pgfplots}
\usetikzlibrary{external}

\pgfplotsset{compat=newest}
\usetikzlibrary{plotmarks}
\usepackage{grffile}

\usetikzlibrary{shapes,arrows,fit,calc,positioning,matrix}
\tikzstyle{box} = [draw, rectangle, thick, node distance=1.7cm, minimum height=1cm, align=center]
\tikzstyle{container} = [draw, rounded corners, rectangle, dashed, inner sep=0.3cm ,transform shape=false]
\tikzstyle{line} = [draw, very thick, -latex']
\tikzstyle{sum} = [draw,circle,thick,inner sep=0mm,minimum size=3mm]
\tikzstyle{branch} = [circle,inner sep=0pt,minimum size=2mm,fill=black,draw=black]

\usetikzlibrary{decorations.markings}

\pgfplotsset{every axis/.append style={legend style={font=\small}}}
\pgfplotsset{every axis label/.append style={font=\small}, tick label style={font=\small}} 
\pgfplotsset{every axis/.append style={line join=round}}
\usepgfplotslibrary{statistics}

\graphicspath{{figure/}}

\let\oldtwocolumn\twocolumn
\renewcommand\twocolumn[1][]{%
	\oldtwocolumn[{#1}{
		\begin{center}
			\centering
			\includegraphics[width=\textwidth, viewport=0 0 780 220, clip]{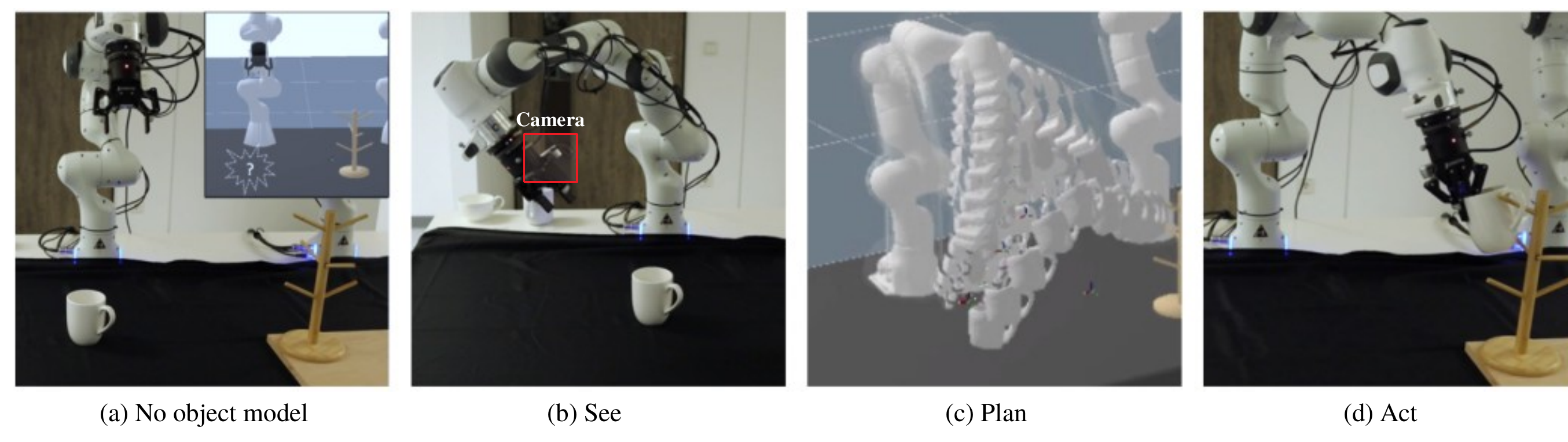}
			\captionof{figure}{Unlike static objects and a robot's own body, 3D models of objects to manipulate are often unavailable. The proposed Deep Visual Constraints represent an object, directly from images, as a continuous function over the 3D space and predict the task constraint values based on such representation. This so-called implicit representation can naturally describe object's rigid transformations in $SE(3)$, enabling efficient optimization-based manipulation planning. The overall pipeline as well as demonstrations are well-visualized in \url{https://youtu.be/r__mIGTu6Jg}}
			\label{fig:realRobot}
		\end{center}
	}]
}

\begin{document}

\title{Deep Visual Constraints: Neural Implicit Models for Manipulation Planning from Visual Input}

\author{Jung-Su Ha \par~~~ Danny Driess \par~~~ Marc Toussaint \\
		Learning \& Intelligent Systems Lab, TU Berlin, Germany}

\maketitle

\begin{abstract}
	Manipulation planning is the problem of finding a sequence of robot configurations that involves interactions with objects in the scene, e.g., grasping and placing an object, or more general tool-use. To achieve such interactions, traditional approaches require hand-engineering of object representations and interaction constraints, which easily becomes tedious when complex objects/interactions are considered. Inspired by recent advances in 3D modeling, e.g.\ NeRF, we propose a method to represent objects as continuous functions upon which constraint features are defined and jointly trained. In particular, the proposed pixel-aligned representation is directly inferred from images with known camera geometry and naturally acts as a perception component in the whole manipulation pipeline, thereby enabling long-horizon planning \textit{only from visual input}.
	\textbf{Project page:} \url{https://sites.google.com/view/deep-visual-constraints}
\end{abstract}

\IEEEpeerreviewmaketitle
\section{Introduction}
Dexterous robots should be able to flexibly interact with objects in the environment, such as grasping and placing an object, or more general tool-use, to achieve a certain goal. 
Such instances are formalized as manipulation planning, a type of motion planning problem that solves not only for the robot's own movement but also for the objects' motions \textit{subject to} their interaction constraints.
Therefore, designing interaction constraint functions, which we also call interaction features, is at the core of achieving the robot dexterity.
Traditional approaches rely on hand-crafted constraint functions based on geometric object representations such as meshes or combinations of shape primitives.
However, when considering large varieties of objects and interaction modes, such traditional approaches have long-standing limitations in two aspects:
(i) The representations have to be inferred from raw sensory inputs like images or point clouds -- raising the fundamental problem of perception and shape estimation. 
(ii) With increasing generality of object shapes and interaction, representation's complexity grows, thereby making hand-engineering of the interaction features inefficient.
However, if the aim is manipulation skills, the hard problem of precise shape estimation and the feature engineering might be unnecessary.

\textit{What is a good object representation}?
Considering the representation will be used to predict interaction features, we expect it to encode primarily task-specific information rather than only geometric. We also expect some of the information to be shared across different interaction modes.
In other words, good representations should be task-specific so that the feature prediction can be simplified and, at the same time, be task-agnostic to enable synergies between the tasks. E.g., mug handles are called handles because we can handle the mug through them and also, once we learn the notion of a handle, we can play around with the mug through the handle in many different ways.
Also, from the perception standpoint, good representations should be easy to infer from raw sensory inputs and should be able to trade their accuracy (if bounded) in favor of the feature prediction.

To this end, we propose a data-driven approach to learning interaction features that are conditioned on object images. 
The whole pipeline is trained end-to-end directly with the task supervisions so as to make the representation and perception \textit{task-specific} and thus to simplify the interaction prediction.
The object representation acts as a bottleneck and is shared across multiple features so that the \textit{task-agnostic} aspects can emerge.
We propose the representation to be a $d$-dimensional continuous function over the 3D space~\cite{park2019deepsdf,mildenhall2020nerf}. 
In particular, the proposed implicit neural representation is pixel-aligned, meaning that the function takes as input images from multiple cameras (e.g.\ stereo) and, assuming known camera poses and intrinsics, computes a representation at a certain 3D location using image features at the corresponding 2D pixel coordinates. Once learned, the interaction features can be used by a typical constrained optimal control framework to plan dexterous object-robot interaction.
We show that making use of the learned constraint models within Logic-Geometric Programming (LGP)~\cite{toussaint2018differentiable} enables planning various types of interactions with complex-shaped objects \textit{only} from images.
Since the representations generalize well, the learned constraint models are directly applicable to manipulation tasks involving unseen objects.
To summarize, our main contributions are
\begin{itemize}
	\item To represent objects as neural implicit functions upon which interaction constraint functions are trained,
	\item An image-based manipulation planning framework with the learned features as constraints,
	\item Comparison to non pixel-aligned, non implicit function, and geometric representations,
	\item Demonstration in various manipulation scenarios ranging from basic pick-and-hang~\href{https://youtube.com/playlist?list=PL9pnj8nG83OfROuTSSxCEego78gzeUsA3}{[video1]} to long-horizon manipulation \href{https://youtube.com/playlist?list=PL9pnj8nG83OfNsYhRQ9gzhceT1PT8IXkq}{[video2]}, zero-shot imitation \href{https://youtube.com/playlist?list=PL9pnj8nG83Oe3dMI_dX-7Xd69XXNWONIy}{[video3]}, and sim-to-real transfer \href{https://www.youtube.com/playlist?list=PL9pnj8nG83OdYYLQsYRxSKycasWW4Eo4n}{[video4]}.
\end{itemize}

\section{Related Work}
\subsection{Neural Implicit Representations in 3D Modeling}
Implicit neural representations have recently gained increasing attention in 3D modeling.
The core idea is to encode an object or a scene in the weights of a neural network, where the network acts as a direct mapping from 3D spatial location to an implicit representation of the model, such as occupancy measures~\cite{mescheder2019occupancy}, signed distance fields (SDF)~\cite{park2019deepsdf,atzmon2020sal}, or radiance fields~\cite{mildenhall2020nerf}.
In contrast to explicit representations like voxels, meshes or point clouds, the implicit representations don't require discretization of the 3D space nor fixed shape topology but rather continuously represent the 3D geometry, thereby allowing for capturing complex shape geometry at high resolutions in a memory efficient way.

There have been attempts to associate these 3D representations with 2D images using the principle of camera geometry.
Exploiting the camera geometry in a forward direction, i.e., 2D projection of 3D representations, yields a differentiable image rendering procedure and this idea can be used to get rid of 3D supervisions.
For example, \citet{sitzmann2019scene,niemeyer2020differentiable,yariv2020multiview,mildenhall2020nerf,henzler2021unsupervised,reizenstein2021common} showed that the representation networks can be trained without the 3D supervision by defining a loss function to be difference between the rendered images and the ground-truth.
Another notable application of this idea is view synthesis. 
Based on the differentiable rendering, \citet{park2020latentfusion,chen2020category,yen2020inerf} addressed unseen object pose estimation problems, where the goal is to find object's pose relative to the camera that produces a rendered image closest to the ground truth.
By conditioning 3D representations on 2D input images, one can expect the amortized encoder network to directly generalize to novel 3D geometries without requiring any test-time optimization.
This can be done by introducing a bottleneck of a finite-dimensional \textit{global} latent vector between the images and representations, but these global features often fail to capture fine-grained details of the 3D models~\citep{peng2020convolutional}.
To address this, the camera geometry can be exploited in the inverse direction to obtain pixel-aligned \textit{local} representations, i.e., 3D reprojection of 2D image features.
\citet{saito2019pifu} and \citet{xu2019disn} showed that the pixel-aligned methods can establish rich latent features because they can easily preserve high-frequency components in the input images.
Also, \citet{yu2020pixelnerf} and \citet{trevithick2021grf} incorporated this idea within the view-synthesis framework and showed that their convolutional encoders have strong generalizations. 

While the above work investigates implicit neural representation to model shapes or appearances, our work makes use of it to model physical interaction feasibility and thereby to provide a differentiable constraint model for robot manipulation planning.

\subsection{Object/Scene Representations for Robotic Manipulations}
Several works have proposed data-driven approaches to learning object representations and/or interaction features which are conditioned on raw sensory inputs, especially for grasping of diverse objects.
One popular approach is to train discriminative models for grasp assessments.
For example, \citet{ten2017grasp,mahler2017dex,van2020learning} trained a neural network that, for given candidate grasp poses, predicts their grasp qualities from point clouds.
In addition, \citet{breyer2020volumetric,jiang2021synergies} proposed 3D convolutional networks that take as inputs a truncated SDF and candidate grasp poses and return the grasp affordances.
Similarly, \citet{zeng2020tossingbot,zeng2020transporter} addressed more general manipulation scenarios such as throwing or pick-and-place, where a convolutional network outputs a task score image.
On the other hand, neural networks also have been used as generative models.
For example, \citet{mousavian20196} and \citet{murali20206} adopted the approach of conditional variational autoencoders to model the feasible grasp pose distribution conditioned on the point cloud.
\citet{sundermeyer2021contact} proposed a somewhat hybrid method, where the network densely generates grasp candidates by assigning grasp scores and orientations to the point cloud.
\citet{you2021omnihang} addressed the object hanging tasks from point clouds where the framework first makes dense predictions of the candidate poses among which one is picked and refined. 
Compared to these works, our framework takes advantage of a trajectory optimization to jointly optimize an interaction pose sequence instead of relying on exhaustive search or heuristic sampling schemes, thus not suffering from the high dimensionality nor the combinatorial complexity of long-horizon planning problems.

Another important line of research is learning and utilizing keypoint object representations.
\citet{manuelli2019kpam,gao2021kpam,qin2020keto,turpin2021gift} represented objects using a set of 3D semantic keypoints and formulated manipulation problems in terms of such the keypoints.
Similarly, \citet{manuelli2020keypoints} learned the object dynamics as a function of keypoints upon which a model predictive controller is implemented.
Despite their strong generalizations to unseen objects, the keypoint representations require semantics of the keypoints to be predefined.

The representation part of our framework is closely related to dense object descriptions proposed by ~\citet{florencemanuelli2018dense,florence2019self}.
The idea is to train fully-convolutional neural networks that maps a raw input image to pixelwise object representations which directly generalize to unseen objects.
As a recent concurrent work from \citet{simeonovdu2021ndf} also proposed, our implicit representation extends this pixel-wise representation to the 3D space. Compared with those existing work, ours is learned via task supervisions in conjunction with the task feature heads and thus can be seamlessly integrated into general sequential manipulation planning problems.
Another recent related work from \citet{yuan2021sornet} proposed learning object-centric representations that are used to predict the symbolic predicates of the scene which in turn enables symbolic-level task planning.
\citet{21-driess-CORL} formulated manipulation planning problems solely in terms of SDFs as representations and proposed to learn manipulation constraints as functionals of SDFs.
More recently, \citet{2022-driess-compNerfPreprint} trained implicit object encoders together with differentiable image rendering decoders and used a graph neural network to model dynamics, based on which an RRT-based method can plan sequential manipulations in the latent space.
In contrast to the above, our proposed representations are trained in conjunction with multiple task prediction heads and can be seamlessly integrated into sequential manipulation planning schemes that generate motions flexibly blending diverse interactions together.

\section{Deep Visual Constraints (DVC) via Implicit Object Representation}\label{sec:featurePred}
\begin{figure}[h]
	\centering
	\scalebox{0.65}{
		\begin{tikzpicture}
		\tikzset{block/.style={rectangle, draw,fill=white}}

		\pgfdeclarelayer{background}
		\pgfdeclarelayer{foreground}
		\pgfsetlayers{background,main,foreground}

		\node[block, draw, align=center, rounded corners] at (0,0) (pose) {robot or static\\frame's pose\\$\vq\in SE(3)$};
		\node[block, draw, above=.3cm of pose, align=center, rounded corners] (imgs) {obj images,\\cam poses,\\intrinsics: $\mathcal{V}$};
		
		\node[block, draw, right= .7 of pose, align=center, rounded corners] (p) {interaction points\\$\{\vp_1,...,\vp_{K}\}$};
		
		\node[right= 0.5 of p] (pk) {$\vp_k$};

		\node[block, draw, above right= -0.26 and 0.5 of pk, align=center] (backbone) {$\psi(\vp; \mathcal{V})$\\backbone\\(PIFO)};
		
		\node[right= 0.3 of backbone] (yk) {$\vy_k$};

		\node[right= 0.4 of yk] (concat) {\Large$\bigoplus$};	
		\node[below= 0 of concat] (concat2) {Concat.};	
		\node[block, draw, right=0.4 of concat, align=center] (FH) {task\\head};
		\node[right= 0.6 of FH] (h) {$\vh$};

		\draw[->, very thick] (pose) -- (p);

		\draw[->, very thick] (imgs.east) -| ++(1.5,0.0) -| +(0.7,-0.85) -- (backbone.170);
		\draw[->, very thick] (p) -- (pk);
		\draw[->, very thick] (pk)  -| ++(.3,0.0)-| ++(.3,0.5) -- (backbone.193);
		\draw[->, very thick] (backbone.east) -- (yk);
		\draw[->, very thick] (yk) -- (concat);
		\draw[->, very thick] (yk) -- (FH);
		\draw[->, very thick] (FH) -- (h);

		\node[above right = 0 and -0.8 of backbone, align=center] (s) {shared\\(Sec.\ref{sec:PIFO})};
		\node[above right = .3 and -1.7 of FH, align=center] (t) {task-specific\\(Sec.\ref{sec:head})};

		\node[below left = 0 and .1 of p, font=\LARGE] (tem1) {};
		\begin{pgfonlayer}{background}
		\node [rounded corners, draw, fill=blau!10, fit={(t) (tem1)}] {};
		\end{pgfonlayer}

		\node[below left = 0 and 0 of pk, font=\LARGE] (tem2) {};
		\begin{pgfonlayer}{background}
		\node [rounded corners, draw, fill=blau!30, fit={(tem2) (s)}] {};
		\end{pgfonlayer}

		\end{tikzpicture}
	}
	\caption{Feature prediction of DVC. The backbone, conditioned on a set of posed images, computes representation vectors at queried 3D spatial locations $\vp_k\in\R^3$, and the task head predicts the interaction feature based on the obtained representation vectors $\vy_k\in\R^d$.}\label{fig:overall}
\end{figure}
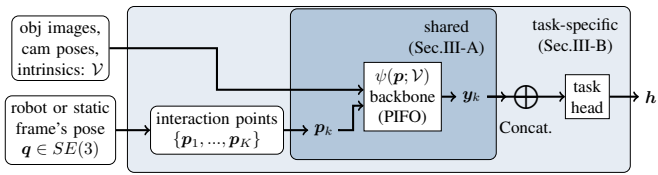
Given $N_\text{view}$ images with their camera poses/intrinsics, $\mathcal{V} = \{(\mathcal{I}^1, \mT^1, \mK^1), ..., (\mathcal{I}^{N_\text{view}}, \mT^{N_\text{view}}, \mK^{N_\text{view}})\}$ with $\mathcal{I}\in\R^{3\times H\times W}$ (we considered $H=W=128$) and $\mT,\mK\in\R^{4\times 4}$, we build an interaction feature as a neural network:
\begin{align}
h = \phi_\text{task}(\vq; \mathcal{V}), \label{eq:feature}
\end{align}
where $\vq\in SE(3)$ is the pose of the robot/static frame interacting with the object; the interaction feature $h\in\R$, analogous to energy potentials, is zero when feasible and non-zero otherwise, which will act as an equality constraint in manipulation planning.
As shown in Fig.~\ref{fig:overall}, the feature prediction framework consists of two parts: the representation backbone which serves as an implicit representation of an object, and the task heads that make feature predictions.
Notably, while the multiple task heads individually model different interaction constraints, the backbone is shared across them, allowing for learning more general object representation. 

\begin{figure*}[t]
	\centering
	\includegraphics[width=.99\columnwidth]{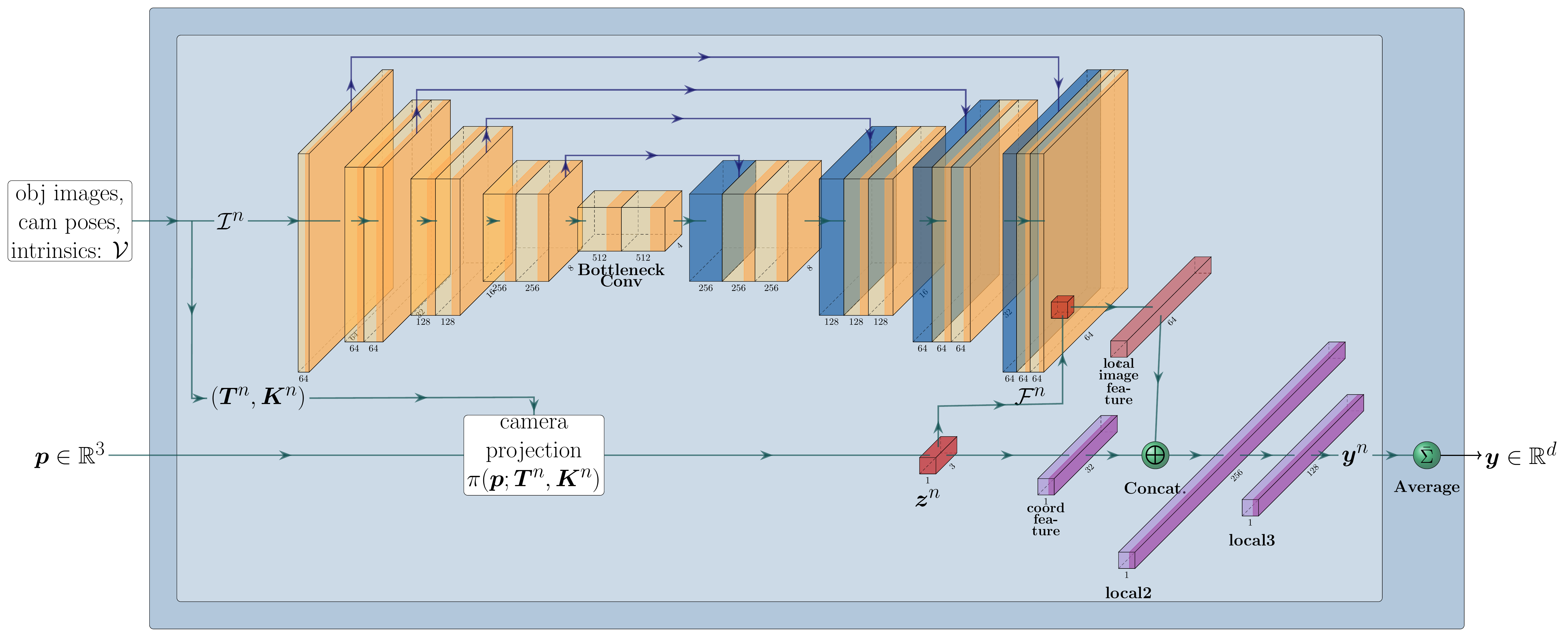}
	\caption{PIFO (i) encodes the images $\mathcal{I}$ as pixel-wise feature maps $\mathcal{F}$ via U-net, (ii) projects the query point $\vp$ into the pixel coordinate $\vz$ using camera geometry $(\mT, \mK)$, and (iii) computes the representation vector $\vy^d$ by extracting the image features at the projected points.}\label{fig:backbone}
\end{figure*}
\subsection{Pixel-Aligned Implicit Functional Object (PIFO)} \label{sec:PIFO}
The proposed implicit object representation is a mapping:
\begin{align}
\vy = \psi(\vp; \mathcal{V}),
\end{align}
where $\vp\in\R^3$ and $\vy\in\R^d$ are a queried 3D position and a representation vector at that point, respectively.
This function, implemented as a neural network as depicted in Fig.~\ref{fig:backbone}, consists of three parts: image encoder, 3D reprojector, and feature aggregator. The first two compute a representation vector from each image and the last one combines them. 

{\bf Image Encoder:} This module takes as input an image and computes a feature map (the pathway from $\mathcal{I}^n$ to $\mathcal{F}^n$ in Fig.~\ref{fig:backbone}). We adopted the hourglass network architecture, especially with ResNet-34 as its downward path and two residual layers with $3\times3$ convolutions followed by up-convolution as the upward path:
\begin{align}
\mathcal{F}^n = UNet(\mathcal{I}^n),~\forall n\in\{1,...,N_\text{view}\},
\end{align}
which results in a feature map $\mathcal{F}^n\in\R^{64\times64\times64}$ that captures both local and global information in the input image.

{\bf 3D Reprojector:} To endow the network with the multi-view consistency, all the 3D operations are performed in the view space. 
The 3D reprojector, the pathway from $(\mT^n, \mK^n)$, $\mathcal{F}^n$ and $\vp$ to $\vy^n$ in Fig.~\ref{fig:backbone}, transforms a queried point, $\vp$, into the image coordinate including depth, $\pi(\vp; \mT, \mK)=\vz\in\R^3$ and extracts the local image feature at the projected point from the feature map, $\mathcal{F}$, via bilinear interpolation.
Finally, the extracted feature and the coordinate feature, which is computed through a couple of fully connected layers (FCLs), are passed to a couple of FCLs to get a representation vector at $\vp$ for a single image, i.e., $\forall n\in\{1,...,N_\text{view}\},$
\begin{align}
\vy^n = MLP(\mathcal{F}^n(\vz^n), \vz^n),~\vz^n = \pi(\vp; \mT^n, \mK^n).
\end{align}

{\bf Feature Aggregator:} This module is the pathway from $\vy^n$ to $\vy$ in Fig.~\ref{fig:backbone}, which aggregates the representation vectors from multiple views into one vector. Among many permutation-invariant options, like summation or more sophisticated attention mechanisms, we simply take the averaging operation for it, i.e.,\
\begin{align}
\vy = \frac{1}{N_\text{view}}\sum_{n=1}^{N_\text{view}}\vy^n.
\end{align}

\subsection{Interaction Task Feature Prediction} \label{sec:head}
A task head evaluates the interaction constraint violation, $h$, for a given robot/static frame's pose, $\vq$, using the object representation \textit{function over 3D}, $\psi(\cdot)$.
To this end, we rigidly attach a set of keypoints to the robot frame at which the backbone is queried, i.e., $\forall k\in\{1,...,K\}$, 
\begin{align}
\vy_k = \psi(\vp_k;\mathcal{V}),~\vp_k = \mR(\vq)\hat\vp_k+\vt(\vq),
\end{align}
where $\hat\vp_k$ is $k^\text{th}$ keypoint's local coordinate, and $\mR(\vq)$ and $\vt(\vq)$ denote the rotation matrix and the translation vector of $\vq$, respectively.
Finally, the task head, based on the resulting representation vectors, predicts a constraint value through a couple of FCLs:
\begin{align}
h = MLP(\vy_1,...,\vy_{K}).
\end{align}

\section{Training} \label{sec:training}
In this paper, we consider manipulation scenarios where a robot arm, Franka Emika Panda, or two manipulate mugs.
The shapes of mugs are diverse and the scene contains multiple hooks on which a mug can be hung.
Formulating such problems requires three types of learned interaction features: an SDF feature for collision avoidance and grasping/hanging features, so we prepared the dataset for each.

\subsection{Data Generation}
We took 131 mesh models of mugs from ShapeNet~\cite{shapenet2015} and convex-decomposed those meshes. The meshes are translated and randomly scaled so that they can fit in a bounding sphere with a radius of $10\sim15\text{ cm}$ at the origin. 
For each mug, we created the following dataset.

{\bf Posed Images:} The posed image data consists of 100 images ($128\times128$) with the corresponding camera poses and intrinsic matrices generated by the OpenGL rendering. Azimuths and elevations of the cameras are sampled such that they are uniformly projected onto the unit sphere, while their distances from the object center are random. The azimuth, elevation and distance fully determine the camera's positions, and the camera's orientations are set such that the cameras are upright and face the object center. For the intrinsics, we used the field of view $fov = 2\arcsin(d/r)$, where $d$ is the camera distance from the object center and $r$ is the radius of the object's bounding sphere, so that the object spans the entire image. Lighting is also randomized.

{\bf SDF:} We sampled 12,500 3D points and precomputed their signed distance values, i.e., the distance of a point from the object surface with the sign indicating whether or not the point is inside the surface. Following the approach of DeepSDF~\cite{park2019deepsdf}, we sampled more aggressively near the object surface to foster the learning of the object geometry.

{\bf Grasping \& Hanging:} The grasping and hanging data are 1,000 feasible grasping and hanging poses of the gripper and the hook, respectively. For grasping, we used an antipodal sampling scheme, similarly to \cite{acronym2021}, to create candidate gripper poses and checked their feasibility using Bullet~\cite{coumans2021}. For hanging, we randomly sampled collision-free hook poses and checked if it's kinematically trapped by the mug in the directions perpendicular to the hook's main axis.

\begin{figure}[t]
	\centering
	\subfigure[SDF]{
		\includegraphics[width=.32\columnwidth]{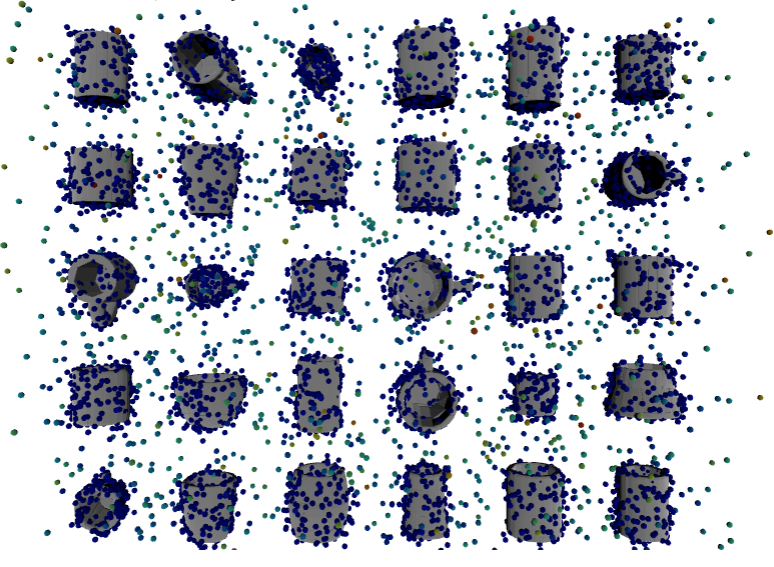}}
	\subfigure[Grasp]{
		\includegraphics[width=.32\columnwidth]{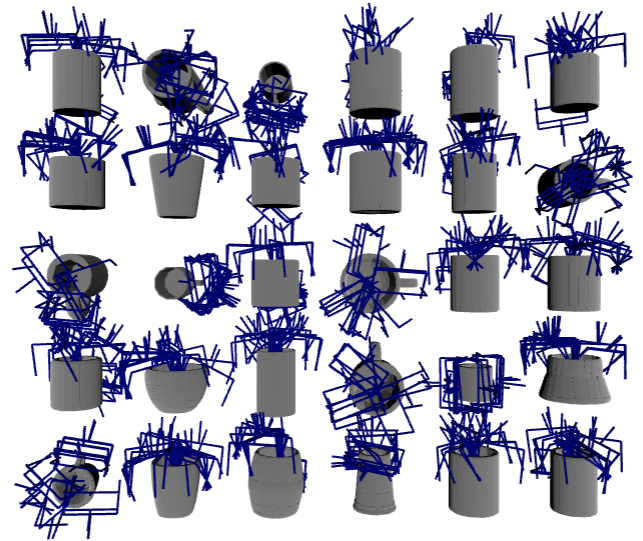}}
	\subfigure[Hang]{
		\includegraphics[width=.32\columnwidth]{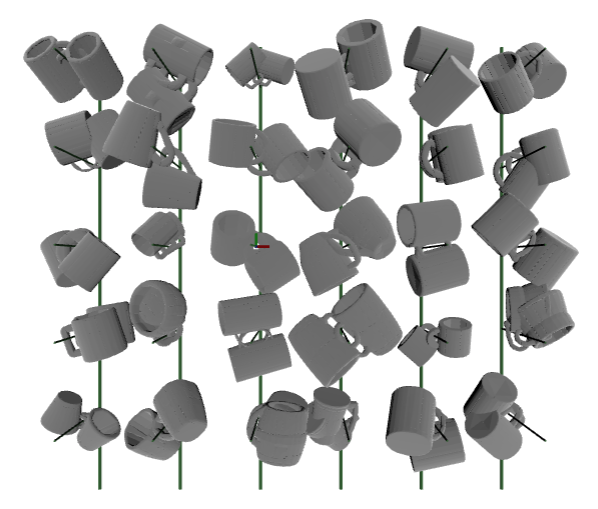}}
	\subfigure[Camera]{
		\includegraphics[width=.49\columnwidth, viewport=0 0 800 780, clip]{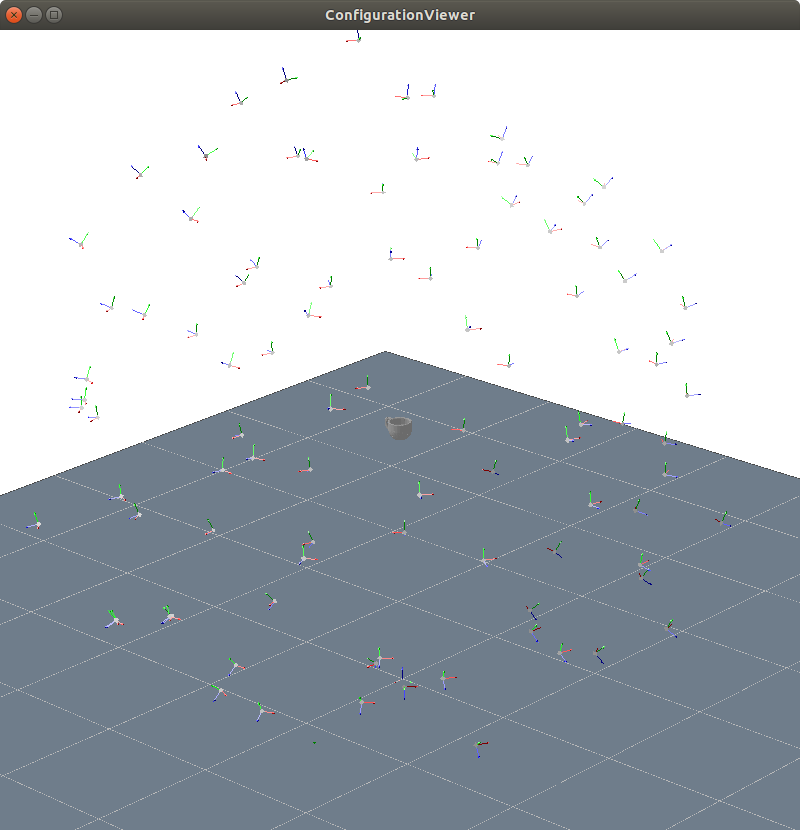}\label{fig:camera}}
	\subfigure[Image]{
		\includegraphics[width=.49\columnwidth]{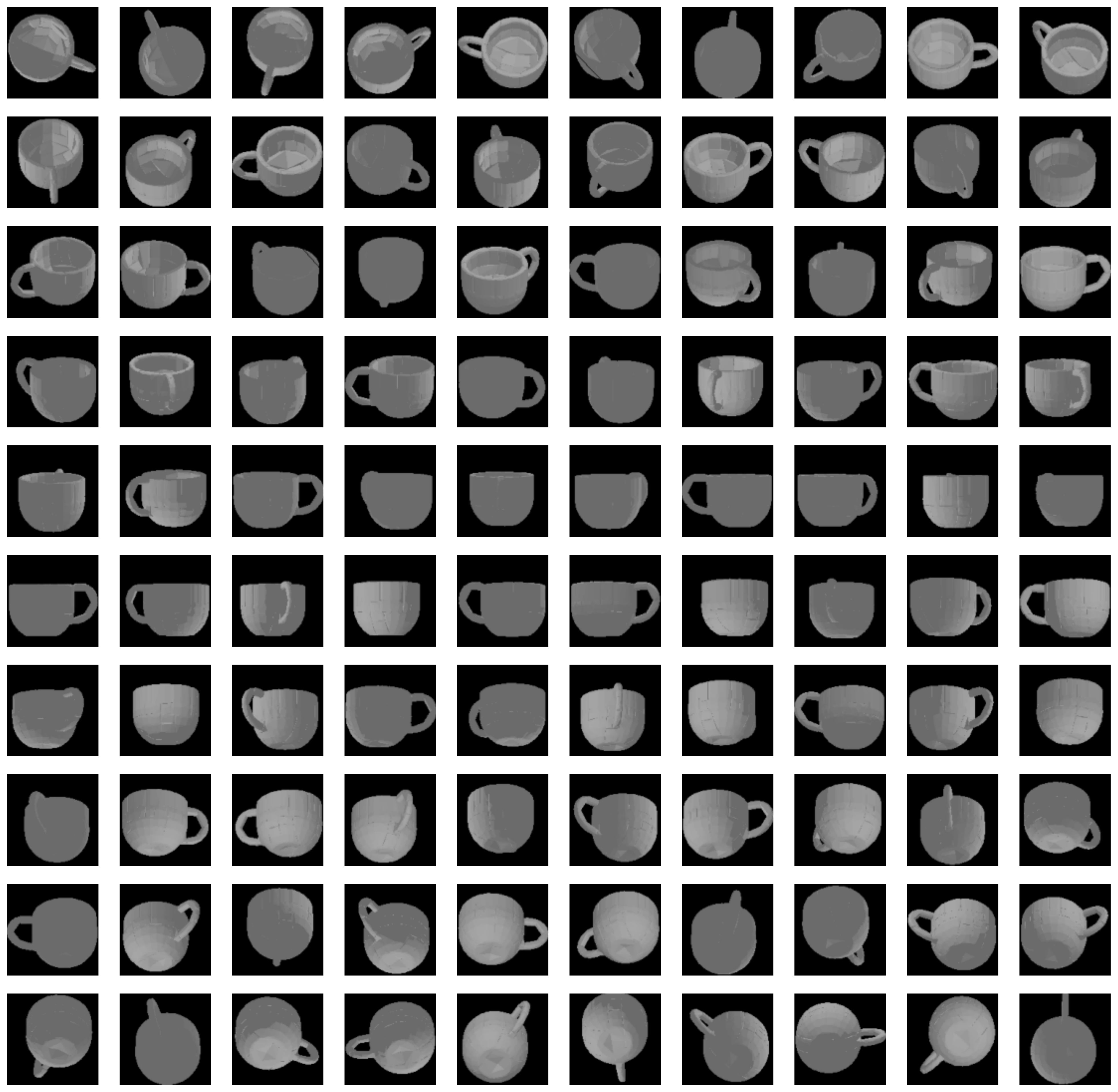}\label{fig:image}}
	\caption{Data Generation} \label{fig:data}
\end{figure}

Fig.~\ref{fig:data} shows some rough looks of the generated data. In the end, we have a dataset of:
\scriptsize\begin{align}
\left\{\left(\mathcal{I}^{1:100}, \mT^{1:100}, \mK^{1:100}, \vp^{1:12500}, SDF^{1:12500}, \vq_\text{grasp}^{1:1000}, \vq_\text{hang}^{1:1000}\right)^{(i)}\right\}_{i=1}^{131}, \nonumber
\end{align}\normalsize
which we divided into 78 training, 25 validation, 28 test sets.

\subsection{Data Augmentation}
\begin{figure}[t]
	\centering
	\subfigure[Before augmentation]{
		\includegraphics[width=.8\columnwidth]{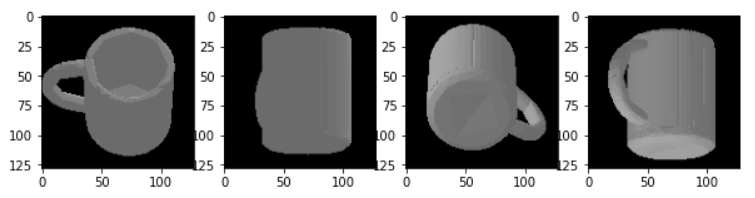}}
	\subfigure[After augmentation]{
		\includegraphics[width=.8\columnwidth]{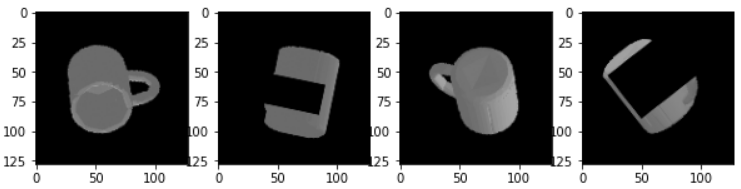}}
	\caption{Image Data Augmentation} \label{fig:data2}
\end{figure}
While randomizing the azimuth, elevation and distance of the camera provides all possible appearances of the object, it still cannot account for varying roll angles of the camera (i.e.\ image rotations) and off-centered images.
To show the network all possible images that it can encounter when deployed later and to mitigate the size-ambiguity issue, we propose to use a data augmentation technique based on Homography warping: In each iteration, for a randomly sampled set of images, we artificially perturb the roll angle of each camera and the estimated object center position (at which the cameras are looking). Also, \textit{fov} is modified as if the radius of the bounding sphere is $15\text{ cm}$ so that smaller objects can appear smaller in the transformed images. This results in new rotation matrices, $\hat\mR$, and intrinsic matrices, $\hat\mK$, of the cameras.
Because the original and new cameras are at the same position, images taken from them can be transformed one another through the Homography warping, as also illustrated in Fig.~\ref{fig:mv_overview}. 
Therefore, we compute the corresponding Homography transformation matrix and warp the images accordingly:
\begin{align}
\mathcal{W}(\hat{\mR},\hat{\mK}):   \begin{bmatrix} u \\ v \\ 1 \end{bmatrix} \mapsto w\hat{\mK}\hat{\mR}^{T}\mR\mK^{-1} \begin{bmatrix} u \\ v \\ 1 \end{bmatrix}. \label{eq:warping}
\end{align}
Random cutouts are also applied to address occlusion.
Fig.~\ref{fig:data2} depicts how this image augmentation works.

For grasping and hanging, i.e., $\text{task}\in\{\text{grasp},\text{hang}\}$, we generate random poses $\hat{\vq}_\text{task}\in SE(3)$ in each iteration as a weighted sum of a (randomly picked) feasible pose and a random pose $\hat{\vq}_\text{task} = t \vq_\text{feasible} + (1-t) \vq_\text{rand},~t\sim\mathcal{U}(0,1)$ where the position of $\vq_\text{rand}$ is from the normal distribution and its quaternion is sampled uniformly, to encourage more precise prediction around the constraint manifolds.
The training target is then, similarly to \cite{atzmon2020sal}, the unsigned distances (in $SE(3)$) of $\hat{\vq}_\text{task}$ from the set of the feasible poses:
\begin{align}
d_\text{task} = \min_{j\in\{1,\ldots,1000\}}||\vq-\vq^j_\text{task}||_2.
\end{align}

\subsection{Loss Function}
\begin{figure}[t]
	\centering
	\subfigure{
		\includegraphics[width=.35\columnwidth]{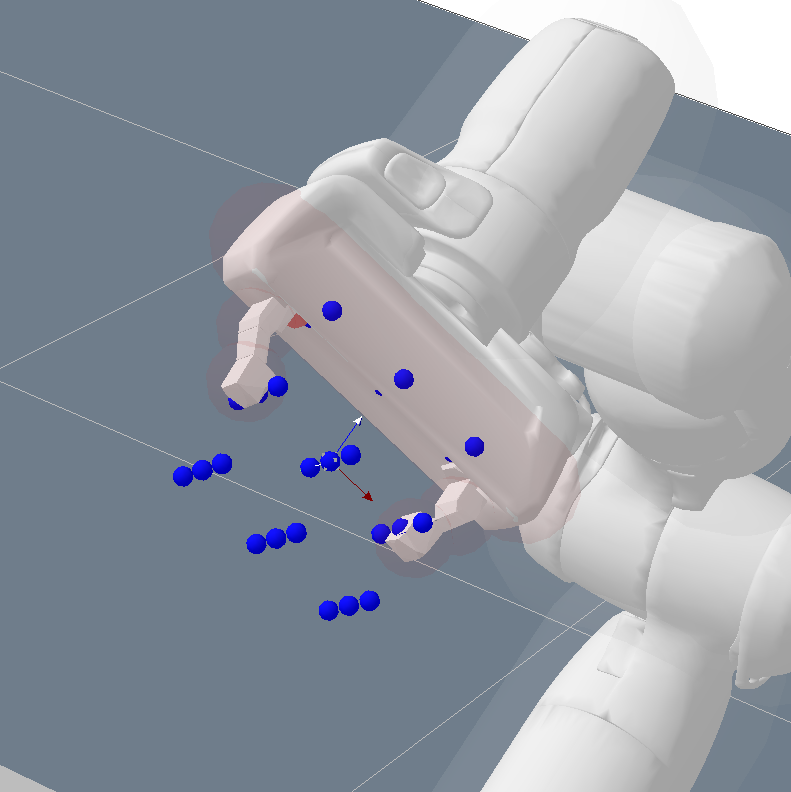}}
	\subfigure{
		\includegraphics[width=.35\columnwidth]{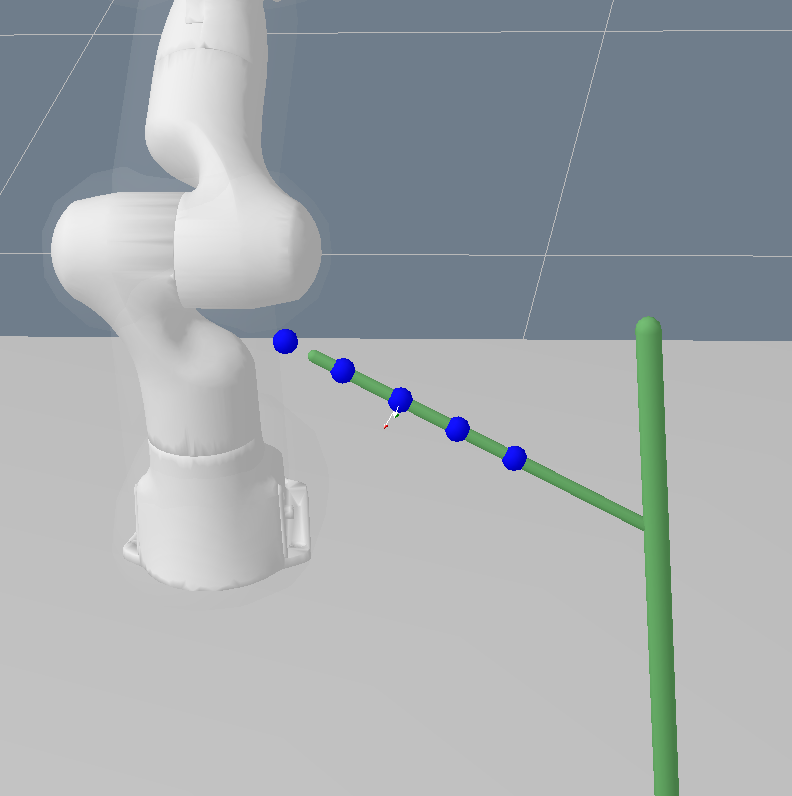}}
	\caption{The grasp and hang interaction points are defined as $(3\times3\times3)$ grid points around the gripper center and 5 points along the hook's main axis, respectively.
	}\label{fig:keypoints}
\end{figure}

The whole architecture, backbone and three task heads, is trained end-to-end.
In each iteration, we choose a minibatch of mugs for which a subset of augmented images with their camera parameters,  $\hat{\mathcal{V}} = \{(\hat{\mathcal{I}}^1, \hat{\mT}^1, \hat{\mK}^1), ..., (\hat{\mathcal{I}}^{N_\text{view}}, \hat{\mT}^{N_\text{view}}, \hat{\mK}^{N_\text{view}})\}$, a subset of SDF data, $\left(\vp^{1:N_\text{SDF}}, SDF^{1:N_\text{SDF}}\right)$, and the grasping/hanging data, $\left(\hat{\vq}_\text{task}^{1:N_\text{task}}, d_\text{task}^{1:N_\text{task}}\right)$, are sampled.
The images are encoded only once per iteration and then the SDF, grasping, hanging features are queried at the sampled points and poses.
The overall loss is given as 
\begin{align}
\mathcal{L}_\text{total} = \mathcal{L}_\text{sdf} + \mathcal{L}_\text{grasp} + \mathcal{L}_\text{hang},
\end{align}
where we used a typical $L1$ loss for SDFs, i.e. 
\begin{align}
\mathcal{L}_\text{sdf} = \frac{1}{N_\text{SDF}}\sum_{i=1}^{N_\text{SDF}} |\phi_\text{sdf}(\vp^i)-SDF^i|,
\end{align}
and the sign-agnostic $L1$ loss in \cite{atzmon2020sal} for grasping and hanging, i.e., $\forall\text{task}\in\{\text{grasp, hang}\}$
\begin{align}
\mathcal{L}_\text{task} = \frac{1}{N_\text{task}}\sum_{i=1}^{N_\text{task}} \left|\left|\phi_\text{task}(\hat{\vq}_\text{task}^i; \hat{\mathcal{V}})\right|-d^i_\text{task}\right|. \label{eq:SAL}
\end{align}
We used $N_\text{views}=4,~N_\text{SDF}=300,~N_\text{grasp}=100,~N_\text{hang}=100$ and the considered interaction points are shown in Fig.~\ref{fig:keypoints}.

\section{Sequential Manipulation Planning with DVC}\label{sec:planning}
\begin{figure}[h]
	\centering
	\scalebox{0.75}{
		\begin{tikzpicture}
		\tikzset{block/.style={rectangle, draw,fill=white}}
		
		\pgfdeclarelayer{background}
		\pgfdeclarelayer{foreground}
		\pgfsetlayers{background,main,foreground}
		\node[block, draw, align=center, rounded corners] at (0,0) (robotConfig) {robot\\configuration\\$\vx\in\R^{n_x}$};
		
		\node[block, draw, right=.5cm of robotConfig, align=center] (FK) {forward\\kinematics};
		\draw[->, very thick] (robotConfig) -- (FK);

		\node[block, draw, right=.5cm of FK, align=center, rounded corners] (robotPose) {robot or static\\frame's pose\\$\vq\in SE(3)$};
		\draw[->, very thick] (FK) -- (robotPose);
		
		\node[block, draw, above=.4cm of robotConfig, align=center, rounded corners] (imgs) {raw images\&masks,\\camera poses,\\intrinsics: $\mathcal{V}_\text{raw}$};
		
		\node[block, draw, right=.3cm of imgs, align=center] (MV) {multi-view\\processing\\(Sec.\ref{sec:perception})};
		\draw[->, very thick] (imgs) -- (MV);
		
		\node[block, draw, right=.3cm of MV, align=center, rounded corners] (objImgs) {obj images,\\cam poses,\\intrinsics: $\mathcal{V}_\text{obj}$};
		\draw[->, very thick] (MV) -- (objImgs);
		
		\node[block, draw, right=.5cm of robotPose, align=center] (feature) {\bf Deep\\\bf Visual\\\bf Constraint\\(Sec.\ref{sec:LGP})};
		\node[block, draw, right=.5cm of feature, align=center, rounded corners] (h) {$h$};
		
		\draw[->, very thick] (robotPose) -- (feature);
		\draw[->, very thick] (objImgs.east) -| ++(0.5,0.0) -| (feature.north);
		
		\node[block, draw, below=.3cm of robotConfig, align=center, rounded corners] (objTrans) {object's rigid\\transformation\\$\delta\vq\in SE(3)$};
		\draw[->, very thick] (objTrans.east) -| ++(1.5,0.0) -| (feature.south);
		\draw[->, very thick] (feature) -- (h);
		
		\end{tikzpicture}}
	\caption{DVCs in manipulation planning. The multi-view preprocessing converts the scene images into the object-centric images. The robot or static frames' poses are computed via forward kinematics.}\label{fig:LGPoverview}
\end{figure}
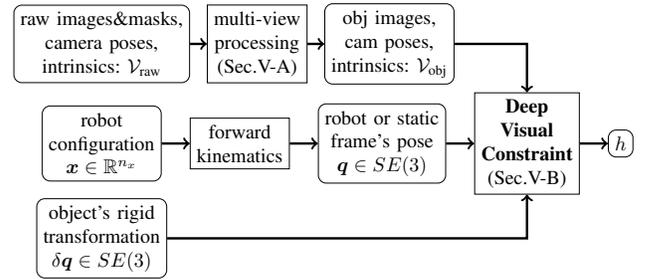
In order to compute a full trajectory of the robot and objects that it interacts with, the learned features can be integrated as differentiable constraints into any constraint-based trajectory optimization framework, for which we adopt Logic-Geometric Programming (LGP)~\cite{toussaint2018differentiable}.
In typical manipulation scenes, however, cameras are equipped such that their views cover a wide range of the environment, so we need to transform the raw images of the entire scene into object-centric ones to pass them to the network.
To that end, as depicted in Fig.~\ref{fig:LGPoverview}, we propose the multi-view preprocessing to compute the object-centric images and corresponding camera extrinsics/intrinsics.
In addition to the multi-view preprocessing, we wrap the learned DVCs with forward kinematics to evaluate the learned constraints from the scene images and the optimization variables, i.e., robot's joint configuration and object's transformation.
Section \ref{sec:perception} discusses the proposed multi-view warping procedure and Section \ref{sec:LGP} presents how the learned features serve as constraints to sequential manipulation planning problems.

\subsection{Multi-View Preprocessing} \label{sec:perception}
\begin{figure}[h]
	\centering\includegraphics[width=.9\columnwidth]{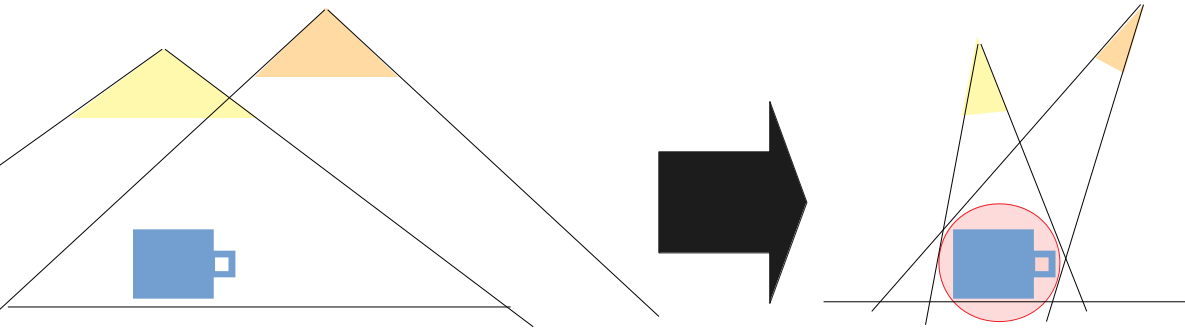}
	\caption{Illustration of Multi-View Preprocessing: Two images taken at the same location but different orientations \& \textit{fov} are related by a homography}\label{fig:mv_overview}
\end{figure}
\begin{figure}[t]
	\centering
	\subfigure[Raw images and masks]{
		\includegraphics[width=1\columnwidth]{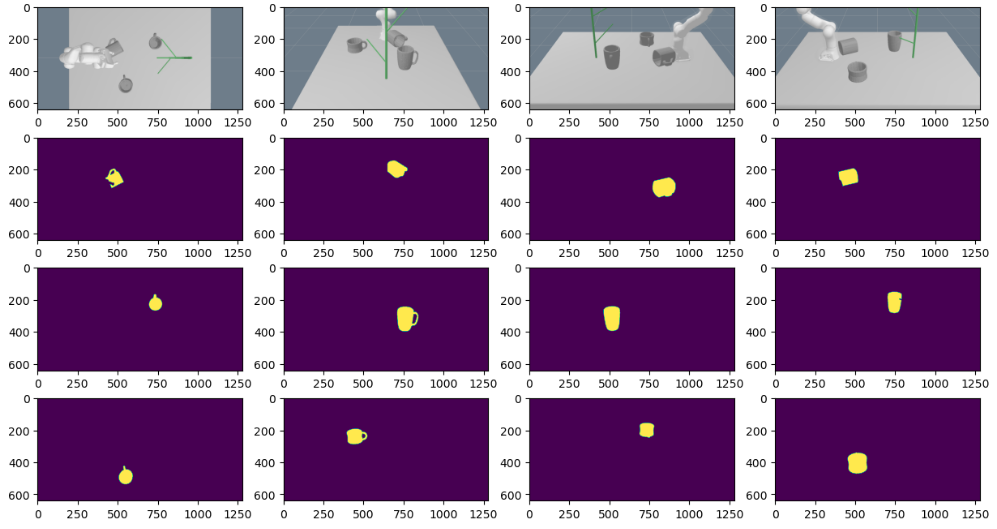}}
	\subfigure[Object-centric images warped via the multi-view processing]{
		\includegraphics[width=1\columnwidth]{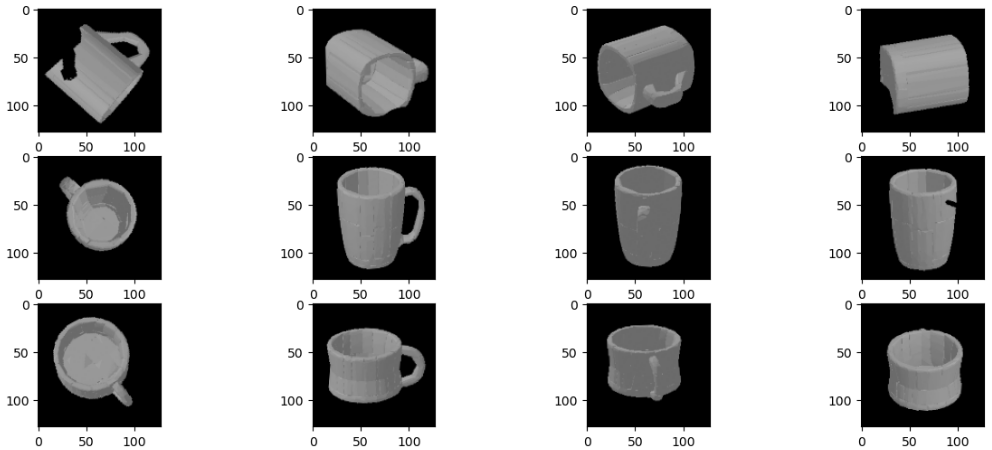}}
	\caption{Multi-view processing}\label{fig:multiview0}
\end{figure}

As illustrated in Fig.~\ref{fig:mv_overview}, multi-view processing finds a bounding ball and warps the raw images via the Homography warping.
Let $\mathcal{M}^n\in\{0,1\}^{W\times H}$ be the object masks available along with the raw images $\mathcal{I}^n_\text{raw},~\forall n=1,...,N_\text{cam}$.
We first find a position and radius of the minimal bounding sphere such that the warped images contain all the object pixels in the original images by solving the following optimization problem:
\begin{align}
&\min_{ \vp\in\R^3, r\in\R^+}  r, \\
&\text{s.t. }\forall_{(u_n, v_n, n)\in \{(u',v',n'); \mathcal{M}_{n'}(u', v')=1,\forall n'\in\{1,...,N_\text{cam}\}\}} : \nonumber\\ &~~~~~~~~~~~~~~~~~~~~~~~~~~~~~~~~||\mathcal{W}(\hat{\mR}^n,\hat{\mK}^n)(u_n, v_n)||_2 < 1, \nonumber
\end{align}
where $\hat{\mR}$ can be obtained from the sphere center $\vp$ and the camera position $\vt$, and $\hat{\mK}$ is computed as $fov = 2\arcsin(||\vt-\vp||_2/r)$; the warping $\mathcal{W}$ can then be defined as in (\ref{eq:warping}).
After solving the above optimization, we fix the camera orientations $\hat{\mR}$, change the intrinsics as if the bounding sphere has a radius of $15 \text{ cm}$ and finally warp the raw images accordingly.
Fig.~\ref{fig:multiview0} shows the raw images from an example environment and the results of the multi-view processing.

\subsection{Logic-Geometric Programming for Manipulation Planning}\label{sec:LGP}
The core concept of manipulation is the rigid transformations of objects~\cite{toussaint2018differentiable,21-driess-CORL}; e.g., while grasped, the object moves with the gripper.
For an object transformed by $\delta\vq\in SE(3)$, we define a rigid transformation of the interaction feature as:
\begin{align}
T(\delta\vq)[\phi_\text{task}](\cdot) := \phi_\text{task}\left(\delta\vq^{-1}\cdot\right),
\end{align}
which is equivalent to rigidly transforming the representation function as $T(\delta\vq)[\psi](\cdot) = \psi\left(\mR(\delta\vq)^T\left(\cdot-\vt(\delta\vq)\right)\right)$\footnote{We dropped $\mathcal{V}$ from $\phi_\text{task}(\vq; \mathcal{V})$ for the simplicity of notation.}.
Through the function composition of the forward kinematics, $FK$, and the (transformed) feature, $\phi_\text{task}$, i.e.,
\begin{align}
H_\text{task}(\vx, \delta\vq) := \left(T(\delta\vq)[\phi_\text{task}]\circ FK\right)(\vx),
\end{align}
we obtain an interaction feature as a function of a robot joint configuration $\vx$ and object's rigid transformation $\delta\vq$. 

Now we are ready to formalize manipulation planning problems.
For an $n_x$-joint robot and $n_o$ rigid objects, LGP is a hybrid optimization problem over the number of phases $K\in\mathbb{N}$, a sequence of discrete actions $\ra_{1:K}$ and sequences of the robot joint configurations $\vx_{1:KT},~\vx\in\R^{n_x}$ and the object's rigid transformations $\delta\vq_{1:KT},~\delta\vq\in SE(3)^{n_o}$.
The trajectory is discretized into $T$ steps per phase.
A discrete action $\ra_{k}$ describes which interaction should be fulfilled at the end of the phase $k$, i.e., which mug to pick or on which hook to hang the grasped mug, and uniquely determines a symbolic state $\rs_k = \text{succ}(\rs_{k-1}, \ra_k)$, i.e., whether each mug is grasped or hung on a particular hook.
Suppose that a discrete action sequence $\ra_{1:K}$ and the corresponding modes $\rs_{1:K}$ with $\rs_K\in\mathcal{S}_\text{goal}$ are proposed by a logic tree search.
We then define the geometric path problem as a $2^\text{nd}$ order Markov optimization~\citep{17-toussaint-Newton}:
\begin{align}\label{eq:LGP}
&\min_{ \substack{\vx_{1:KT}\\\delta\vq_{1:KT}} }  \sum_{t=1}^{KT} f\left(\vx_{t-2:t}\right),\\
&~\text{s.t.}\forall_{\substack{H\in\sH(\rs_k, \ra_k),\\k\in\{1,\ldots, K\}}} :  H\left((\vx_{t-2:t}, \delta\vq_{t-2:t}^i)_{(t,i)\in\mathcal{I}_H(\rs_k,\ra_k)}\right) = 0, \nonumber
\end{align}
where the initial joint states $\vx_{-1:0}$ and objects' transformations $\delta\vq_{-1:0}=0$ are given.
Note that $\delta\vq$ denotes rigid transformations applied to objects' implicit representations, not their absolute poses.
$f$ is a path cost that penalizes squared accelerations of the robot joints, but it can be more general if necessary.
$\sH(\rs_k,\ra_k)$ is a set of constraints the symbolic state and action impose on the geometric path at each phase $k(t) = \lfloor t/T \rfloor$; these constraints include physical consistency, collision avoidance, and the learned interaction constraints that ensure the success of the discrete action $a_k$.
Lastly, $\mathcal{I}_H(\rs_k,\ra_k)$ decides the time slice and object index that are subject to the constraint $H$.
Appendix \ref{app:constraints} introduces the set of imposed constraints in detail.
As all the cost and constraint terms are differentiable and their Jacobians/Hessians are sparse, we can solve this optimization problem efficiently using the augmented Lagrangian method with the Gauss-Newton approximation~\citep{17-toussaint-Newton}.

\section{Experiments}
\subsection{Performance of Learned Features}\label{sec:learned feature}

\begin{table*}[t]
	\caption{Individual Feature Evaluation with 4 views (Training / Test)}
	\label{tab:individual}
	\begin{center}
		\begin{tabular}{lcccc}
			\multicolumn{1}{l}{}   &\multicolumn{1}{c}{IoU} & \multicolumn{1}{c}{ Chamfer-$L_1$ ($\times10^{-3}$)} &\multicolumn{1}{c}{Grasp+c ($\%$)} &\multicolumn{1}{c}{Hang+c ($\%$)}
			\\ \hline 
			PIFO         &  0.816 / 0.656  & 5.26 / 6.90 &  \textbf{88.1 / 82.5} & \textbf{94.0 / 78.9}\\
			Global Image Feature & 0.697 / 0.581 & 7.42 / 9.49  & 82.7 / 75.7  & 91.2 / 78.2 \\
			Vector Object Representation & 0.036 / 0.014 & 38.6 / 39.7 & 0.5 / 0.4 & 0.0 / 0.0 \\
			SDF Object Representation & \textbf{0.845 / 0.667} & \textbf{4.90 / 6.83} & 67.9 / 64.3 & 3.7 / 4.3
			\\ \hline 
			PIFO (2 views)  &  0.760 / 0.577  & 6.14 / 8.84 &  82.9 / 77.1 & 88.2 / 72.1\\
			PIFO (8 views)  &  0.851 / 0.683  & 4.78 / 6.34 &  88.7 / 85.0 & 96.5 / 82.5
			\\ \hline 
			GT Mesh + HE & - & - & 62.8 / 75.0 & 94.9 / 92.9\\ 
			Recon. + HE & - & - & 66.7 / 42.9 & 78.2 / 60.7
		\end{tabular}
	\end{center}
\end{table*}

\begin{figure}[h]
	\centering
	\subfigure[PIFO]{
		\includegraphics[width=1.\columnwidth]{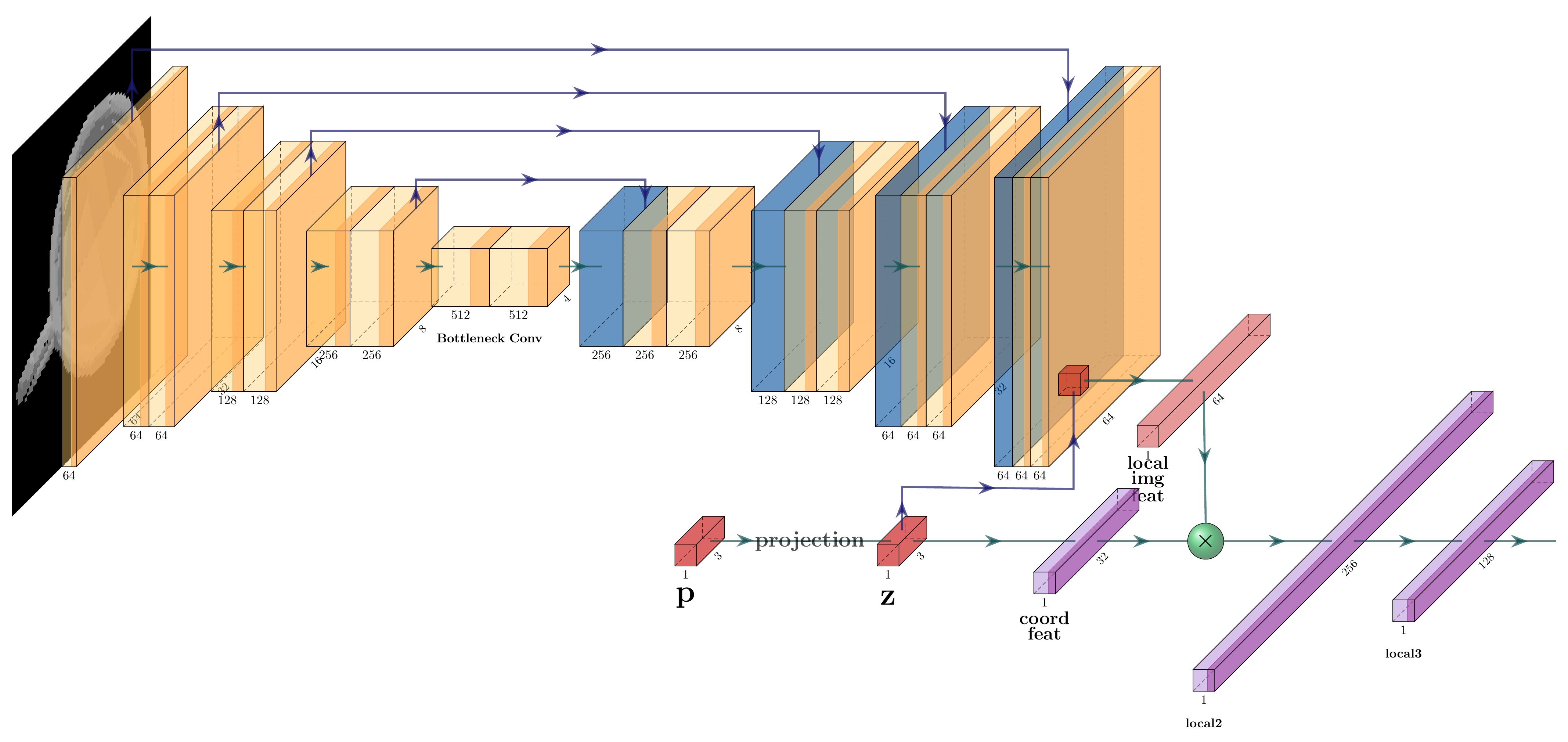}}
	\subfigure[Global Image Feature]{
		\includegraphics[width=.45\columnwidth]{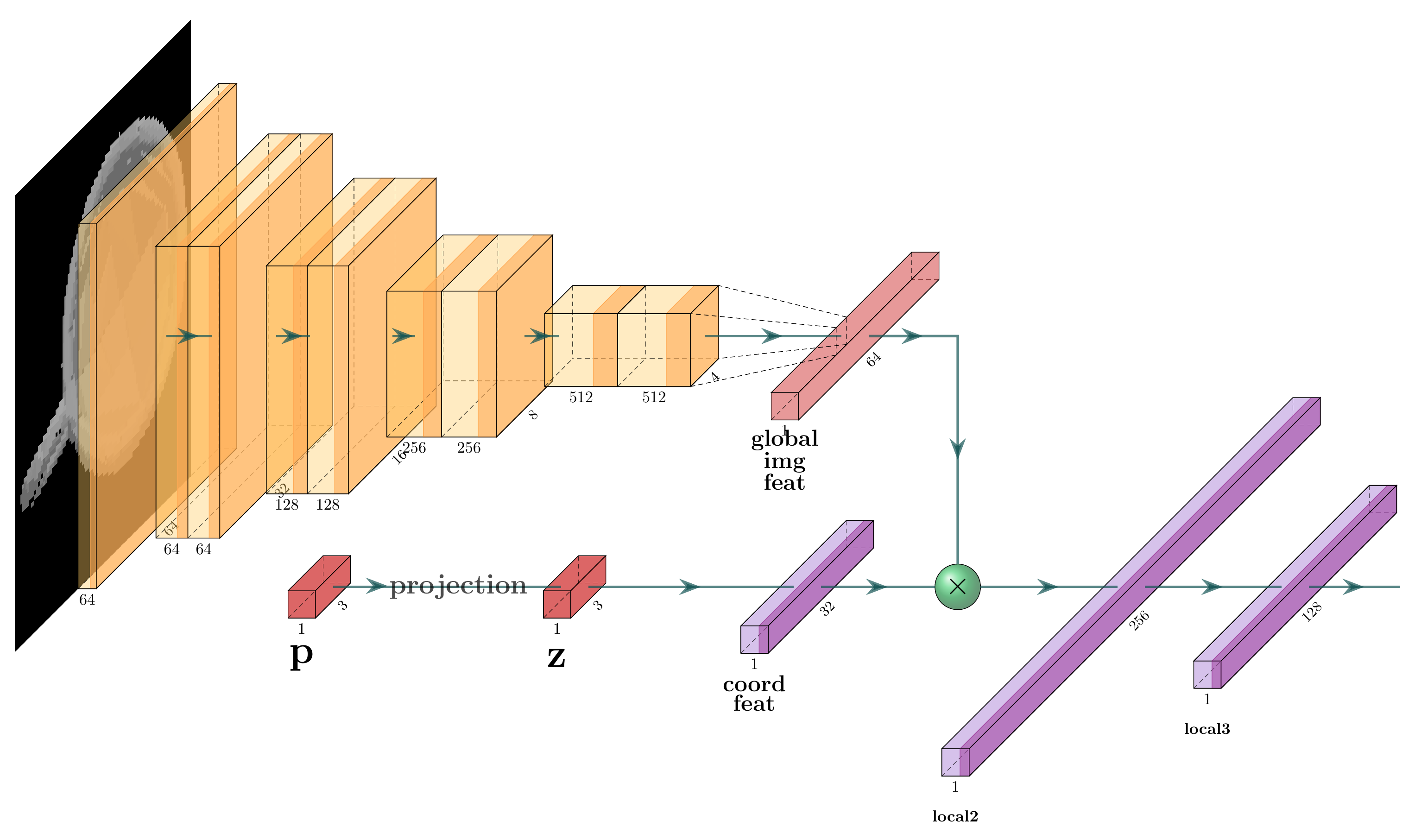}\label{fig:baseline_noPixelAligned}}
	\subfigure[Vector Object Represnetation]{
		\includegraphics[width=.45\columnwidth]{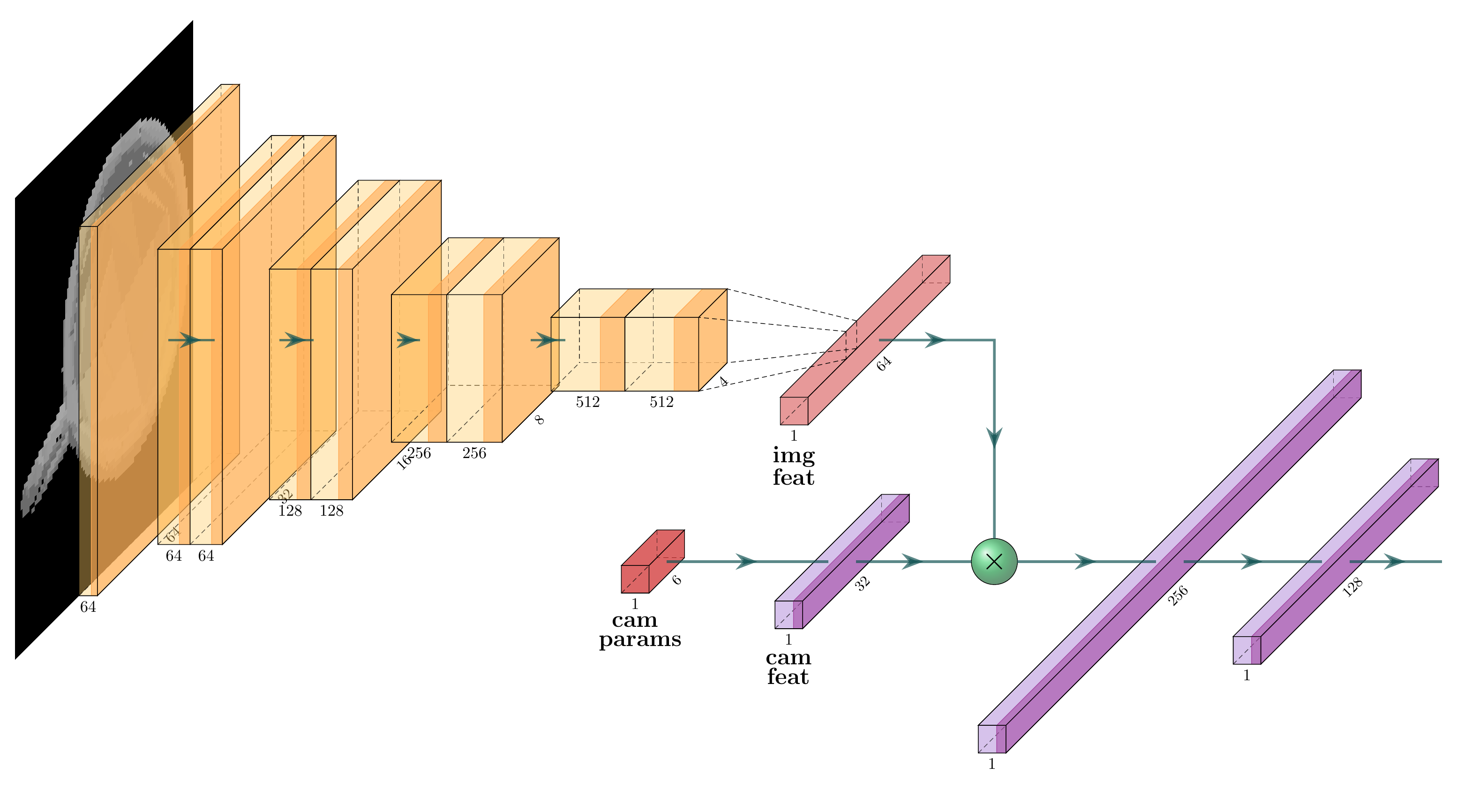}\label{fig:baseline_noImplicitFunction}}
	\caption{Baseline Networks used for comparison.}\label{fig:baseline}
\end{figure}

{\bf Baselines: }
The key techniques of the proposed framework are threefold: the pixel-aligned local feature extraction, the implicit object representation over 3D and the task-guided learning scheme.
To examine the benefits from each component, three baselines are considered.
{\it (i) Global image features}: The first baseline still represents an object as a function but the image encoder outputs a global image feature (as shown in Fig.~\ref{fig:baseline_noPixelAligned}) rather than having the pixel-aligned feature locally extracted; we used the ResNet-34 architecture as the image encoder and fixed the other model specifications.
{\it (ii) Vector object representations}: The second baseline represents an object as a finite-dimensional vector instead of a function; as shown in Fig.~\ref{fig:baseline_noImplicitFunction}, the representation network first computes the image features from the images using ResNet-34 and the camera features from the camera parameters using a couple of FCLs. Two features are then passed to another couple of FCLs to produce the object representation vector. The task heads take as input the frame's pose as well as the object representation vector.
{\it (iii) SDF representations}: The last baseline uses SDFs as object representations; the network architecture for the SDF feature remains the same, but the grasping and hanging heads take as input a set of the keypoints' SDF values instead of the $d$-dimensional representation vectors. The SDF values are detached when passed to the grasping/hanging heads so the backbone is trained by the geometry (SDF) data only.

\begin{figure}[t]
	\centering
	\subfigure[Train Mugs]{
		\includegraphics[width=.49\columnwidth]{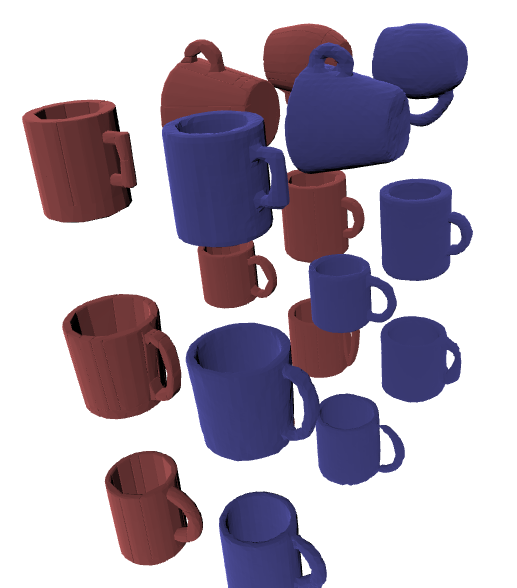}}
	\subfigure[Test Mugs]{
		\includegraphics[width=.49\columnwidth]{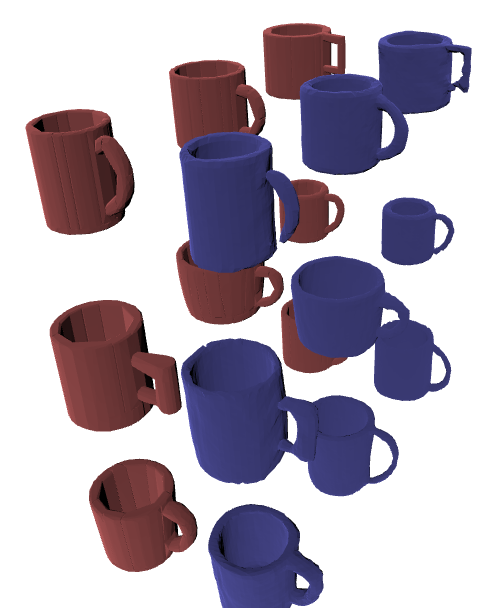}}
	\caption{Reconstruction via marching cube. Red: ground truth, Blue: reconstructed}\label{fig:recon}
\end{figure}

{\bf Evaluation Metric: }
Regarding the shape reconstruction, we report the Volumetric IoU and the Chamfer distance.
To measure these metrics, we randomly sampled 4 images from the dataset and reconstructed the meshes from the learned SDF feature using the marching cube algorithm (See Fig.~\ref{fig:recon}).
The volumetric IoU is the ratio between the intersection and the union of the reconstructed and ground-truth meshes which is (approximately) computed on the $100^3$ grid points around the objects.
To compute the Chamfer distance, we sampled 10,000 surface points from each mesh and averaged the forward and backward closest pair distances.
To evaluate the learned task features, we solved the unconstrained optimization $\hat{\vq}^* = \argmin_{\vq} ||\phi_\text{task}(\vq)||^2,~\text{task}\in\{\text{grasp, hang}\}$ using the Gauss-Newton method. 
Starting from this solution, we then solved the second optimization problem by including the collision feature (details in Appendix \ref{app:collision}), $\vq^* = \argmin_{\vq} ||\phi_\text{task}(\vq)||^2+w_\text{coll}||\phi_\text{coll}(\vq)||^2$.
Because the local optimization method can be stuck at local optima, we ran the algorithm from 10 random initial guesses in parallel and picked the best one.
The optimized pose is finally tested in simulation and the success rates (feasibility) are reported in Table \ref{tab:individual}.

\begin{figure}[t]
	\centering
	\subfigure[]{
		\includegraphics[width=.28\columnwidth]{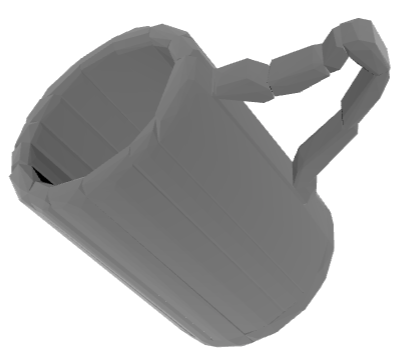}}
	\subfigure[]{
		\includegraphics[width=.28\columnwidth]{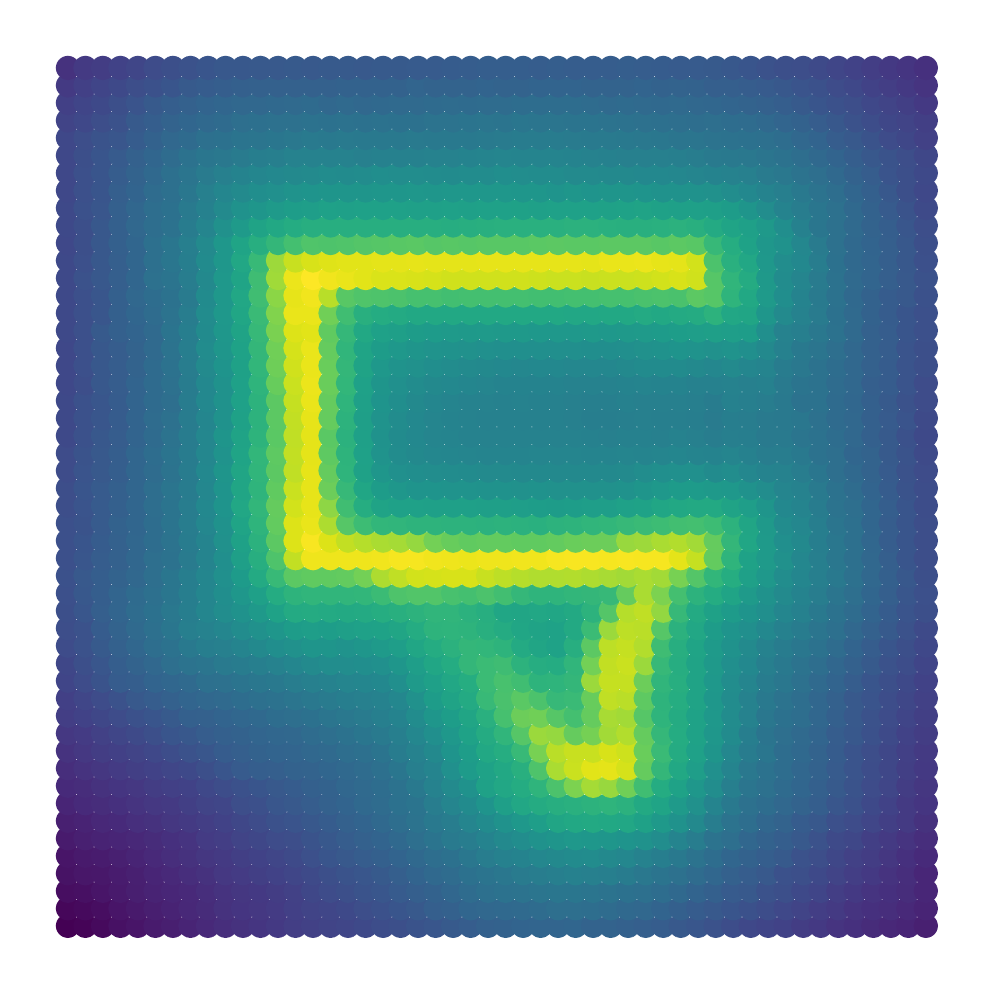}}
	\subfigure[]{
		\includegraphics[width=.28\columnwidth]{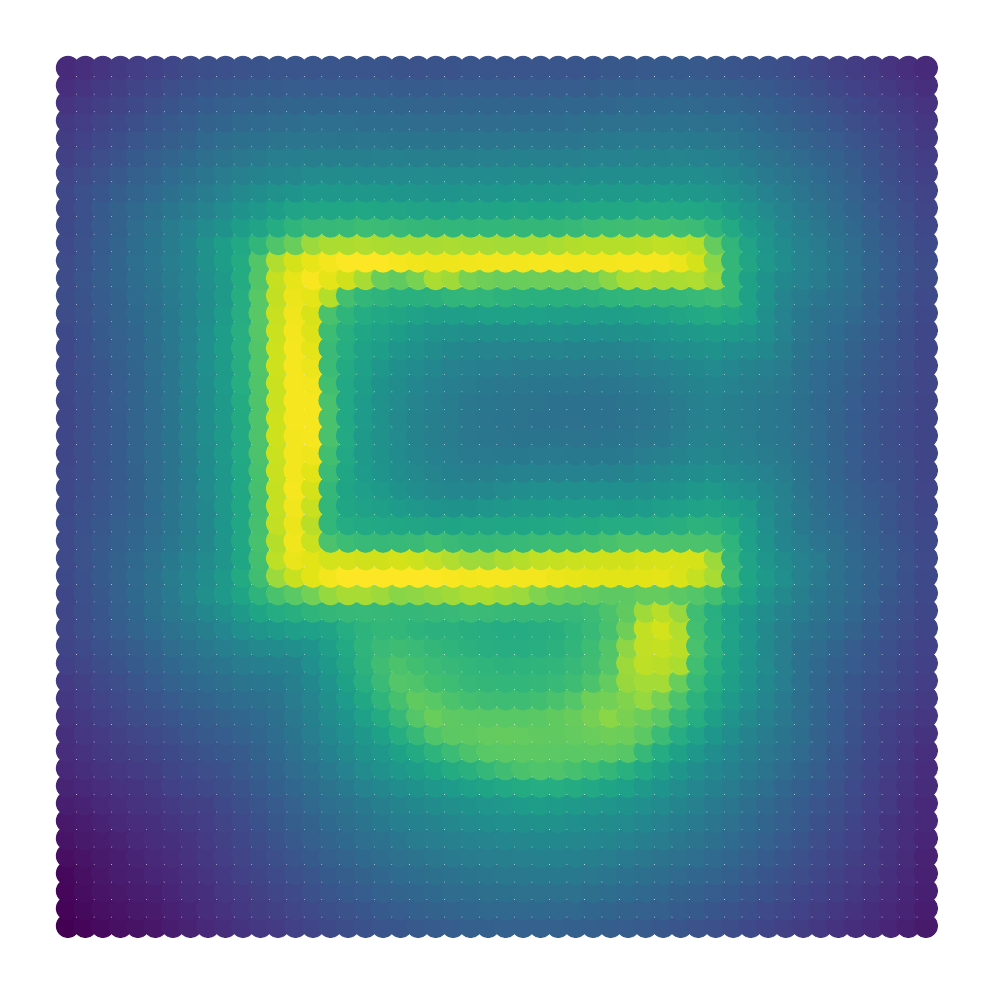}}
	\caption{SDFs predicted by (b) PIFO and (c) the global image feature model.}\label{fig:handle}
\end{figure}

{\bf Result: }
Table~\ref{tab:individual} shows that the SDF representation has the best shape reconstruction performance; PIFO is slightly worse, followed by the other two baselines.
On the other hand, the task performances of PIFO are significantly better than the others.
The SDF representation is especially worse in the hanging task, which implies that SDFs along the line are not sufficient for its feature prediction and our task-guided representation simplifies the feature prediction.
In addition, it can be observed from Fig.~\ref{fig:handle} that the pixel-aligned method was better able to capture fine-grained details than the global image feature which reconstructed the handle shape as being more ``typical".
PIFO was also trained with the different numbers of input images and it can be seen that the more images we put in, the better performance the network shows.
Tables~\ref{tab:individual_sdf} and \ref{tab:individual_opt} report all combinations of the metrics and the number of views.
	
\begin{figure}[t]
	\centering
	\subfigure[]{
		\includegraphics[width=.3\columnwidth, viewport=0 0 800 780, clip]{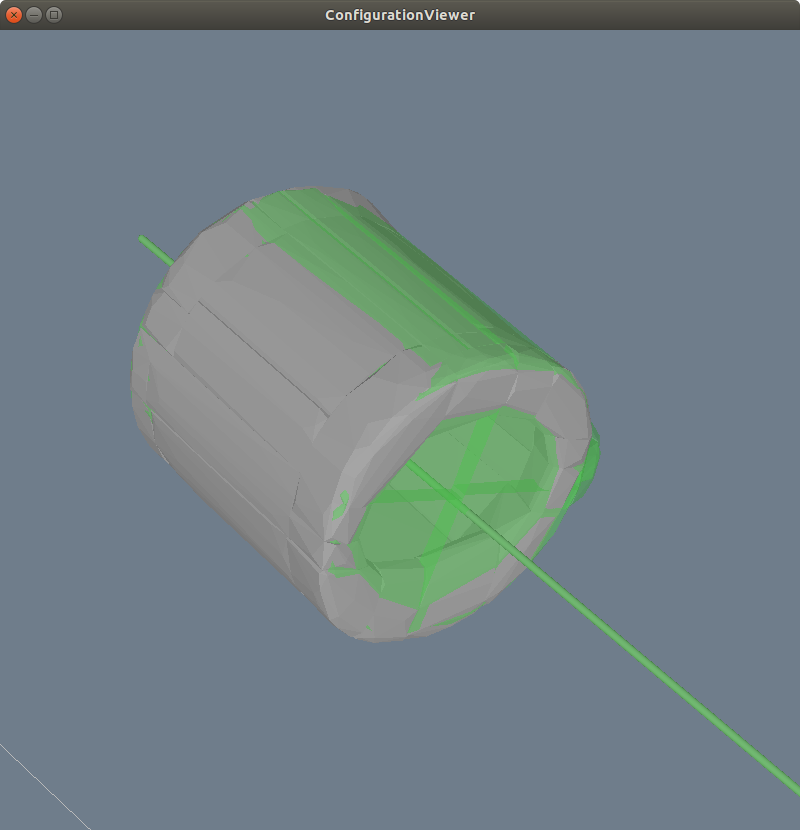}}
	\subfigure[]{
		\includegraphics[width=.3\columnwidth, viewport=0 0 800 780, clip]{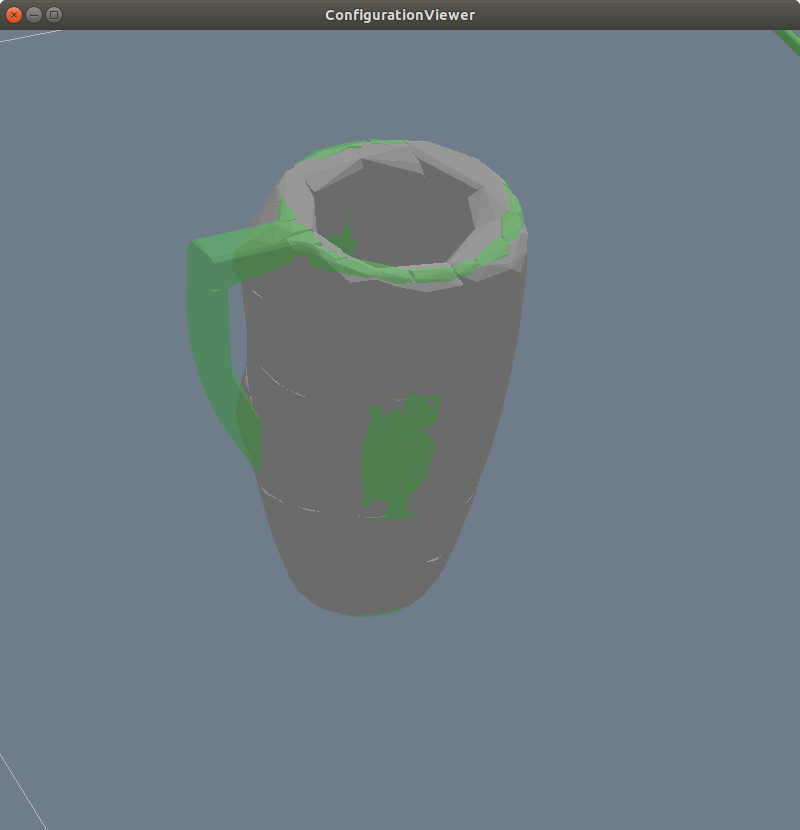}}
	\subfigure[]{
		\includegraphics[width=.3\columnwidth, viewport=0 0 800 780, clip]{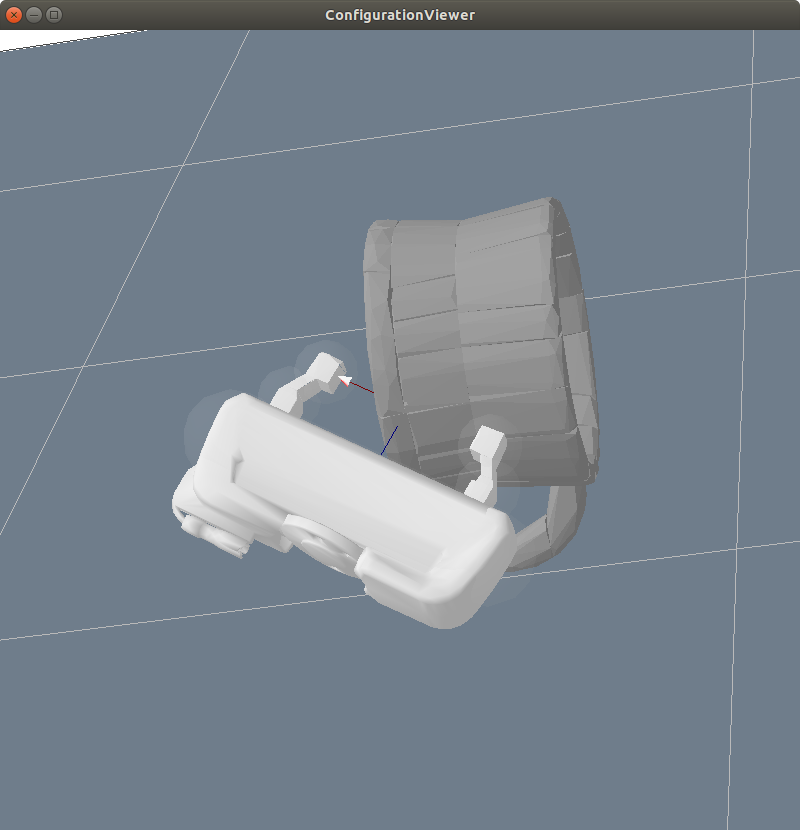}}
	\caption{Some failure cases of hand-engineered features. (a) The hand-engineered feature lead the optimizer to hang the mug through the wrongly generated hole. (Green transparent meshes represent the ground truth.) (b) The handle disappeared in reconstruction, so this part would never be grasped and the mug never be hung. (c) The hand-engineered feature generated a wrong grasping pose on the ground truth mesh.} \label{fig:HE}
\end{figure}
{\bf Hand-Engineered Constraint Models:}
We also compared our model to hand-engineered constraint models, {\it (iv) GT Mesh + HE} and {\it (v) Recon. + HE}, each of which computes constraint values based on the ground-truth meshes and the meshes reconstructed by the above SDF representations. 
Notably, Figs.~\ref{fig:HE}(a)-(b) show how vulnerable the hand-engineered constraints can be to the reconstruction error; i.e., the error is directly associated with the planning result. 
While the perception pipeline for this geometric representation is never encouraged to reconstruct the ``graspable/hangable parts'' more accurately, we can view our end-to-end representation learning via task supervision as a way to do so. 
Moreover, the hand-engineered feature sometimes produces a wrong grasping pose even for the ground truth mesh (e.g., Fig.~\ref{fig:HE}(c)). 
One can argue that a better interaction feature could be hand-designed by investigating the physics and kinematic structures more deeply, but that would require a huge amount of human insights/efforts and thus is inevitably less scalable. 
In contrast, our data-driven approach eliminates this procedure and directly learns the interaction constraint models from empirical success data of physical interactions.

\begin{figure}[t]
	\centering
	\subfigure[Single mug hanging]{
		\includegraphics[width=.31\columnwidth, viewport=100 150 500 550, clip]{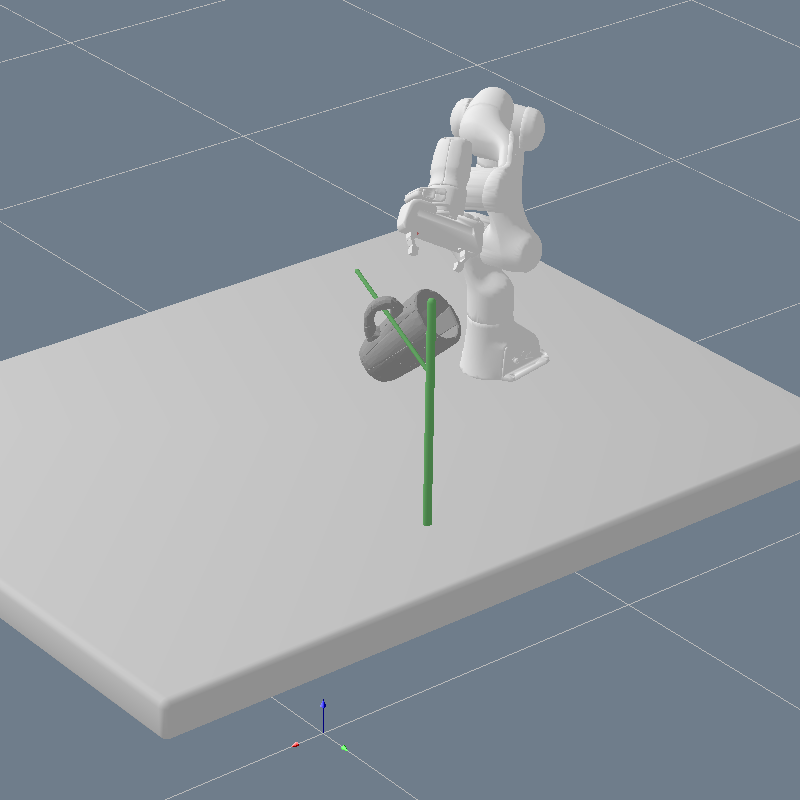}}
	\subfigure[Three-mug hanging]{
		\includegraphics[width=.31\columnwidth, viewport=50 100 450 500, clip]{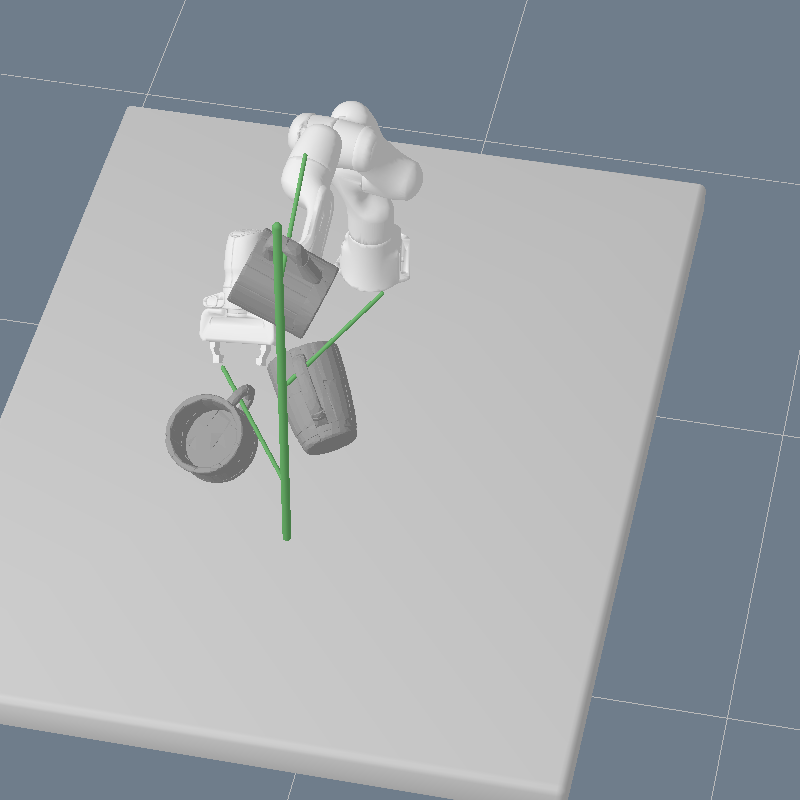}}
	\subfigure[Handover]{
		\includegraphics[width=.31\columnwidth, viewport=50 70 550 570, clip]{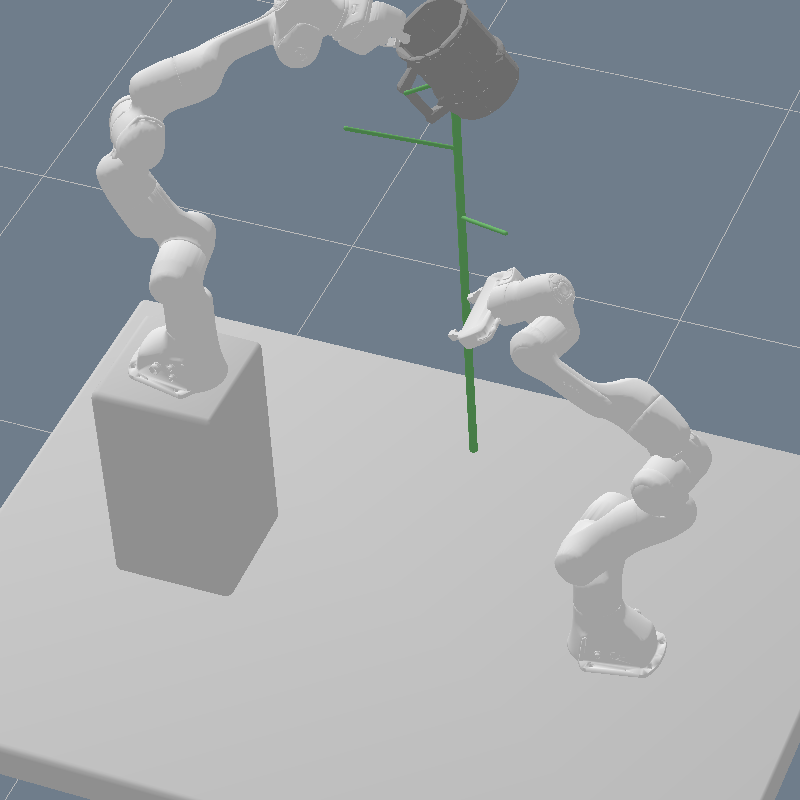}}
	\caption{Sequential manipulation scenarios}\label{fig:long_horizon}
\end{figure}	
\subsection{Sequential Manipulation Planning via LGP}
\begin{figure}[t]
	\centering
	\subfigure[Scene: Four cameras' poses are depicted as coordinate axes where the origin is the camera location, $-z$ axis (blue) is pointing the view direction, and $x$ and $-y$ axes (red and blue) are the directions of $(u,v)$ coordinate of images.]{
		\includegraphics[width=.7\columnwidth, viewport=0 0 800 780, clip]{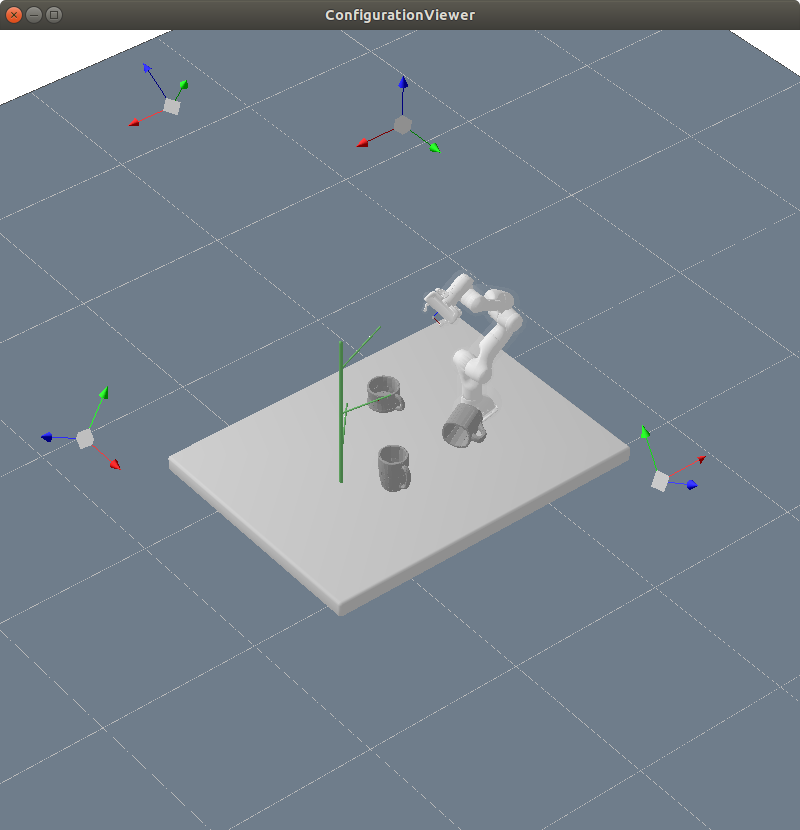}}
	\subfigure[Raw images and masks]{
		\includegraphics[width=.45\columnwidth]{perception1}}
	\subfigure[Warped images (via the multi-view processing)]{
		\includegraphics[width=.45\columnwidth]{perception2}}
	\caption{Multi-view processing}\label{fig:multiview}
\end{figure}

We first considered a basic pick \& hang task as shown in Fig.~\ref{fig:long_horizon}(a).
The environment contains one robot arm, one hook, one mug and 4 cameras (as in Fig.~\ref{fig:multiview}(a)), and the interaction modes are constrained by the discrete action sequence of [(\textsc{Grasp}, gripper, mug), (\textsc{Hang}, hook, mug)].
10 mugs were picked from each of the training and test data sets and their initial poses are randomized.\footnote{Before solving the full trajectory optimization, we first optimized each feature as in Sec.~\ref{sec:learned feature} and added small regularization terms using the optimized poses to guide the optimizer away from local optima.}
When the optimized trajectories are executed in the Bullet simulation, the success rates on the train and test mugs were 50 \% and 40 \%, respectively.
If we allow the method to re-plan and execute when it failed, the success rates increased to 90\% and 70\%, respectively \href{https://youtube.com/playlist?list=PL9pnj8nG83OfROuTSSxCEego78gzeUsA3}{[video1]}.

\begin{figure}[t]
	\centering
	\subfigure[t=5]{
		\includegraphics[width=.32\columnwidth]{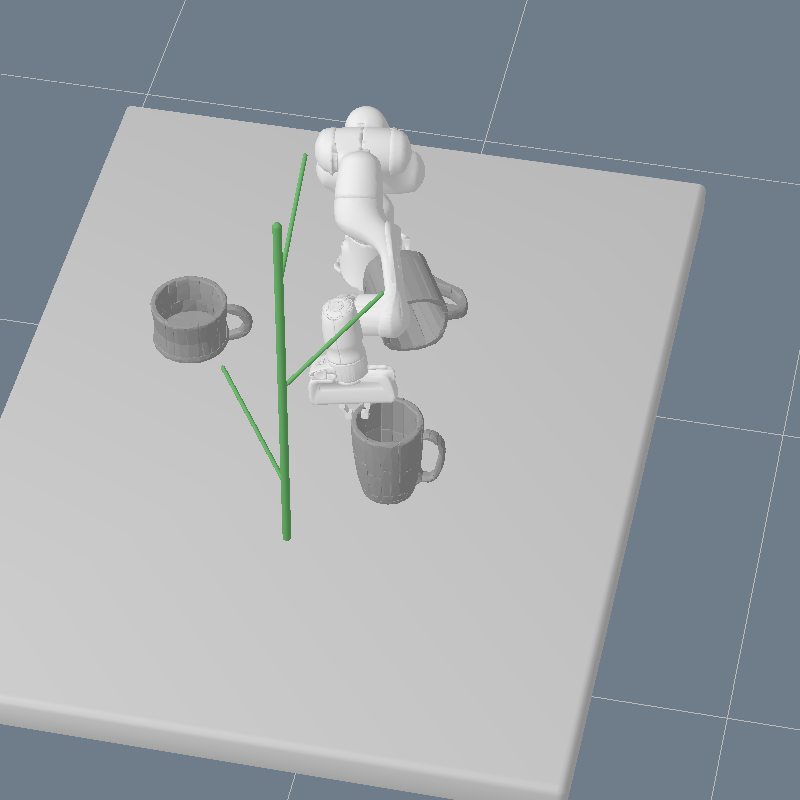}}
	\subfigure[t=8.5]{
		\includegraphics[width=.32\columnwidth]{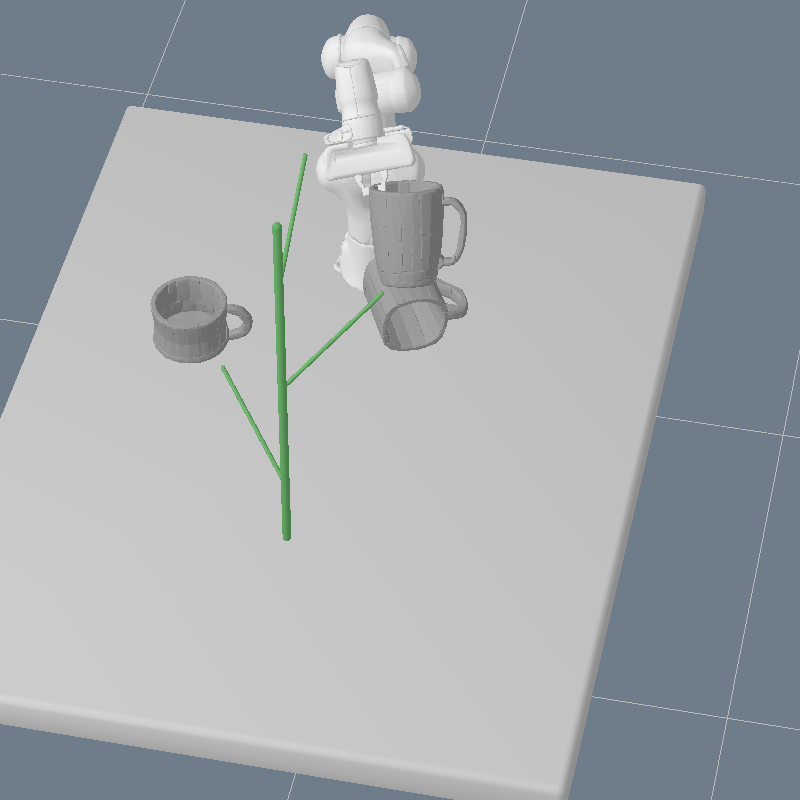}}
	\subfigure[t=10]{
		\includegraphics[width=.32\columnwidth]{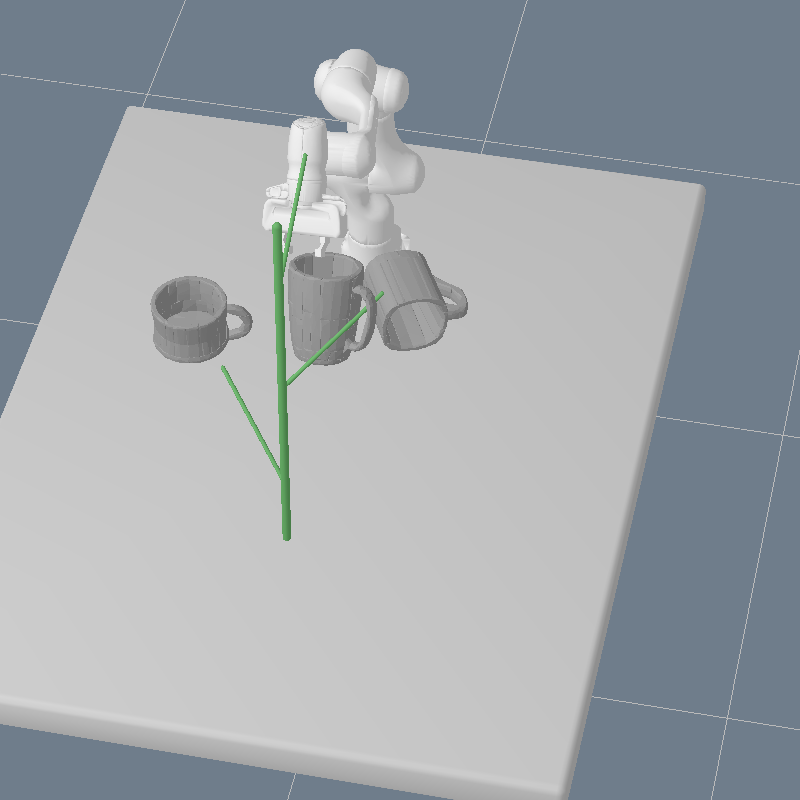}}
	\subfigure[t=15]{
		\includegraphics[width=.32\columnwidth]{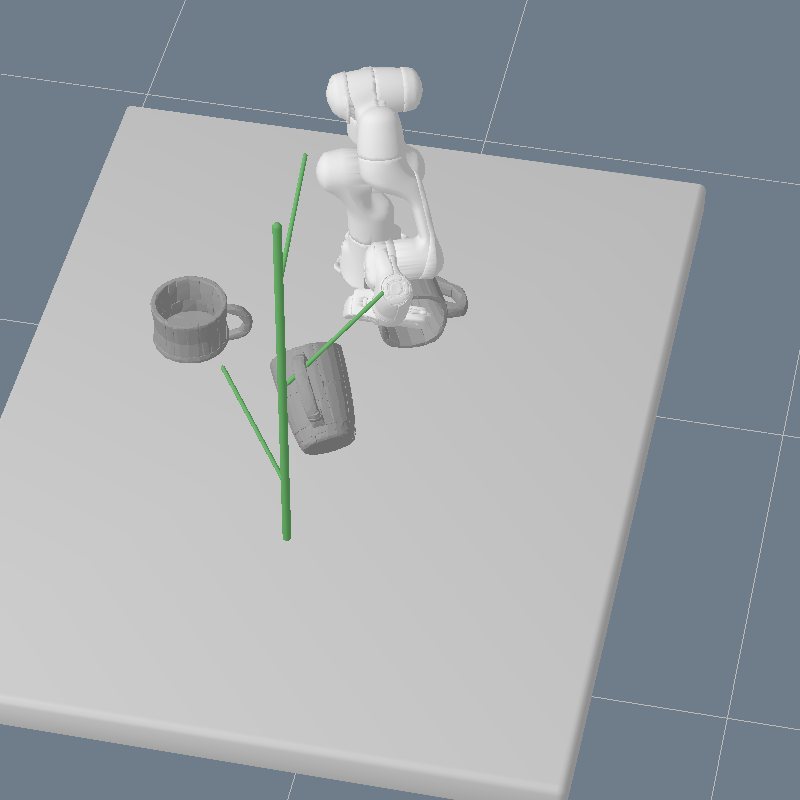}}
	\subfigure[t=18.5]{
		\includegraphics[width=.32\columnwidth]{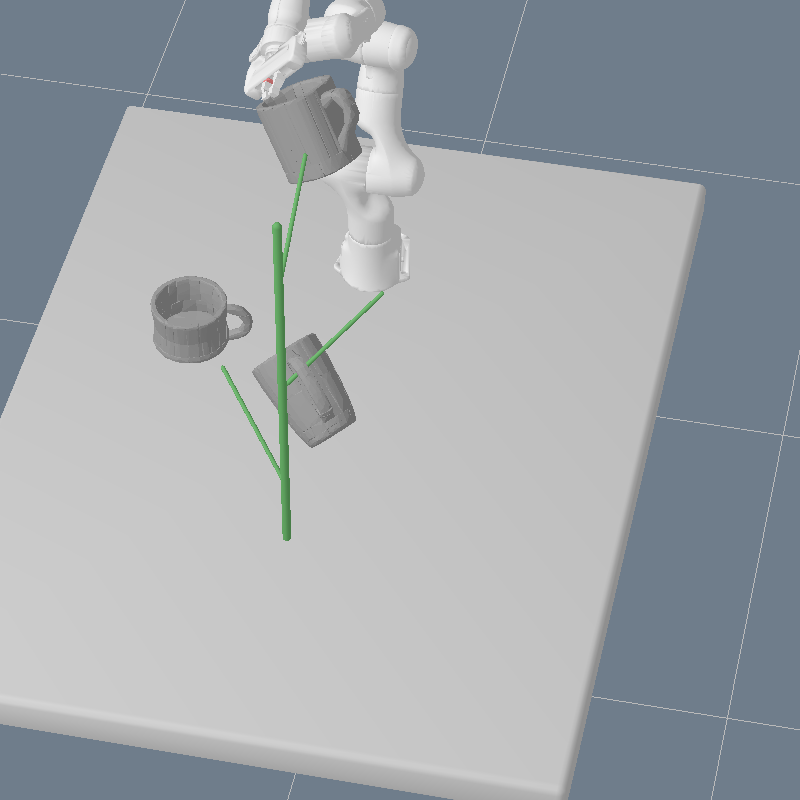}}
	\subfigure[t=20]{
		\includegraphics[width=.32\columnwidth]{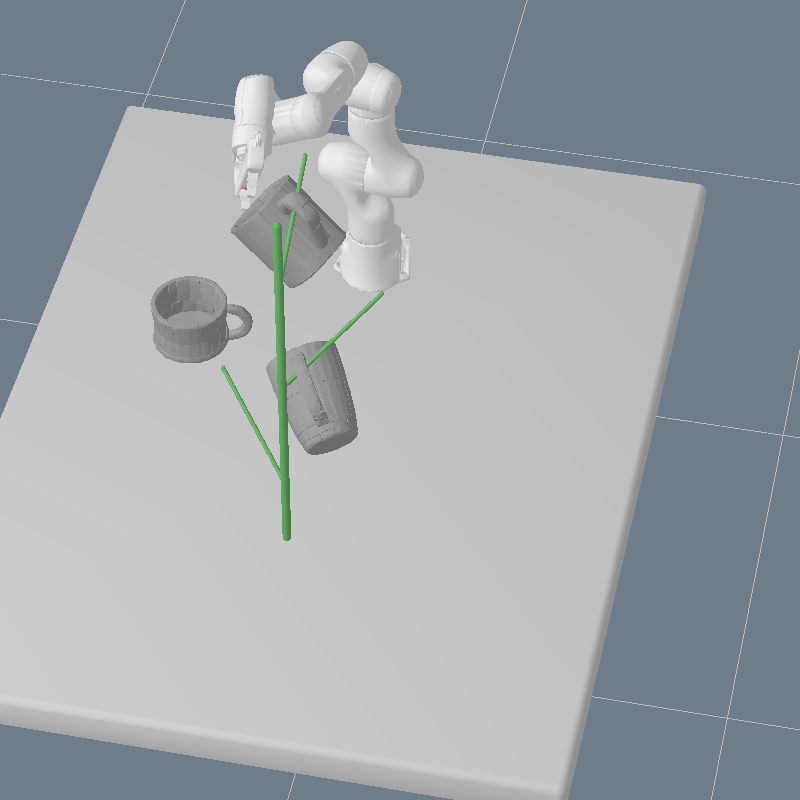}}
	\subfigure[t=25]{
		\includegraphics[width=.32\columnwidth]{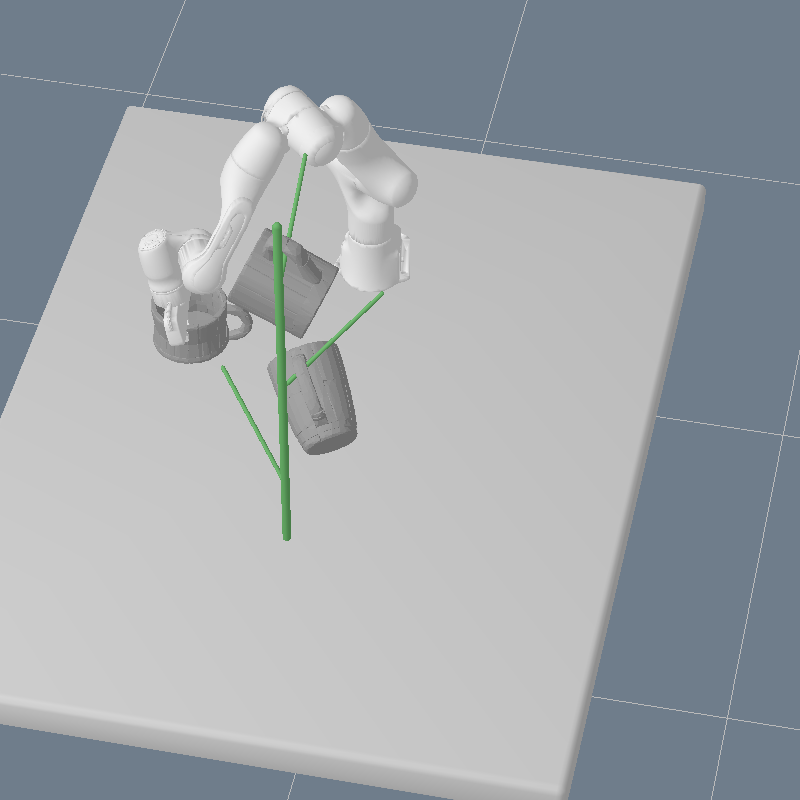}}
	\subfigure[t=28.5]{
		\includegraphics[width=.32\columnwidth]{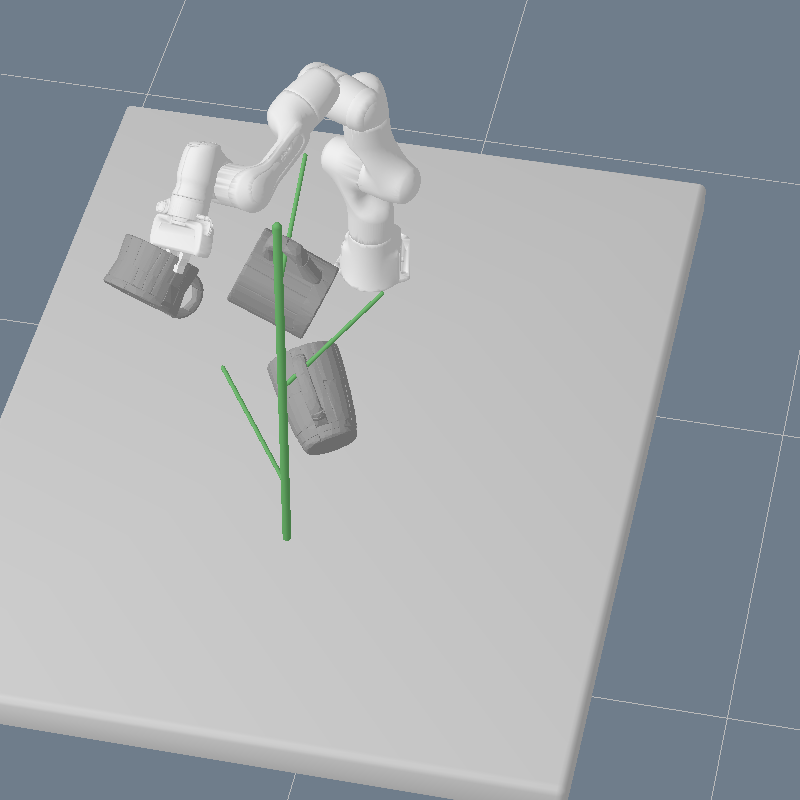}}
	\subfigure[t=30]{
		\includegraphics[width=.32\columnwidth]{multiHang10_2}}
	\caption{The three-mug scenario. 60 steps of robot configurations and rigid transformations of three mugs are jointly optimized via the proposed manipulation framework. This optimization is a 1071-dimensional decision problem (one 7DOF arm for 60 steps and one 7DOF mug for 51, 31, 11 steps = 1071, the mug's rigid transformations before grasped are not included in optimization) and is solved within 1 minute on a standard laptop.}\label{fig:threemug}
\end{figure}

\begin{figure}[t]
	\centering
	\subfigure[t=0]{
		\includegraphics[width=.32\columnwidth]{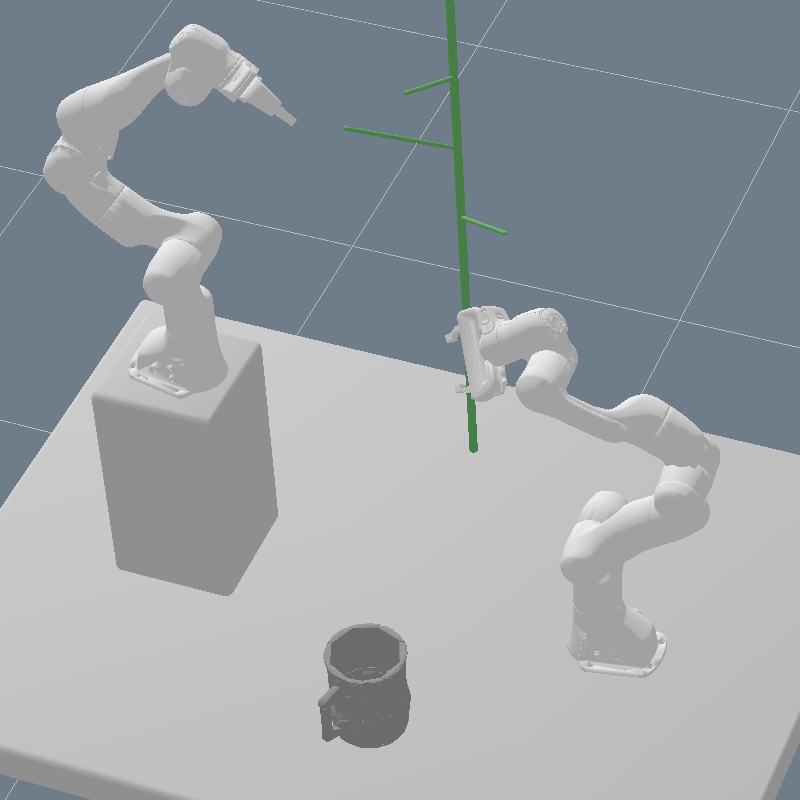}}
	\subfigure[t=5]{
		\includegraphics[width=.32\columnwidth]{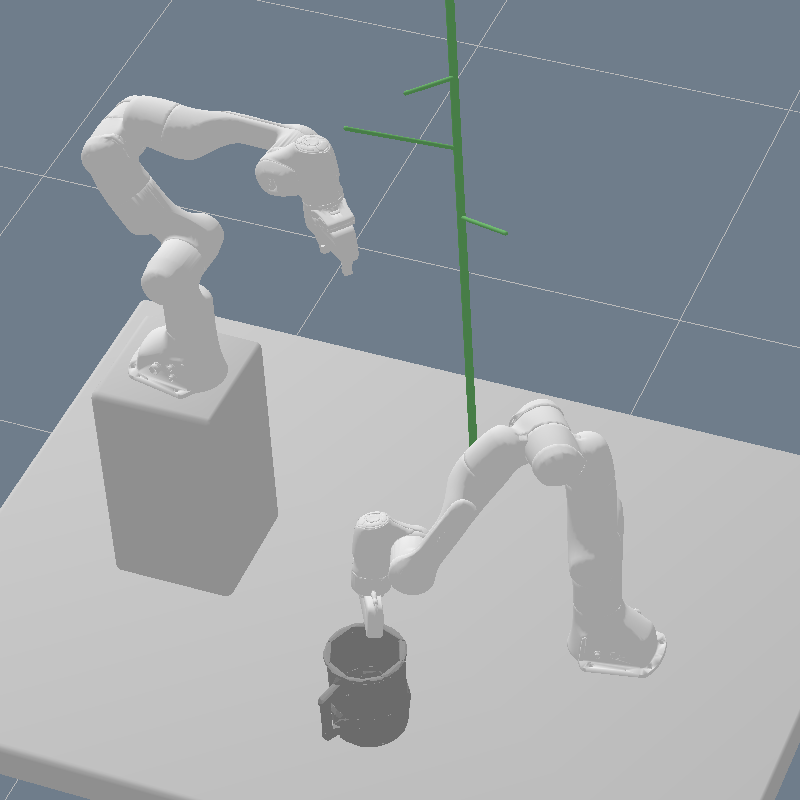}}
	\subfigure[t=10]{
		\includegraphics[width=.32\columnwidth]{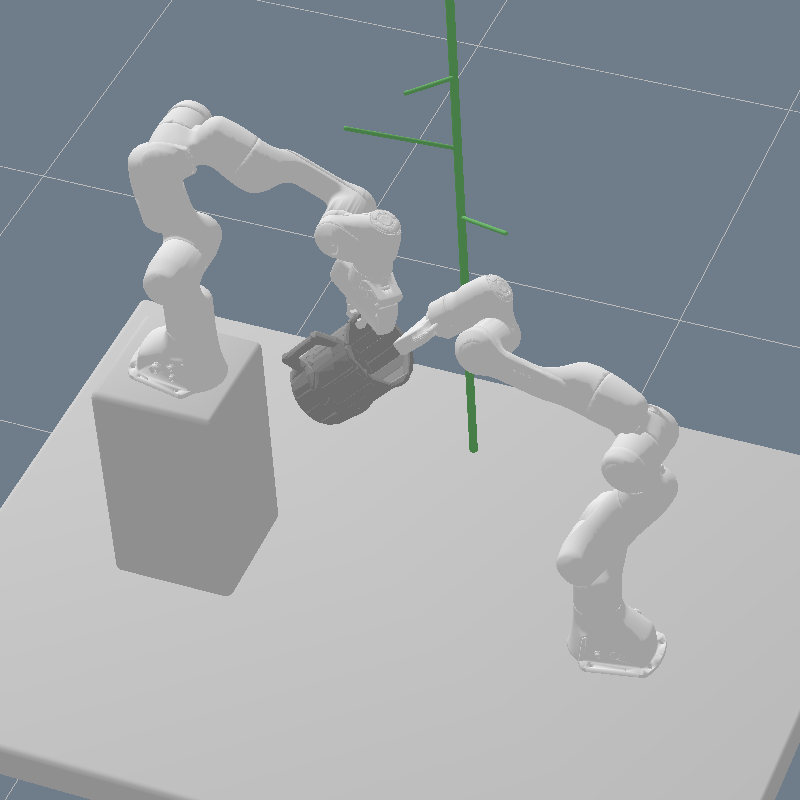}}
	\subfigure[t=13.5]{
		\includegraphics[width=.32\columnwidth]{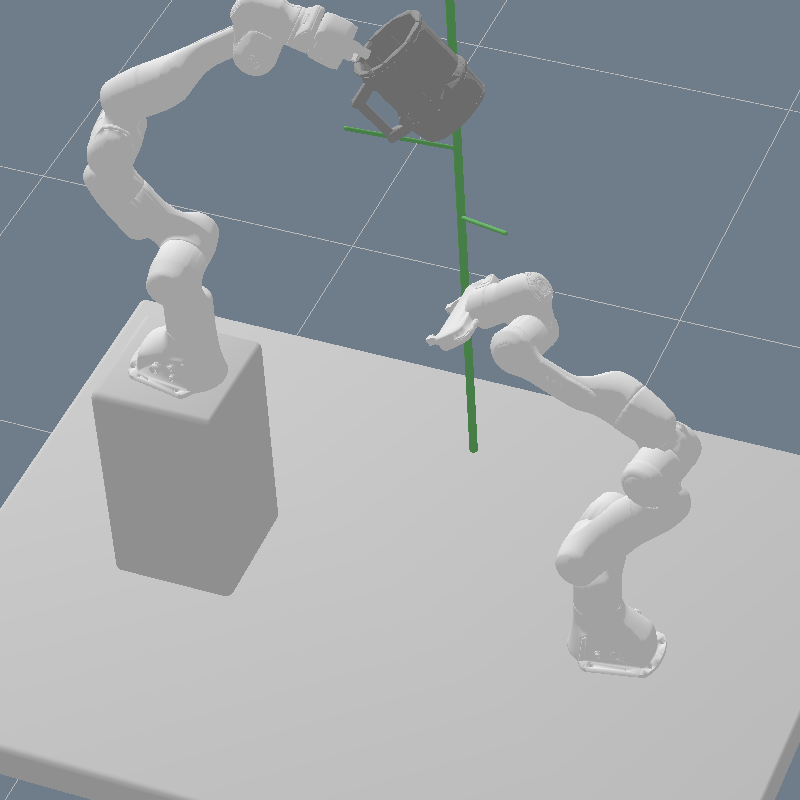}}
	\subfigure[t=15]{
		\includegraphics[width=.32\columnwidth]{handover5}}
	\caption{The handover scenario. 30 steps of the two arms' configurations and rigid transformations of the mug are jointly optimize dvia the proposed manipulation framework. This optimization is a 567-dimensional decision problem (two 7DOF arms for 30 steps and one 7DOF mug for 21 steps = 567, the mug's rigid transformations at the first phase are not included in optimization) and is solved within 1 minute on a standard laptop.}\label{fig:handover}
\end{figure}

To showcase the long-horizon planning capability of LGP, we considered the following two scenarios:
(i) The three-mug scenario consists of 6 discrete phases with [(\textsc{Grasp}, gripper, mug1), (\textsc{Hang}, M\_hook, mug1), (\textsc{Grasp}, gripper, mug2), (\textsc{Hang}, U\_hook, mug2), (\textsc{Grasp}, gripper, mug3), (\textsc{Hang}, L\_hook, mug3)].
(ii) The handover scenario has two arms at different heights and the target hook is placed high, requiring two arms to coordinate a handover motion with the discrete actions [(\textsc{Grasp}, R\_gripper, mug), (\textsc{Grasp}, L\_gripper, mug), (\textsc{Hang}, U\_hook, mug)].
Fig.~\ref{fig:long_horizon} shows the last configurations of the optimized plans; we refer readers to Figs.~\ref{fig:threemug}--\ref{fig:handover} and videos \href{https://youtube.com/playlist?list=PL9pnj8nG83OfNsYhRQ9gzhceT1PT8IXkq}{[video2]} for clearer views. 

\begin{figure}[t]
	\centering
	\subfigure[Sampled grasp]{
		\includegraphics[width=.31\columnwidth, viewport=0 0 800 780, clip]{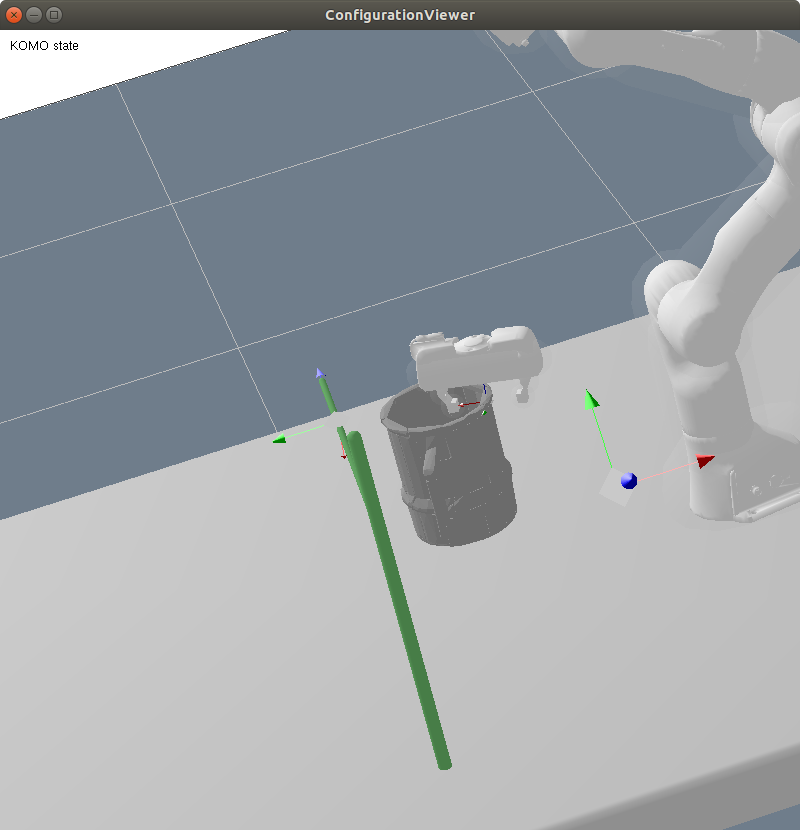}}
	\subfigure[Sampled hang]{
		\includegraphics[width=.31\columnwidth, viewport=0 0 800 780, clip]{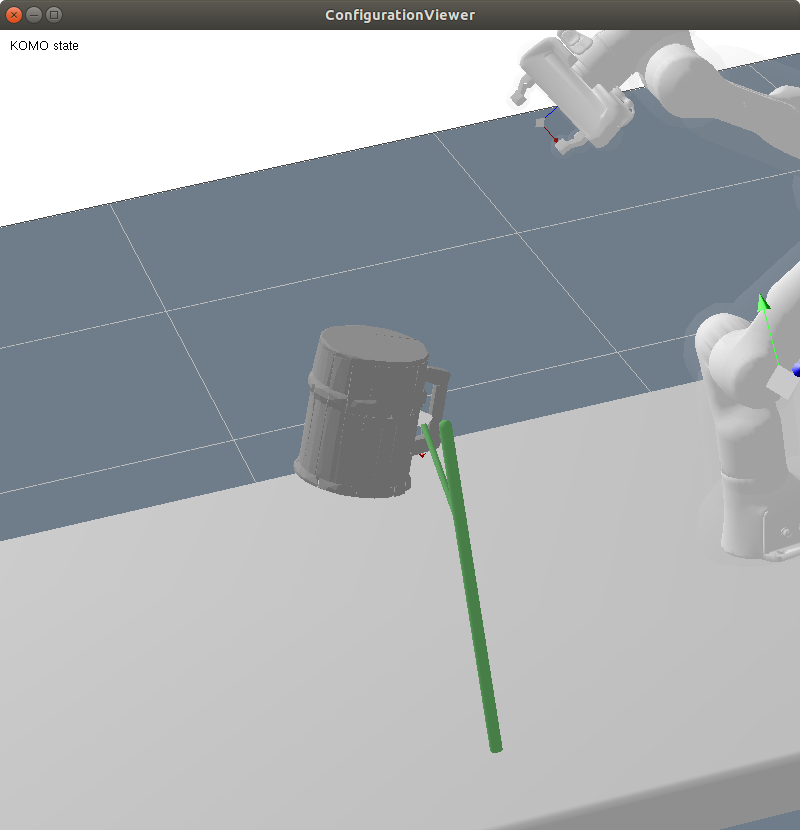}}
	\subfigure[IK (out-of-reach)]{
		\includegraphics[width=.31\columnwidth, viewport=0 0 800 780, clip]{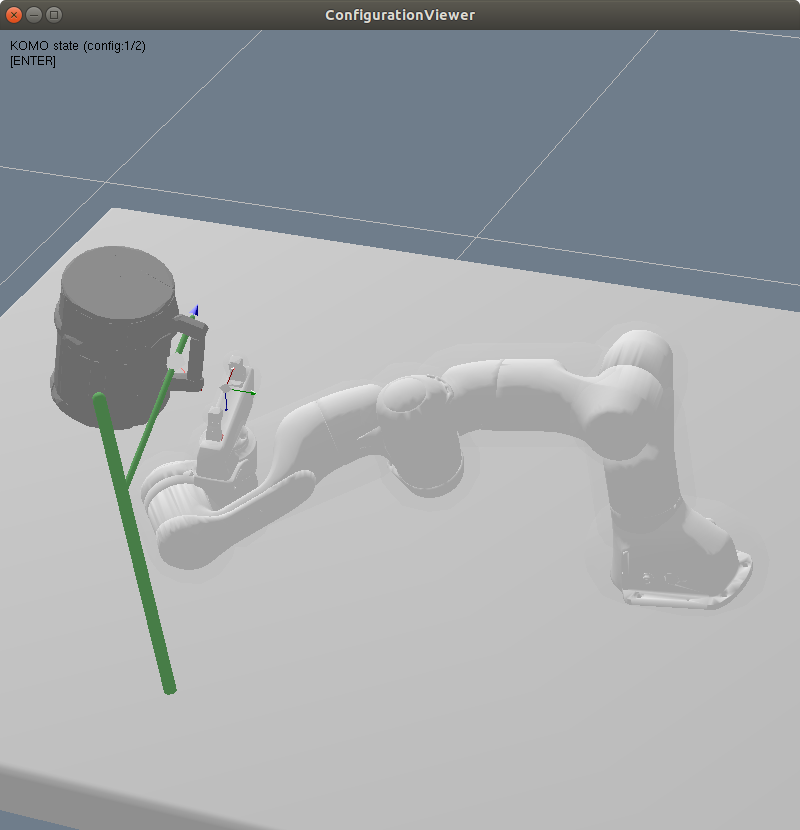}}\\
	\subfigure[Sampled grasp1]{
		\includegraphics[width=.31\columnwidth, viewport=0 0 800 780, clip]{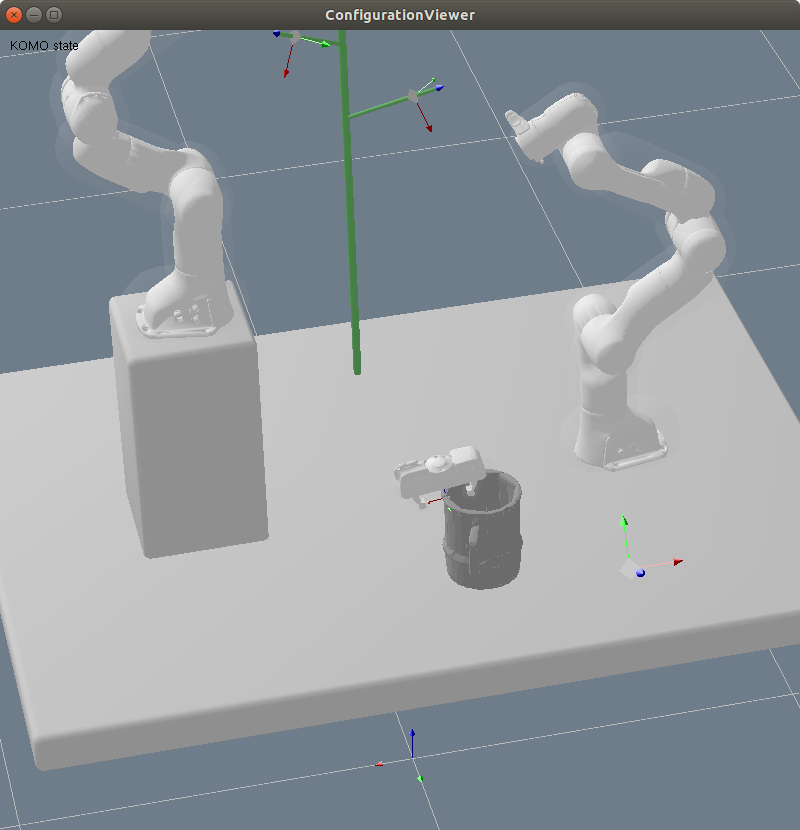}}
	\subfigure[Sampled grasp2]{
		\includegraphics[width=.31\columnwidth, viewport=0 0 800 780, clip]{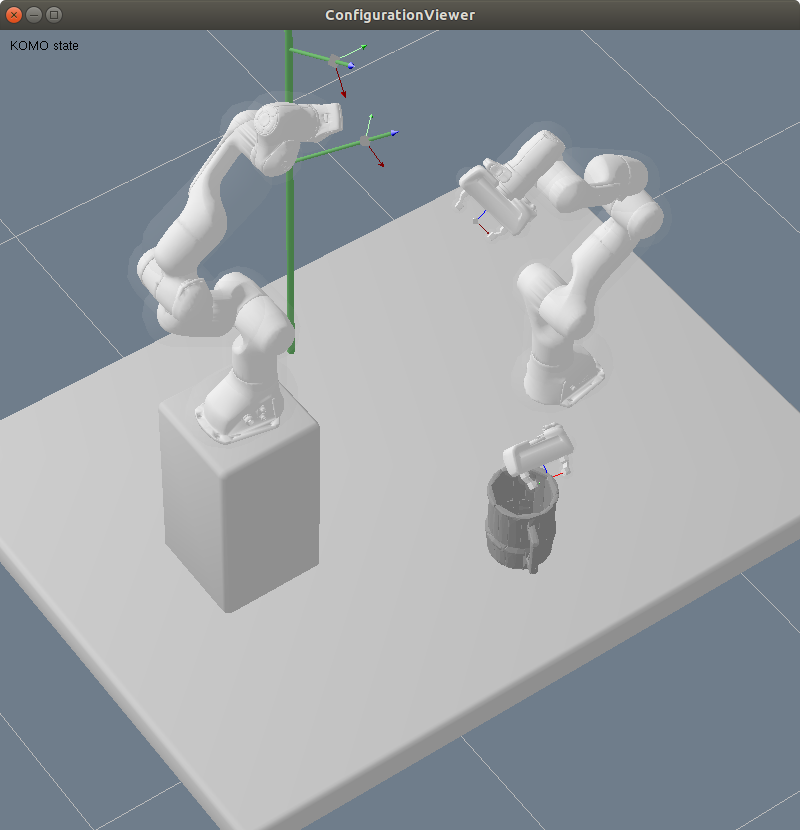}}
	\subfigure[IK (collision)]{
		\includegraphics[width=.31\columnwidth, viewport=0 0 800 780, clip]{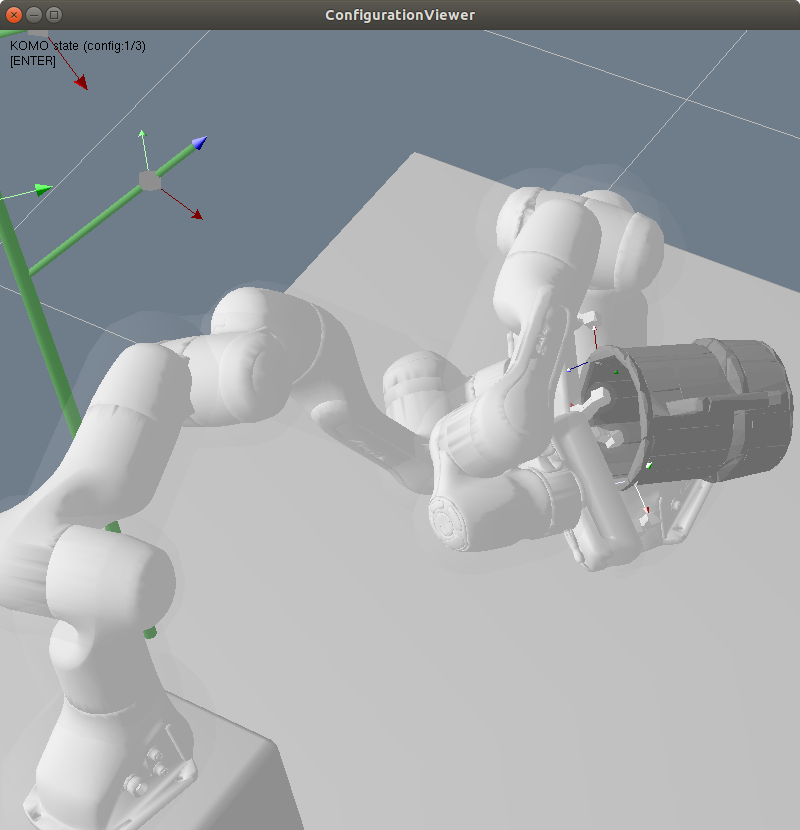}}
	\caption{Inverse kinematics with generative models}\label{fig:gen}
\end{figure}

\begin{figure}[t]
	\centering
	\subfigure[Sampled grasp pose]{
		\includegraphics[width=.32\columnwidth, viewport=0 0 800 780, clip]{genModels/0}}
	\subfigure[Sampled hang pose]{
		\includegraphics[width=.32\columnwidth, viewport=0 0 800 780, clip]{genModels/1}}
	\subfigure[IK result (infiseable)]{
		\includegraphics[width=.32\columnwidth, viewport=0 0 800 780, clip]{genModels/2}}
	\subfigure[Sampled grasp pose]{
		\includegraphics[width=.32\columnwidth, viewport=0 0 800 780, clip]{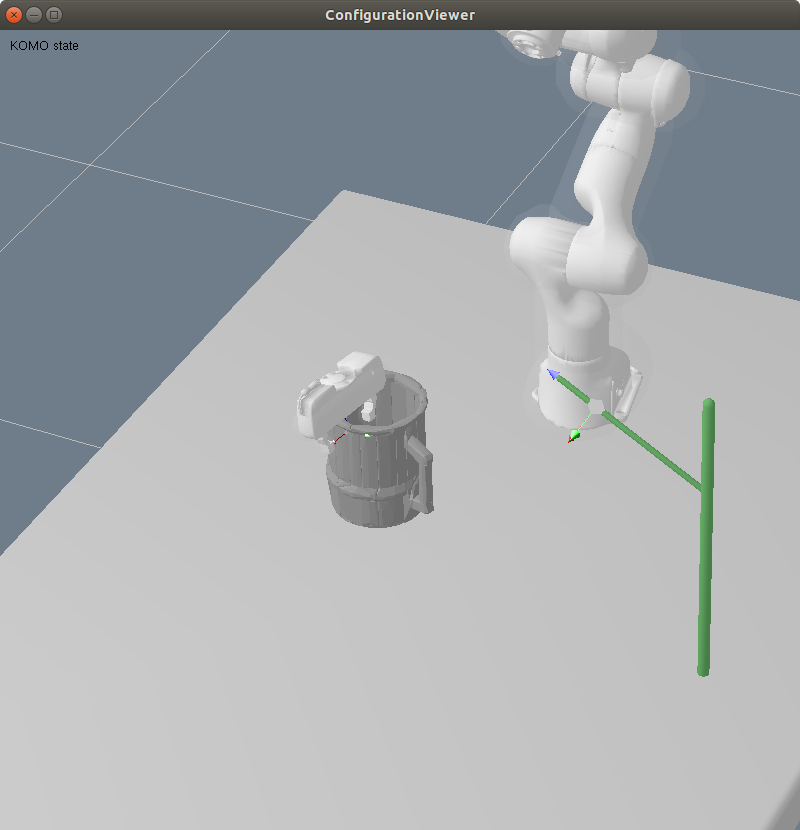}}
	\subfigure[Sampled hang pose]{
		\includegraphics[width=.32\columnwidth, viewport=0 0 800 780, clip]{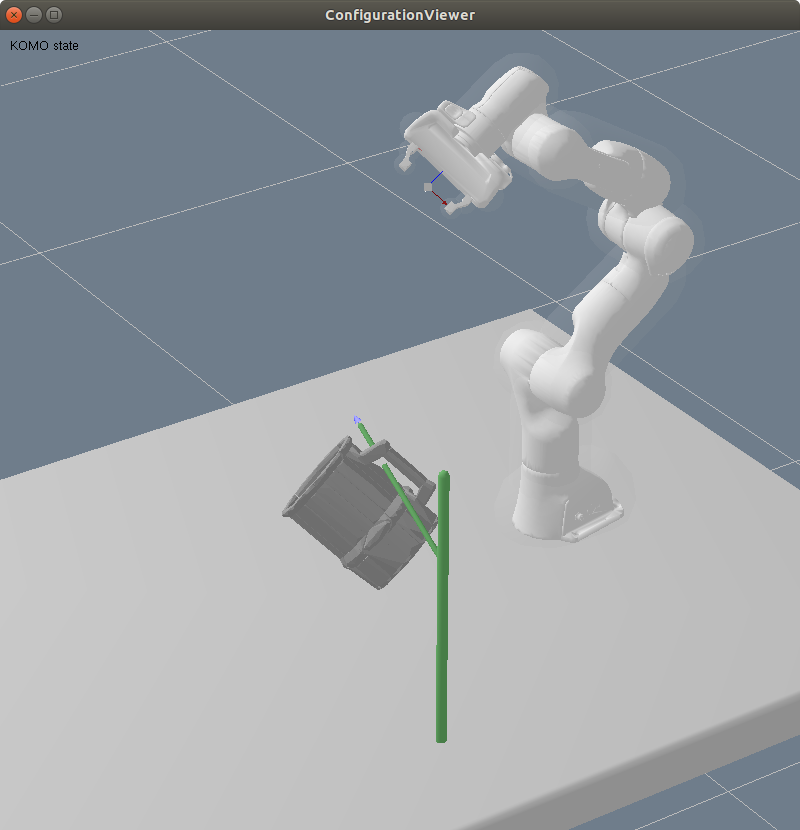}}
	\subfigure[IK result (infiseable)]{
		\includegraphics[width=.32\columnwidth, viewport=0 0 800 780, clip]{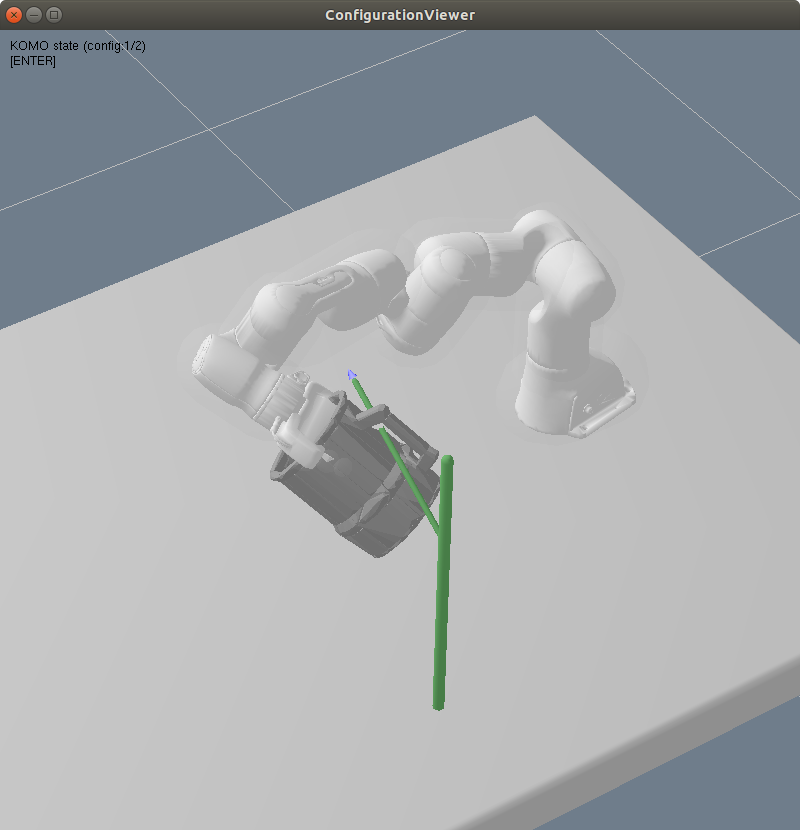}}
	\subfigure[Sampled grasp pose]{
		\includegraphics[width=.32\columnwidth, viewport=0 0 800 780, clip]{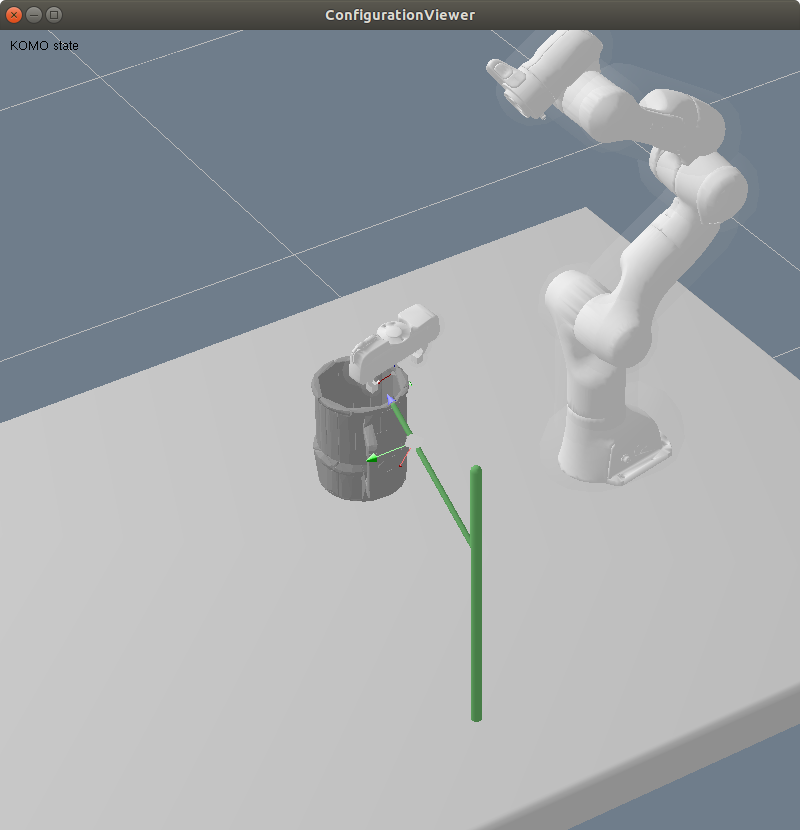}}
	\subfigure[Sampled hang pose]{
		\includegraphics[width=.32\columnwidth, viewport=0 0 800 780, clip]{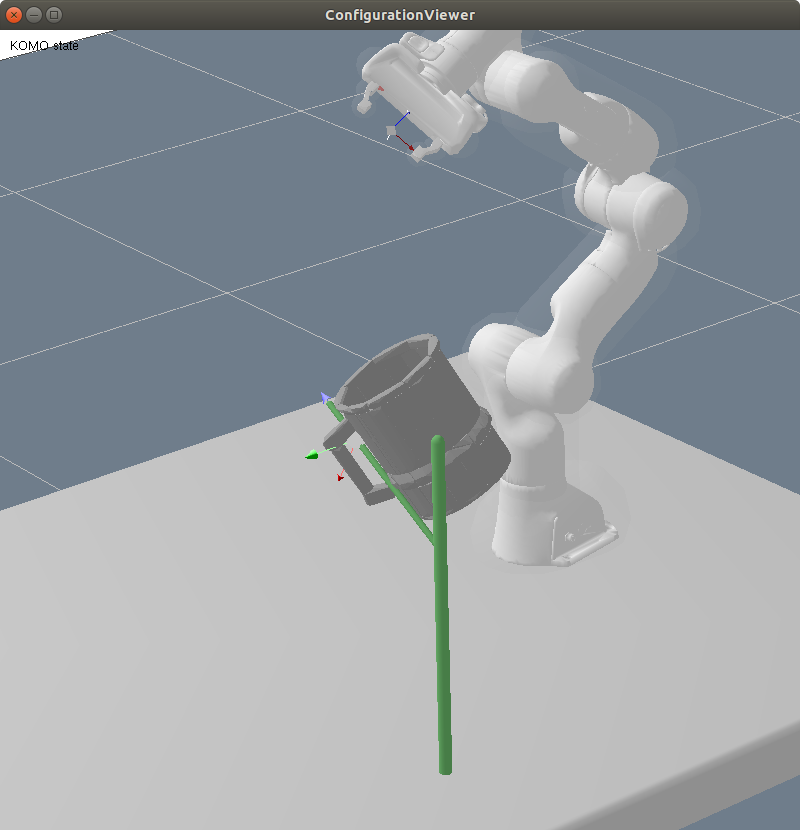}}
	\subfigure[IK result (infiseable)]{
		\includegraphics[width=.32\columnwidth, viewport=0 0 800 780, clip]{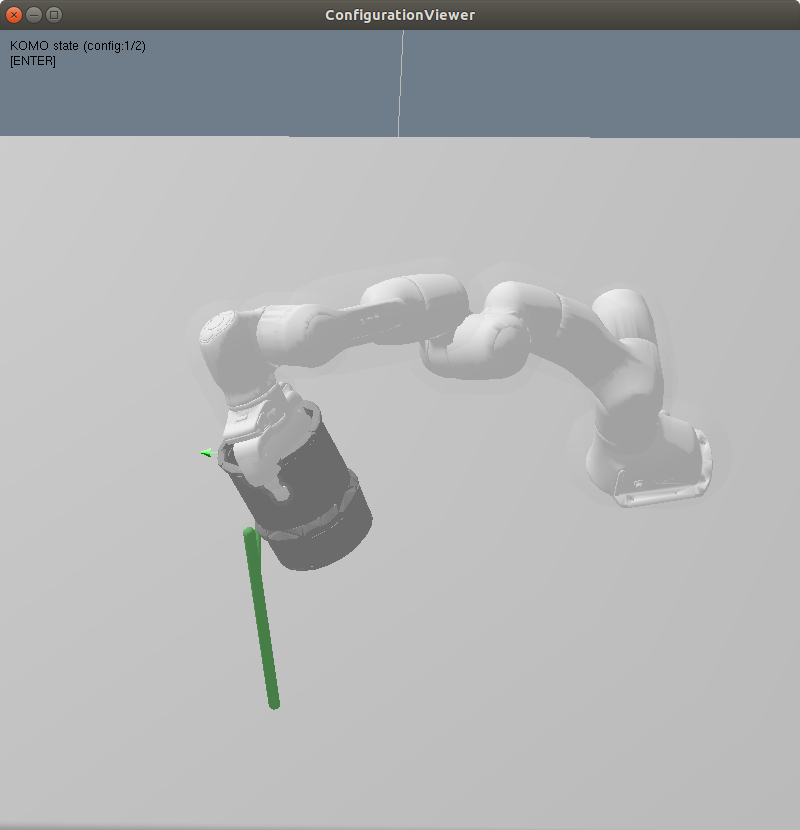}}
	\caption{IK with generative models - Pick \& Hang. Separately generated poses often can not be coordinated due to the kinematic infeasibility, i.e., the robot joint angle limits, or the collision constraints.}\label{fig:gen1}
\end{figure}

\begin{figure}[t]
	\centering
	\subfigure[Sampled grasp1 pose]{
		\includegraphics[width=.32\columnwidth, viewport=0 0 800 780, clip]{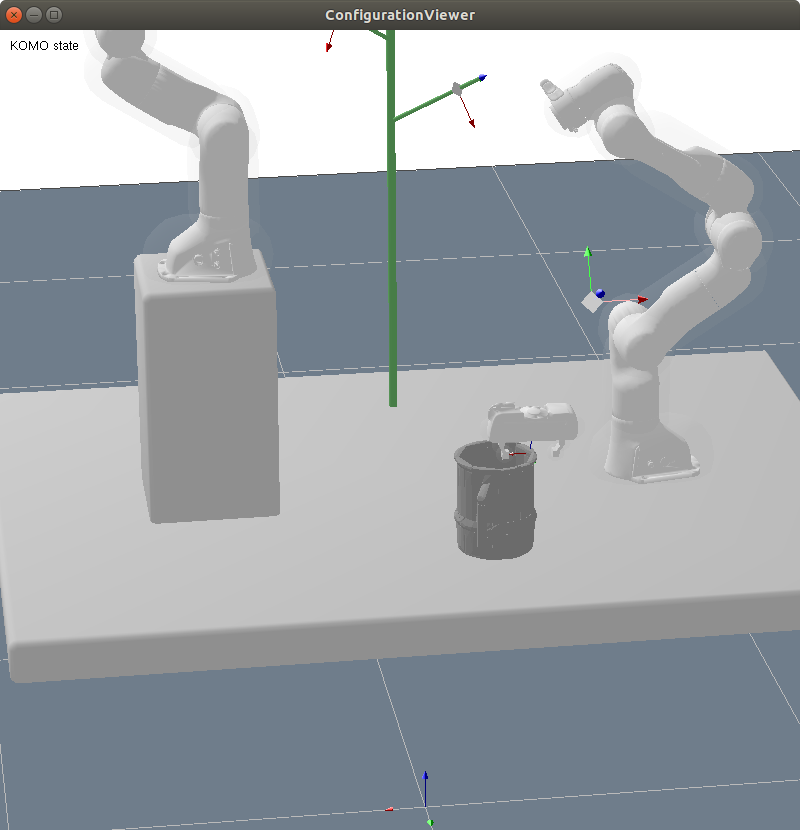}}
	\subfigure[Sampled grasp2 pose]{
		\includegraphics[width=.32\columnwidth, viewport=0 0 800 780, clip]{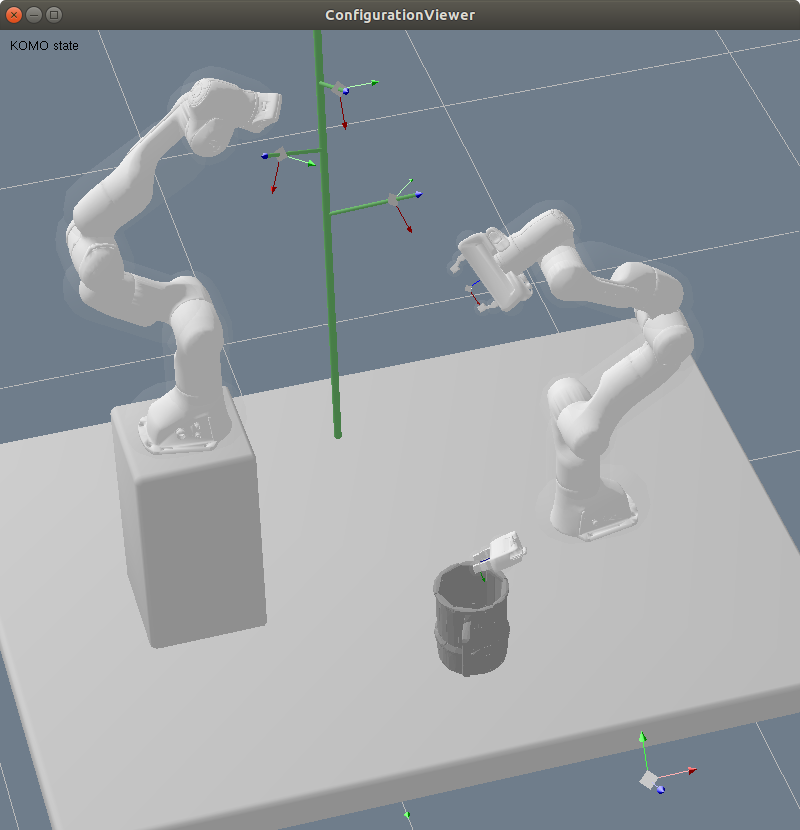}}
	\subfigure[IK result (infiseable)]{
		\includegraphics[width=.32\columnwidth, viewport=0 0 800 780, clip]{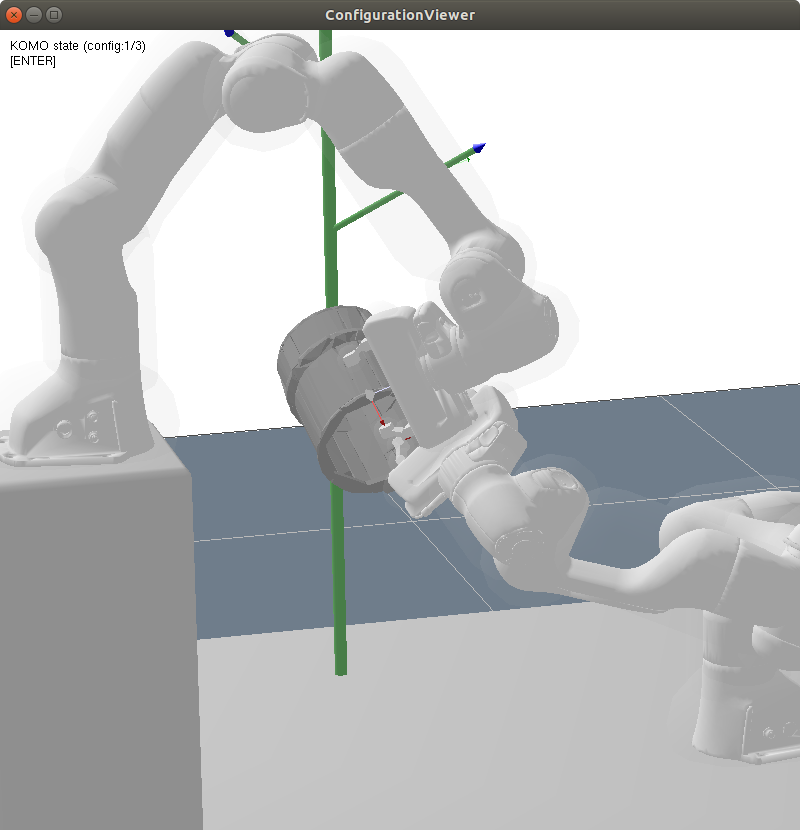}}
	\subfigure[Sampled grasp1 pose]{
		\includegraphics[width=.32\columnwidth, viewport=0 0 800 780, clip]{genModels/26}}
	\subfigure[Sampled grasp2 pose]{
		\includegraphics[width=.32\columnwidth, viewport=0 0 800 780, clip]{genModels/27}}
	\subfigure[IK result (infiseable)]{
		\includegraphics[width=.32\columnwidth, viewport=0 0 800 780, clip]{genModels/28}}
	\subfigure[Sampled grasp1 pose]{
		\includegraphics[width=.32\columnwidth, viewport=0 0 800 780, clip]{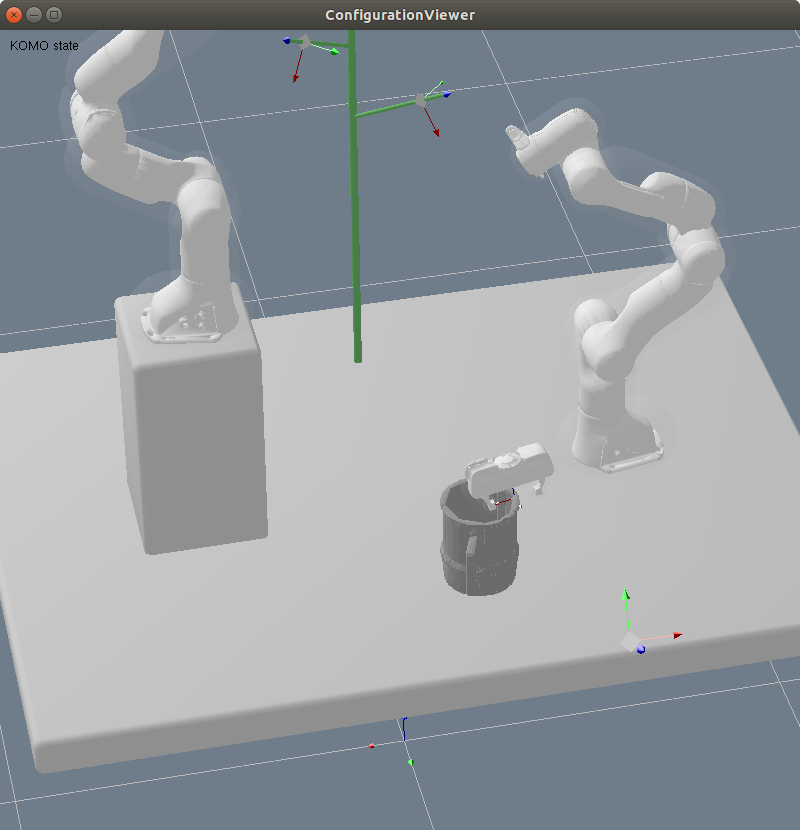}}
	\subfigure[Sampled grasp2 pose]{
		\includegraphics[width=.32\columnwidth, viewport=0 0 800 780, clip]{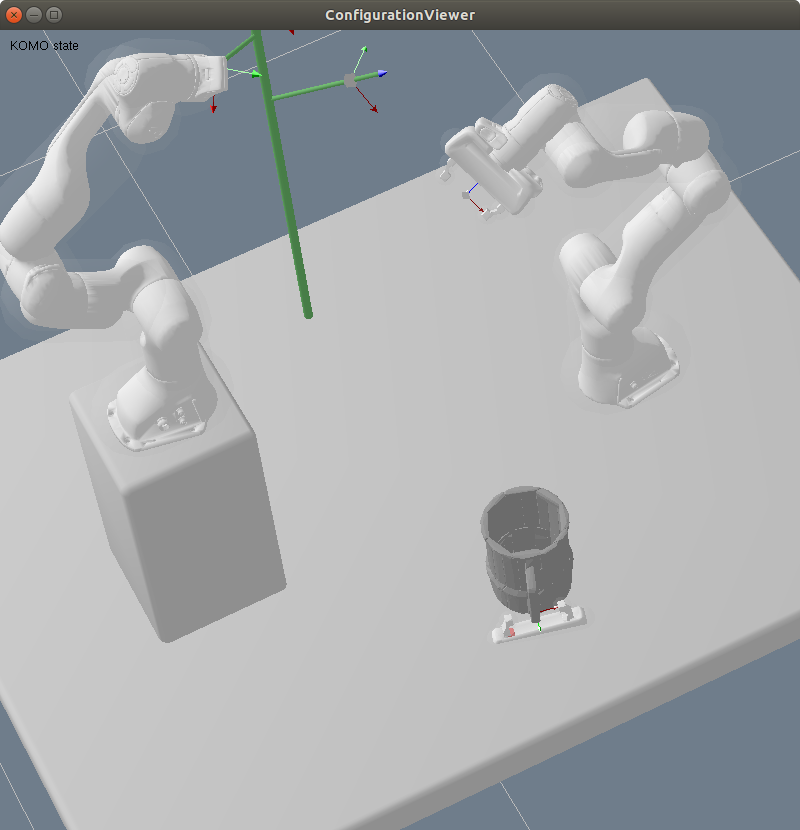}}
	\subfigure[IK result (infiseable)]{
		\includegraphics[width=.32\columnwidth, viewport=0 0 800 780, clip]{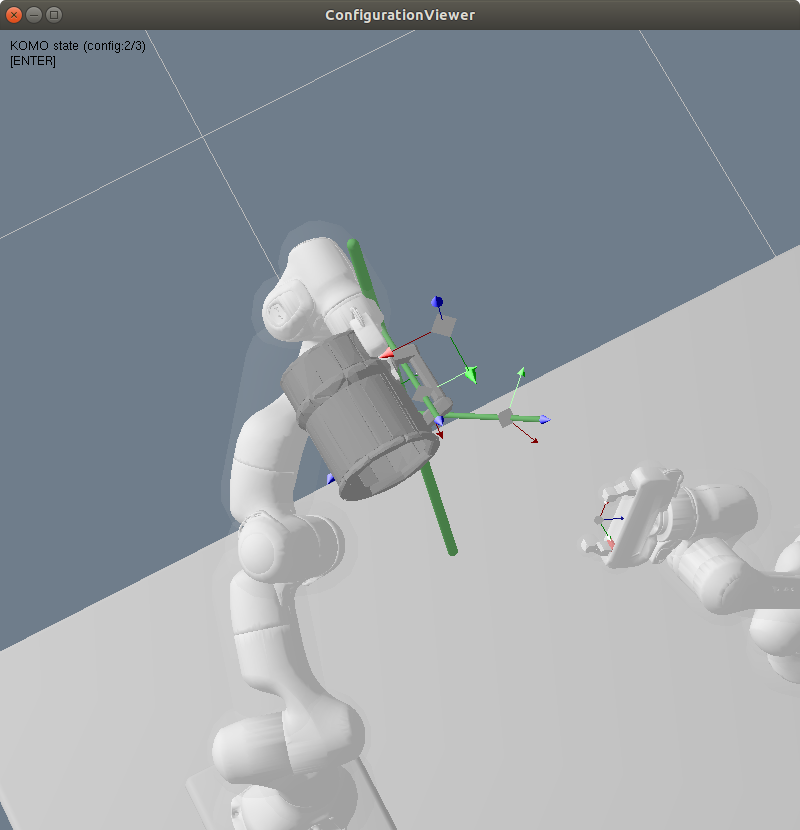}}
	\caption{IK with generative models - Handover. Separately generated poses often can not be coordinated due to the kinematic infeasibility, i.e., the robot joint angle limits, or the collision constraints.}\label{fig:gen2}
\end{figure}

{\bf Inverse Kinematics with Generative Models:} One important attribute of our framework is that, while most existing works train generative models that directly produce the interaction poses, ours models interactions as equality constraints where multiple constraints can be \textit{jointly} optimized with other planning features.
To see the benefits of such joint optimization, we considered the following inverse kinematics problems with a generative model:
For the basic pick \& hang and handover scenarios, we optimized each interaction pose separately as in Sec.~\ref{sec:learned feature} and checked if these individually optimized poses are kinematically feasible when combined together, i.e., whether or not the inverse kinematics problems have a solution.
Even though the mug's initial pose was given such that the first gasping is ensured feasible, 53 out of 100 pairs of grasp and hang poses were infeasible for the pick \& hang scenario and 86 out of 100 sets were infeasible for the handover scenario, i.e., many of the sampled poses led to a collision or an infeasible robot configuration for hanging/handover. Some failure cases are depicted in Fig.~\ref{fig:gen} (more in Figs.~\ref{fig:gen1}--\ref{fig:gen2}).
As the sequence length gets longer, not only should an exponentially larger number of planning problems be solved to find a set of feasible poses, but also the found poses are not guaranteed to be optimal.
The joint optimization with our constraint models doesn't raise such issues.

\subsection{Exploiting Learned Representations: 6D Pose Estimation and Zero-shot Imitation} \label{sec:poseEstim}
\begin{figure}[t]
	\centering
	\subfigure{
		\includegraphics[width=.31\columnwidth, viewport = 0 200 350 550, clip]{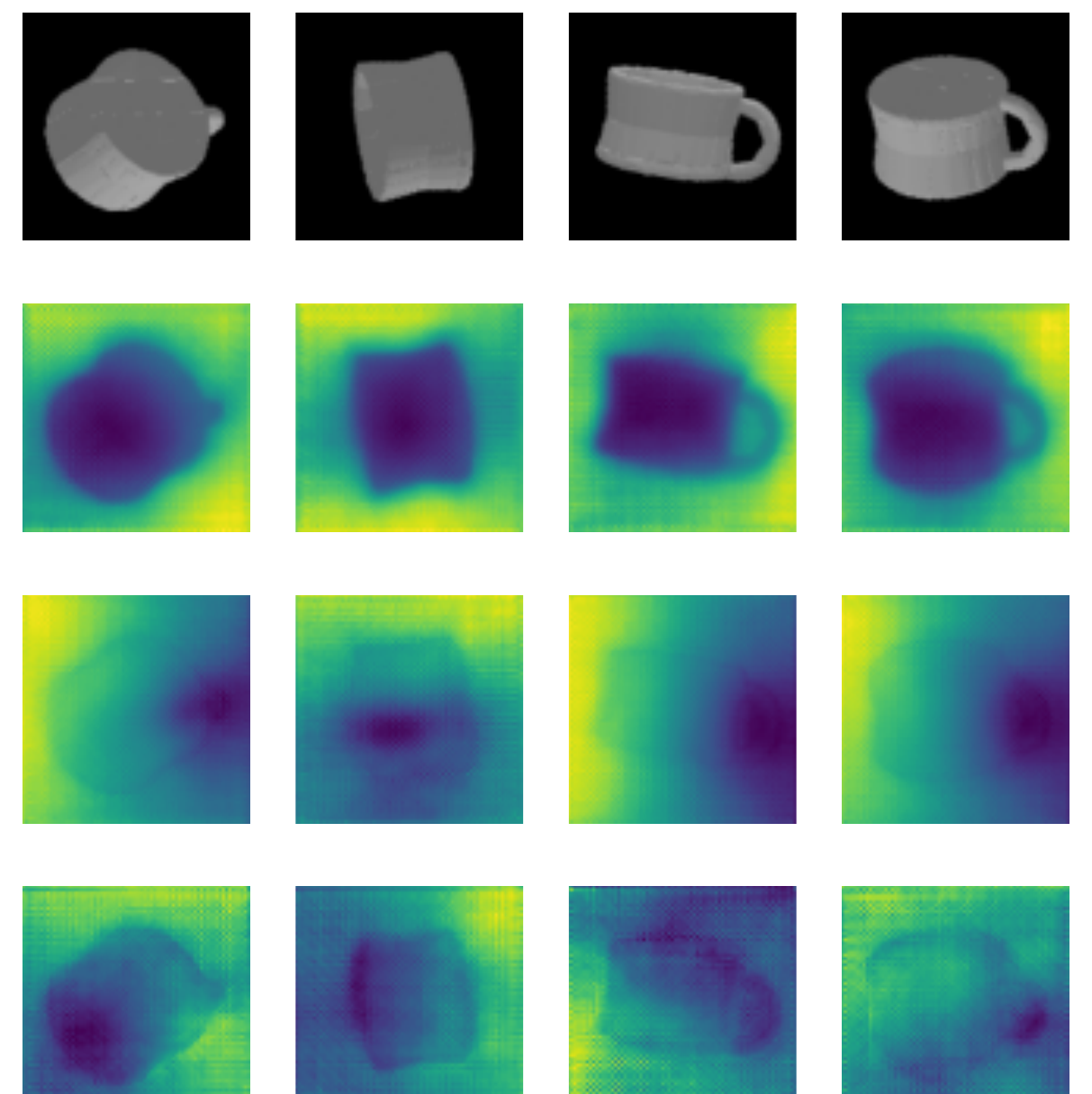}}
	\subfigure{
		\includegraphics[width=.31\columnwidth, viewport = 0 200 350 550, clip]{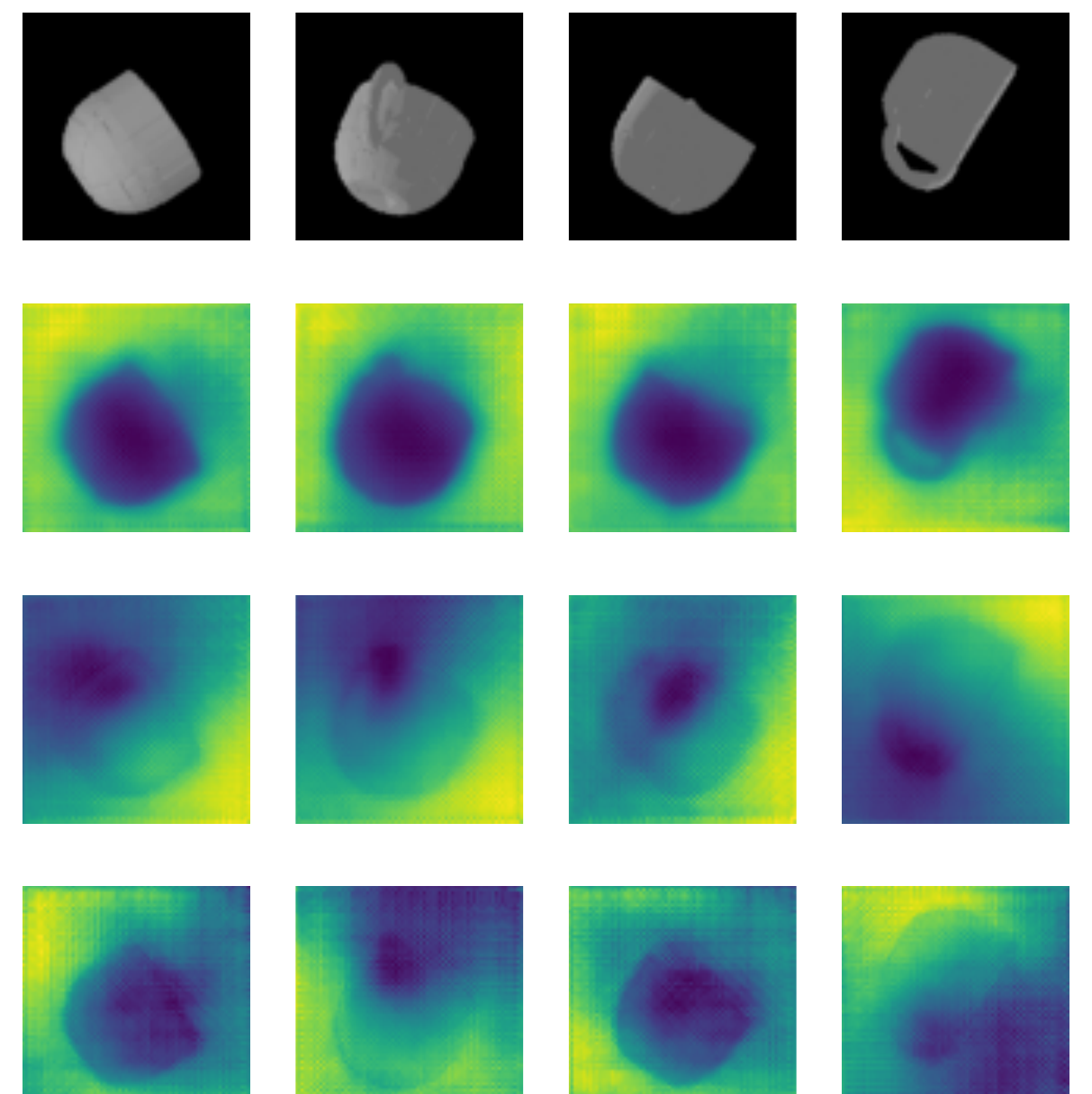}}
	\subfigure{
		\includegraphics[width=.31\columnwidth, viewport = 0 200 350 550, clip]{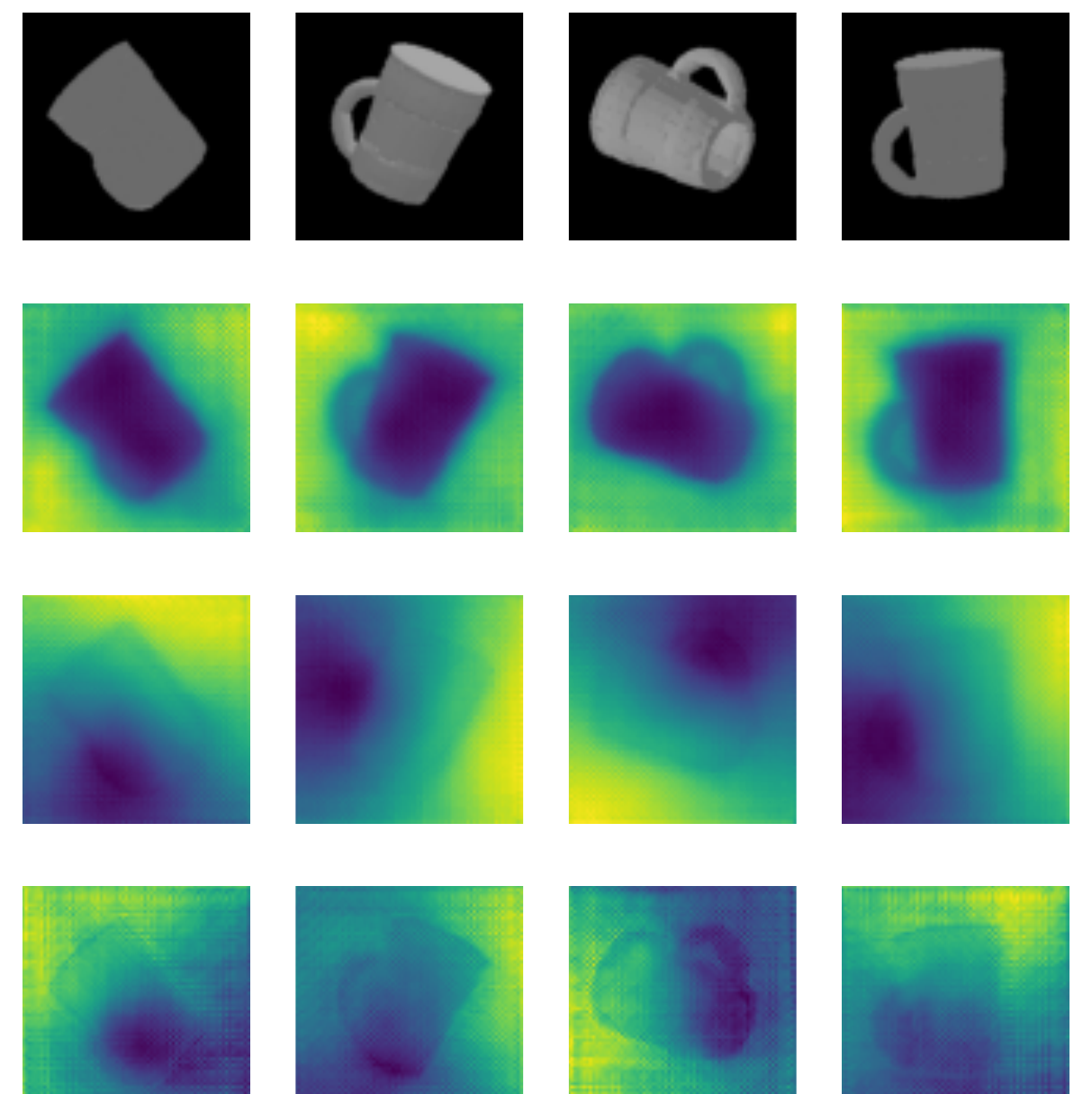}}
	\caption{First 3 principal components from PCA on image features. Each component distinguishes the overall object areas, the handle or rim parts, etc. More images can be found in Figs.~\ref{fig:PCA_img}--\ref{fig:PCA_3d}.
	}\label{fig:PCA_img0}
\end{figure}

\begin{figure}[t]
	\centering
	
	\subfigure[Model (right) and target (left) mugs]{
		\includegraphics[width=.3\columnwidth, viewport=0 0 800 780, clip]{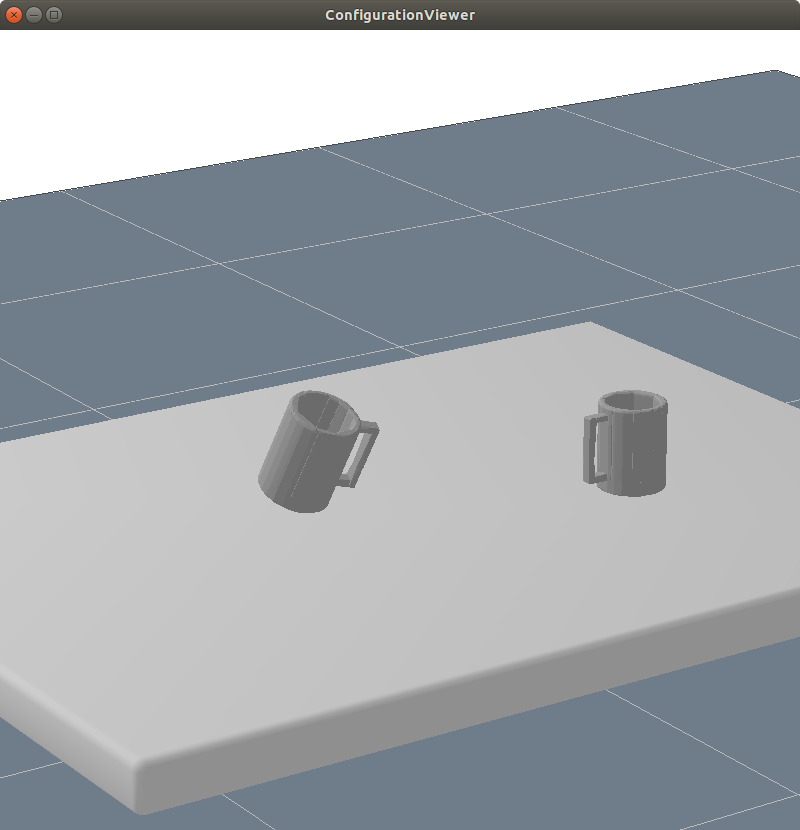}}
	\subfigure[Point clouds for ICP]{
		\includegraphics[width=.3\columnwidth, viewport=0 0 800 780, clip]{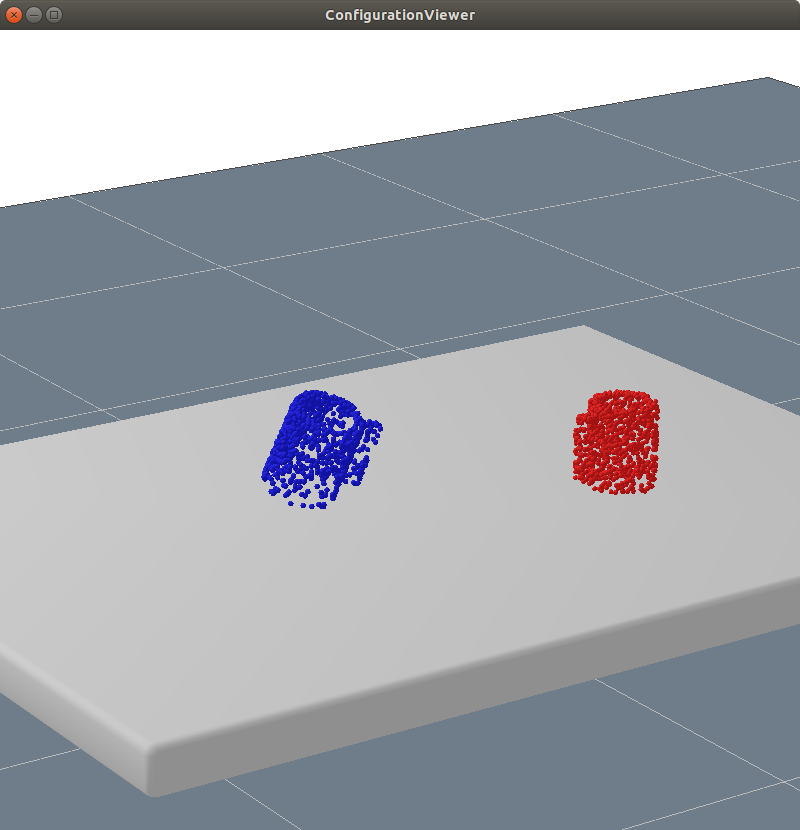}}
	\subfigure[Grid points for FCP]{
		\includegraphics[width=.3\columnwidth, viewport=0 0 800 780, clip]{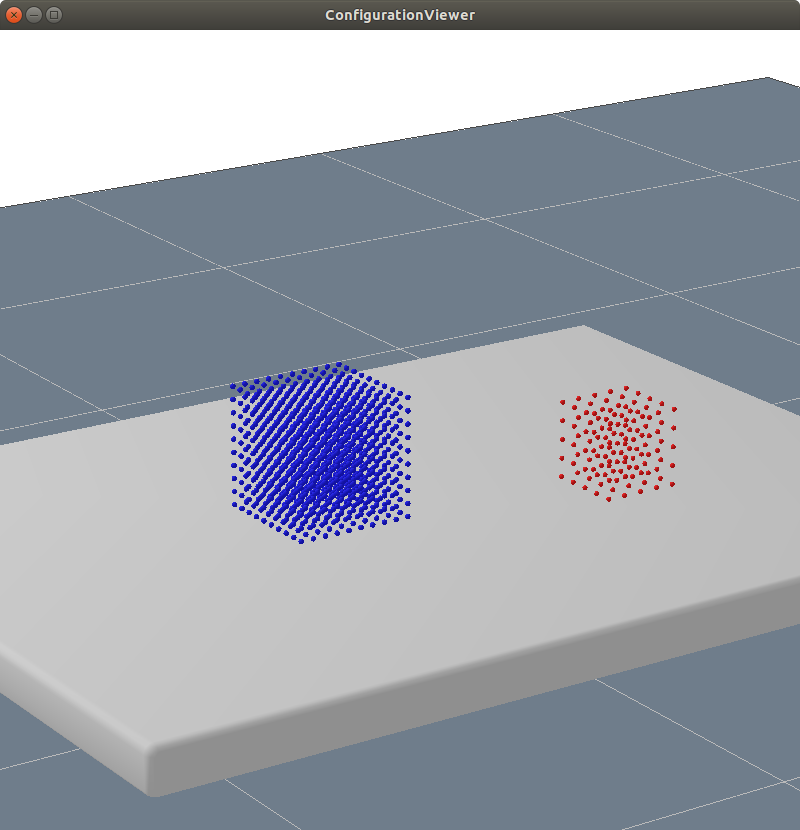}\label{fig:poseEstim_grid0}}
	\subfigure[ICP]{
		\includegraphics[width=.223\columnwidth, viewport=0 0 800 780, clip]{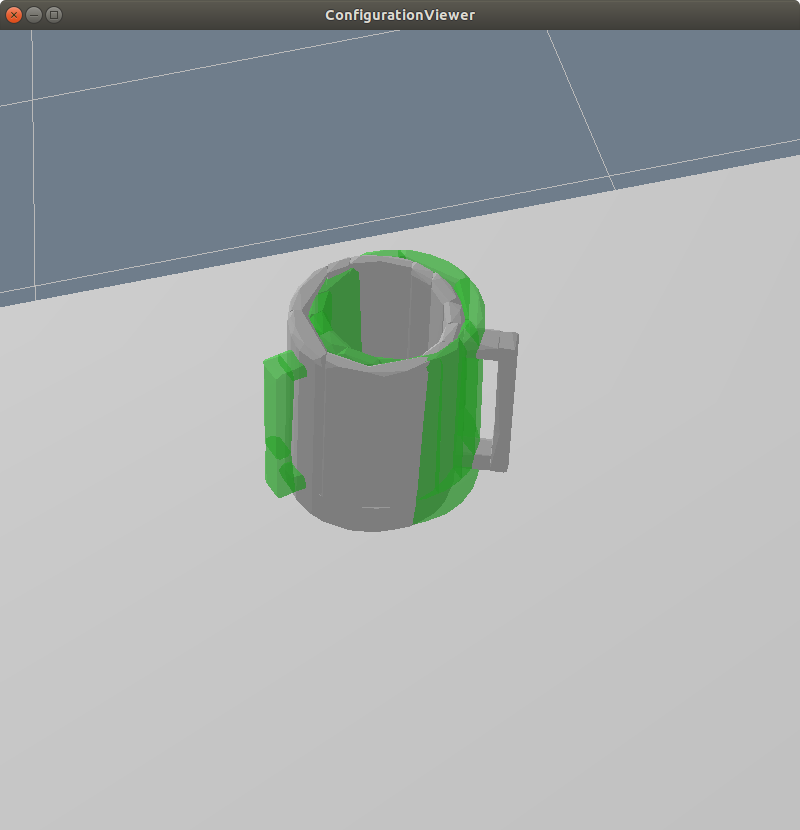}}
	\subfigure[ICP2]{
		\includegraphics[width=.223\columnwidth, viewport=0 0 800 780, clip]{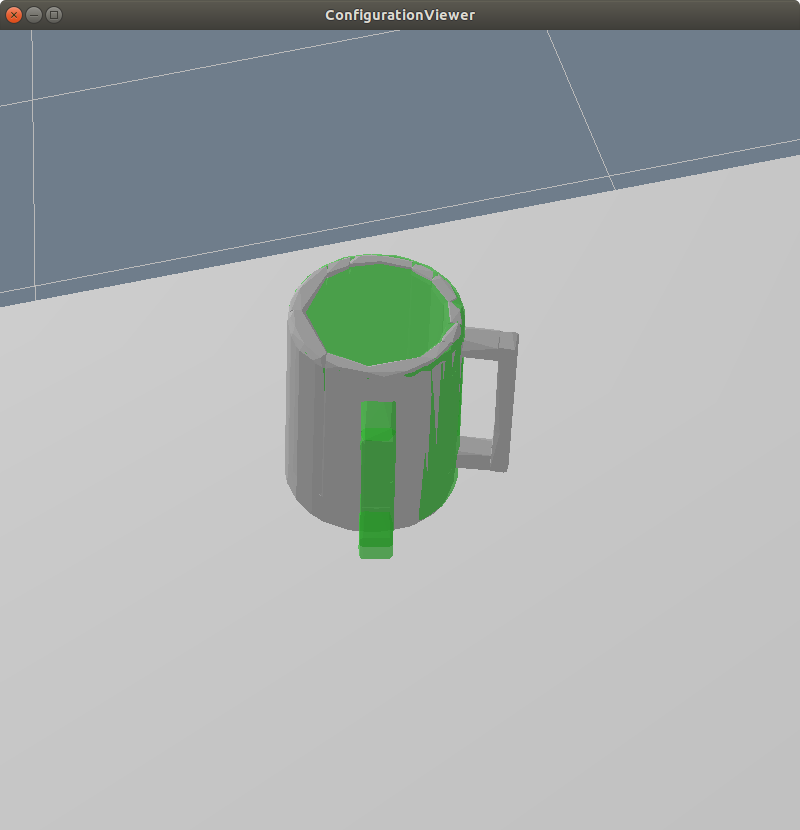}}
	\subfigure[FCP]{
		\includegraphics[width=.223\columnwidth, viewport=0 0 800 780, clip]{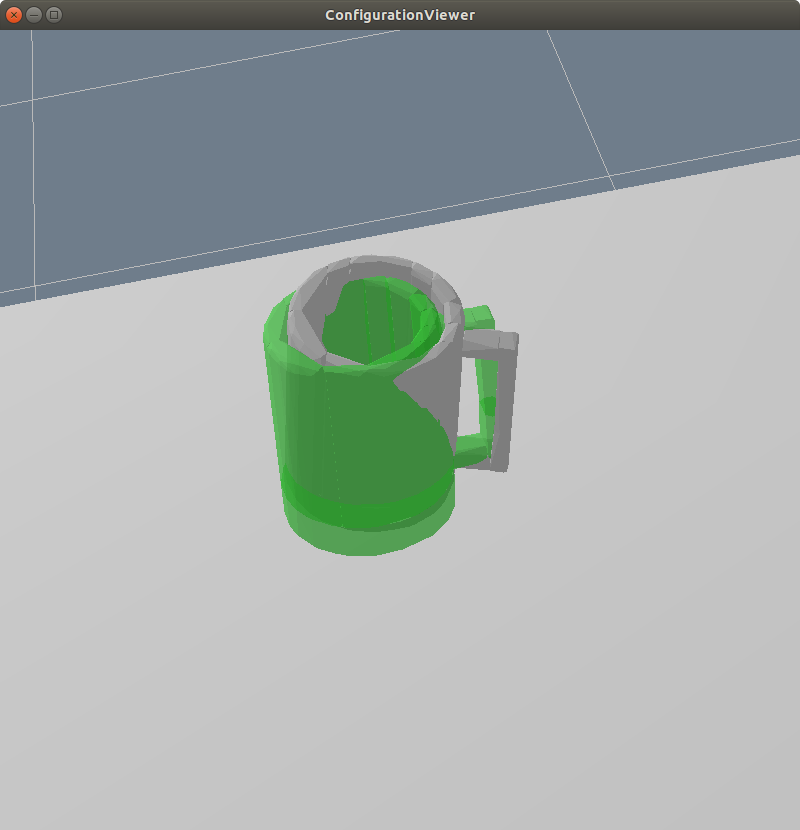}}
	\subfigure[F+ICP2]{
		\includegraphics[width=.223\columnwidth, viewport=0 0 800 780, clip]{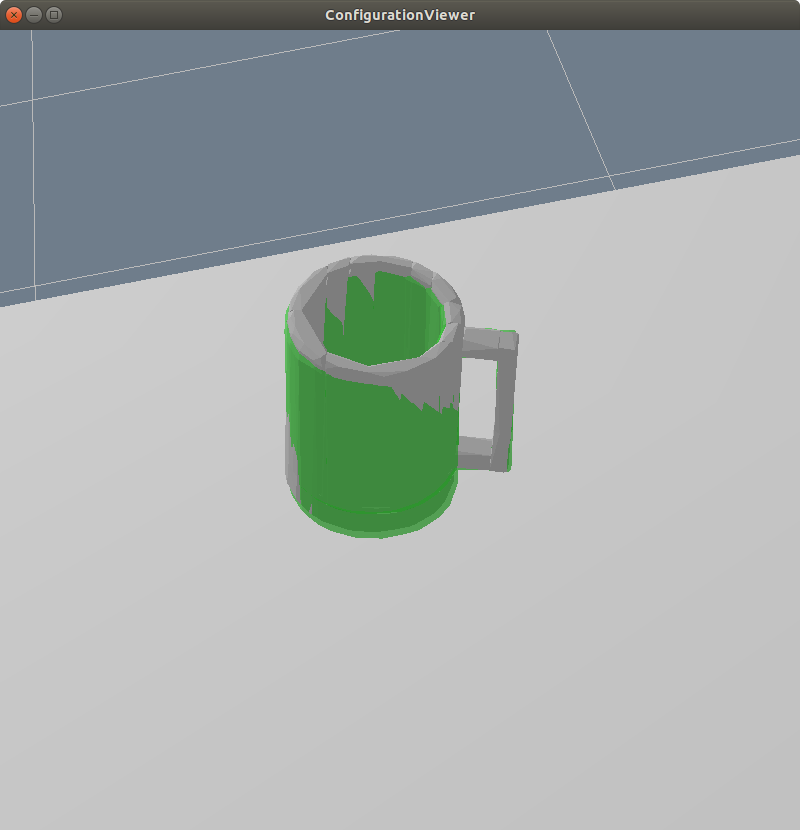}}
	\subfigure[Position error]{
		\includegraphics[width=.47\columnwidth]{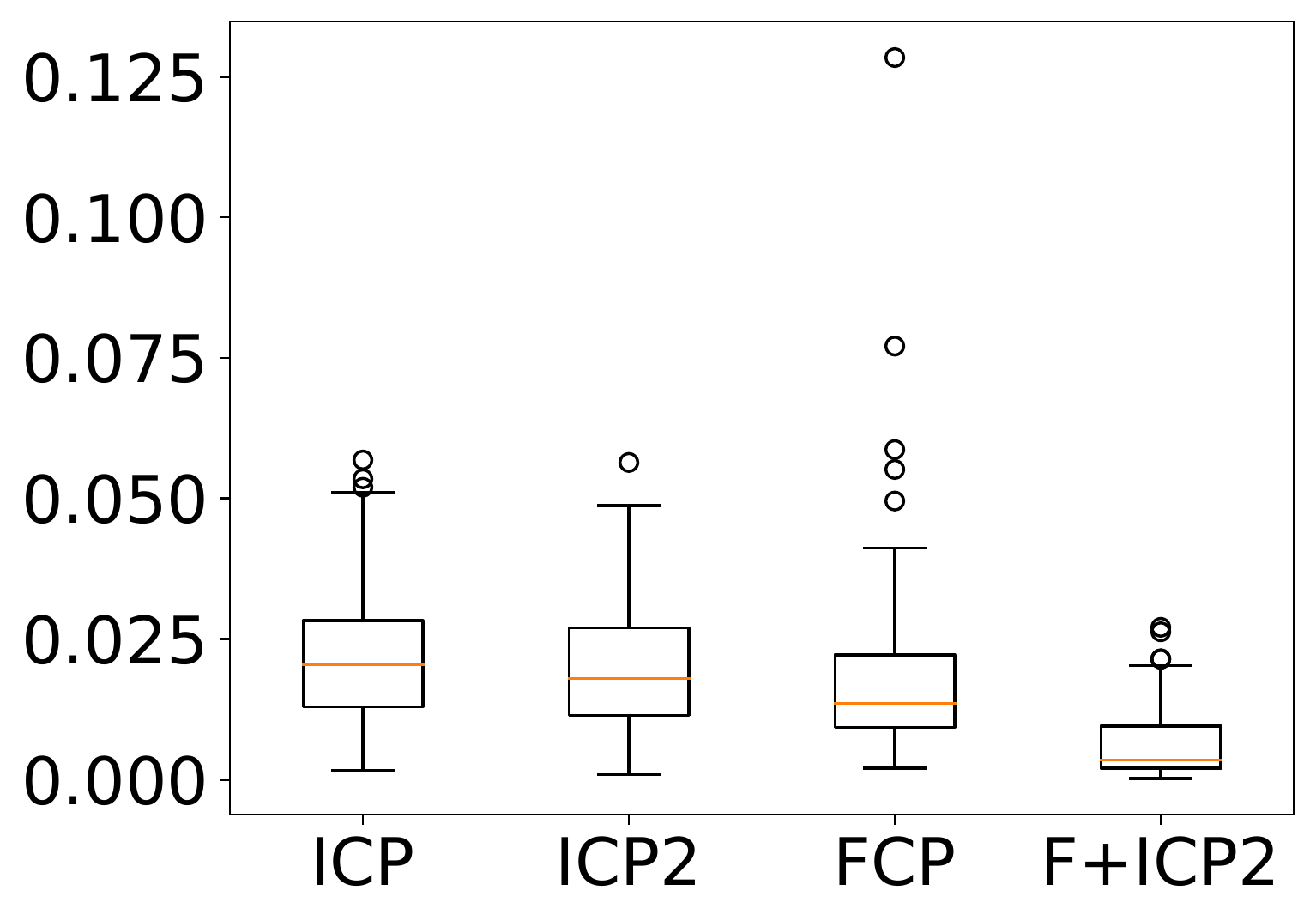}}
	\subfigure[Orientation errors]{
		\includegraphics[width=.45\columnwidth]{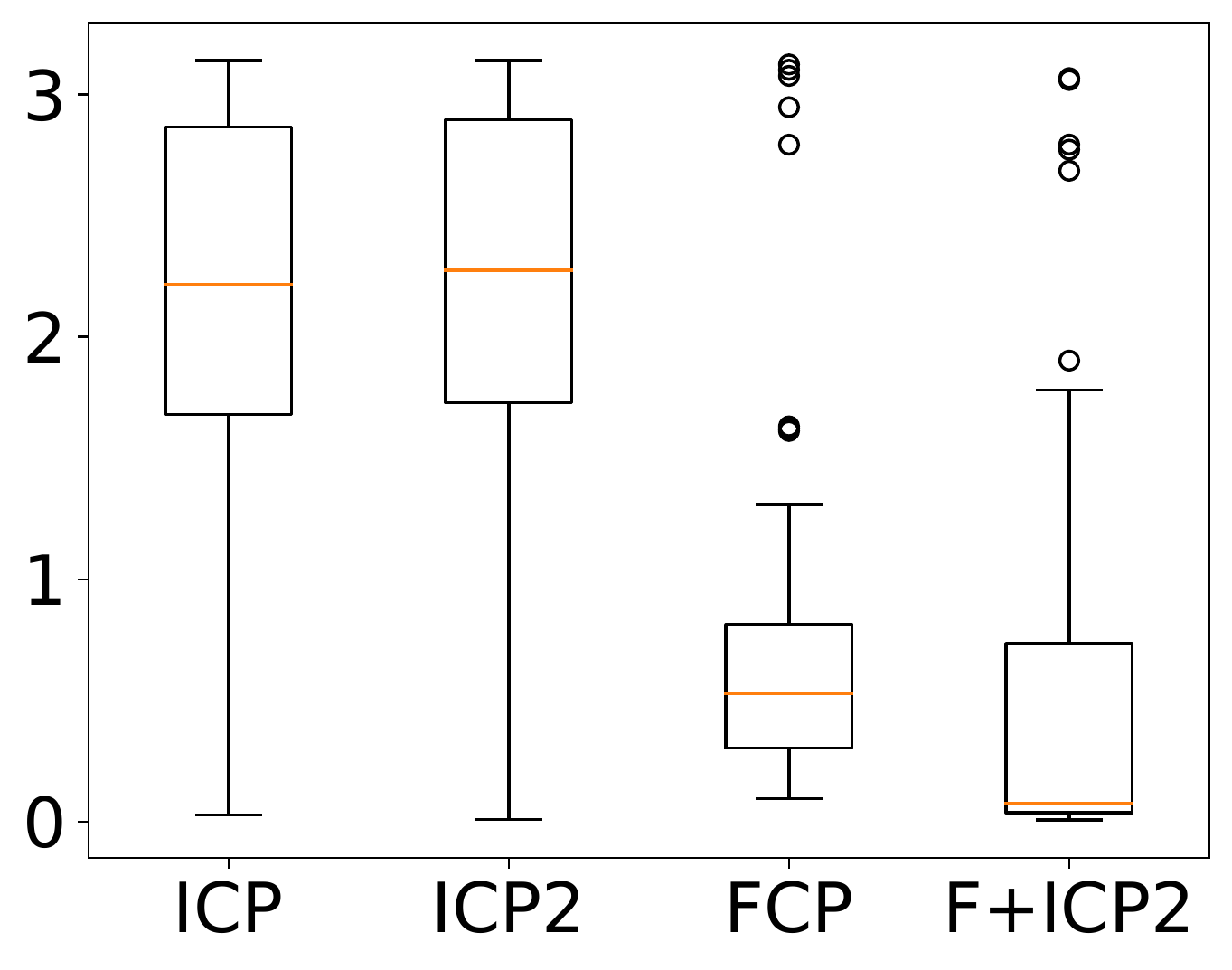}}
	\caption{6D Pose Estimation Results - the estimated poses are applied to the green meshes. ICP easily gets stuck at local optima while FCP produces fairly accurate poses which help F+ICP2 escape the local optima; note that FCP does not iterate to get the results. More images are shown in Fig.~\ref{fig:poseEstim2}.}\label{fig:pose_estim}
\end{figure}	

\begin{figure}[t]
	\centering
	\subfigure[Reference]{
		\includegraphics[width=.31\columnwidth, viewport = 150 150 450 450, clip]{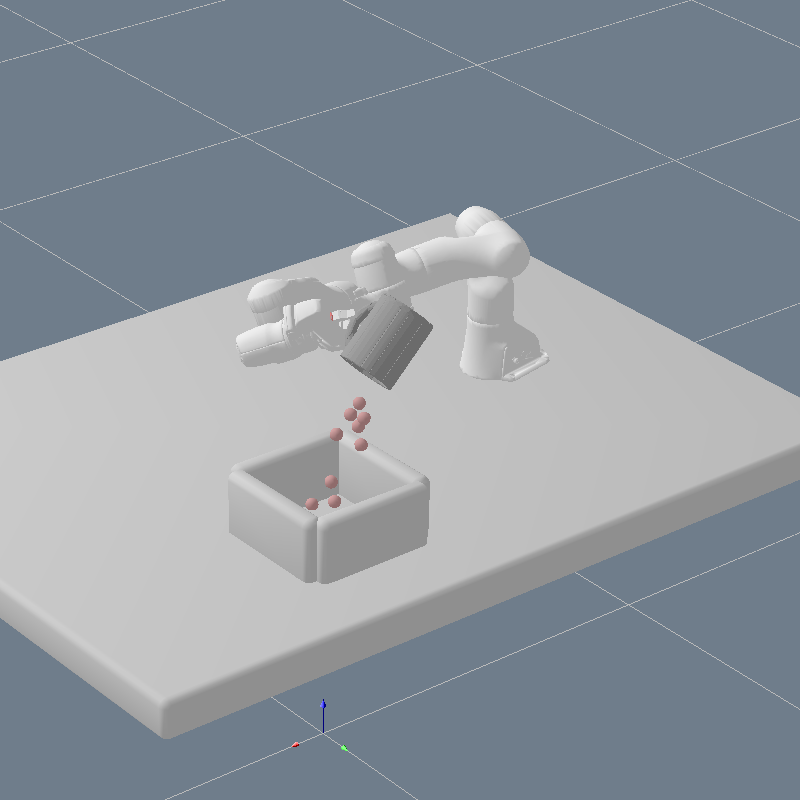}}
	\subfigure[Imitation 1]{
		\includegraphics[width=.31\columnwidth, viewport = 150 150 450 450, clip]{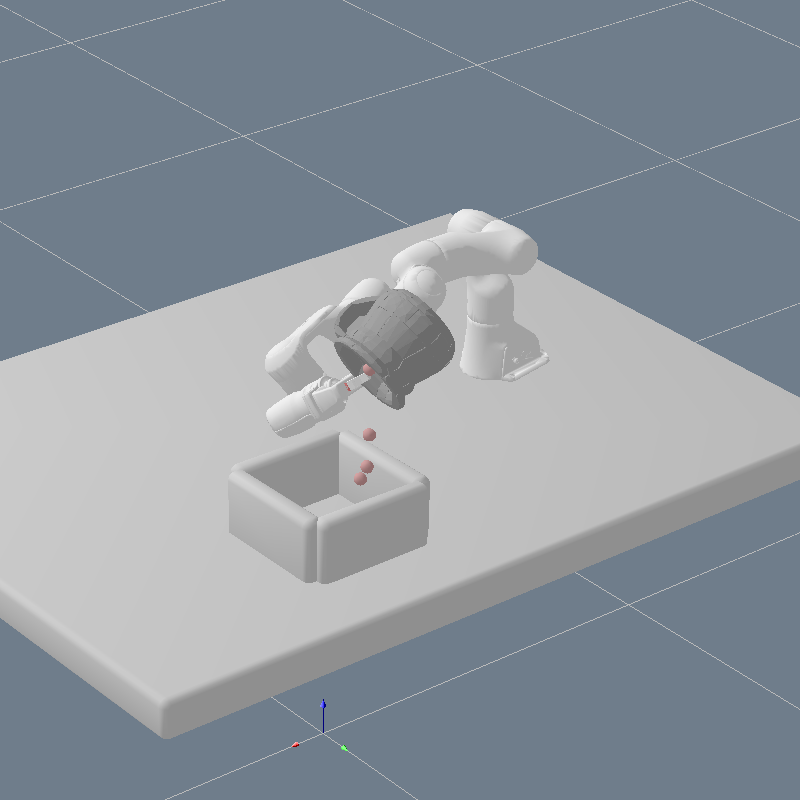}}
	\subfigure[Imitation 2]{
		\includegraphics[width=.31\columnwidth, viewport = 150 150 450 450, clip]{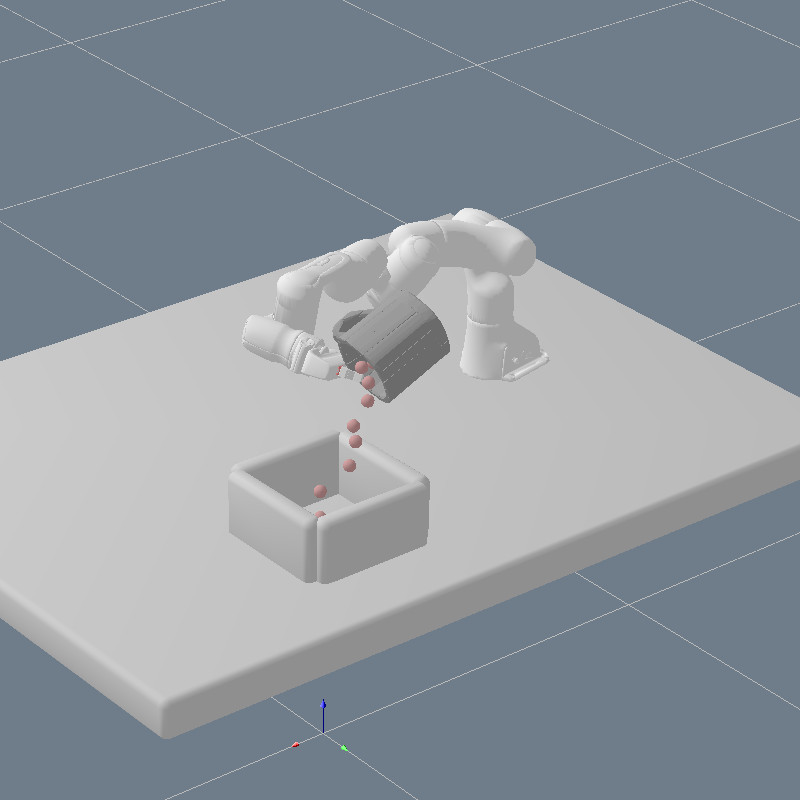}}
	\caption{Zero shot imitation. Detailed views are in Figs.~\ref{fig:ref} -- \ref{fig:imitation_full}.}	\label{fig:imitation}
\end{figure}	

Fig.~\ref{fig:PCA_img0} visualizes three components of the image feature vectors (the outputs of the U-net encoder) from the principal component analysis (PCA).
It can be observed that each component represents a certain property of the objects, such as inside vs.\ outside, handle vs.\ other parts, or above vs.\ below.
This enables the image-based pose estimation which we call feature-based closest point (FCP) matching, i.e., the problem of finding the relative pose of a target mesh w.r.t. a model mesh, without defining any canonical coordinate of the objects. Specifically, the FCP matching works as follows:
\begin{enumerate}
	\item It first queries the backbone at $10^3$ and $5^3$ grid points around the target and the model, respectively, (as shown in Fig.~\ref{fig:poseEstim_grid0}) with their own images.
	\item For each model grid point, the target point is obtained such that their representations are closest.
	\item Finally, it computes a $SE(3)$ pose that minimizes the sum of the model-target pairwise Euclidean distances.
\end{enumerate}
We compared this to the conventional iterative closest point (ICP) algorithm on point clouds, i.e., the problem of finding the relative pose minimizing the Euclidean distance of two sets of point clouds.
The point clouds can be obtained from depth cameras (ICP) or on the surface of the meshes reconstructed via the learned SDF features (ICP2).
The point clouds' size was 1000.
Fig.~\ref{fig:pose_estim} (h, i) shows the position and orientation errors when 131 mugs with random poses were tested.
Notably, as visualized in Fig.~\ref{fig:pose_estim} (d-g), FCP performs much better especially in orientation because, notoriously, ICP easily gets stuck in local optima.
A significant improvement was observed in F+ICP2 where we used the FCP results as starting points of ICP2 (which is performed without depth images).

Another important observation from Fig.~\ref{fig:PCA_img0} is that the semantics of representations are consistent across different objects as well, e.g.\ the handle parts of different mugs have similar representations.
It implies that a pose of one object can be transferred into another through the representation.
We therefore considered an image-based zero-shot imitation scenario, where the environment contains one robot arm, one target mug (filled with small balls) and 4 cameras as shown in Fig.~\ref{fig:imitation}.
We manually designed a pouring motion for one mug and stored the images of pre- and post-pouring postures of the mug, $\mathcal{V}_\text{pre}=(\mathcal{I}_\text{pre}, \mT_\text{pre}, \mK_\text{pre})$ and $\mathcal{V}_\text{post}=(\mathcal{I}_\text{post}, \mT_\text{post}, \mK_\text{post})$, respectively.
For a new mug, we solved LGP with [(\textsc{Grasp}, gripper, mug), (\textsc{PoseFCP}, $\mathcal{V}_\text{pre}$, mug), (\textsc{PoseFCP}, $\mathcal{V}_\text{post}$, mug)], where (\textsc{PoseFCP}, $\cdot$, $\cdot$) imposes the aforementioned FCP constraint at the end of each phase.
That is, the trajectory optimizer tries to match each part of the new object to the corresponding part of the target mug while coordinating the global consistency of the full trajectory (e.g., determining a proper grasp pose for pouring).
Fig.~\ref{fig:imitation} shows the optimized post-pouring posture, which implies that the learned representation allows for imitation of the reference motions \emph{only} from the posed images. The videos can be found at \href{https://youtube.com/playlist?list=PL9pnj8nG83Oe3dMI_dX-7Xd69XXNWONIy}{[video3]}.

\subsection{Real Robot Demonstration}
Fig.~\ref{fig:realRobot} shows our complete framework in the real robot system.
To successfuly apply the learned DVCs to the real robot by closing the sim-to-real gap, we had to extend training to a larger dataset;
specifically, we randomized the material of mugs by adjusting metalness and roughness to get more diverse appearances and also applied more extensive data augmentations, e.g.\ ColorJitter or GaussianBlur.
At test time, we attached RealSense D435 on the gripper and took 8 color images from predefined shooting poses.
We used the pre-trained Mask R-CNN~\cite{he2017mask} to get object masks, which we found provides clear enough segmentation for the network to come up with sensible manipulation plans. 
We refer the readers to the supplementary video (in the caption of Fig. \ref{fig:realRobot}) for visualization of the whole manipulation pipeline.

\section{Discussion} \label{sec:conclusion}
The main idea of the proposed Deep Visual Constrains is twofold:
\begin{enumerate}
	\item Implicit object representations to which manipulation planning algorithms can apply rigid transformations in $SE(3)$ and
	\item the implicit representations trained as a shared backbone of multiple task features, directly via task-supervisions.
\end{enumerate}
Throughout the experiments, we demonstrated the proposed visual manipulation framework both in simulation and with the real robot.
The ablation studies examined each of the proposed techniques, compared to the non-pixel aligned, explicit, and geometric representations as well as the traditional hand-engineered features.
The IK experiments demonstrated the advantage of DVC's joint optimization capability.
We analyzed the learned representations via PCA and found that the generalizable sementics emerged in the representation during training, which enables 6D pose estimation and zero-shot imitation.

Notably, the last finding implies that considering more diverse tasks and objects in our multi-task learning would lead to more generalized representations as well as stronger synergies between individual feature learning, which we leave for future work.
All those task features don't necessarily model physical interaction feasibility for planning; e.g., they can also serve as a value or energy function of a direct control policy and be trained via imitation or reinforcement learning as well~\cite{florence2021implicit,21-driess-ICRA}.

While we have only demonstrated one-robot/static-frame vs.\ one-object interactions, addressing interactions between multiple frames vs.\ object, e.g.\ grasping an object with two hands or elbows, is straightforward by attaching further key interaction points on those frames.
However, interactions between two or more functional objects would require to extend our framework, e.g.\ by concatenating the objects' representation vectors obtained in some pre-defined interaction region, and predicting features based on this concatenation, the details of which we need to leave for future work.
Alternatively, the notion of \textit{Point-of-Attack} can be introduced for some primitive object-object interactions, like touching, inserting, placing and pushing, solely from the geometric feature, SDFs~\citep{xie2016rigid,20-toussaint-RAL,21-driess-CORL}.

Lastly, we would like to emphasize that the idea of DVCs is not limited to RGB input. Point clouds can be considered by replacing the U-net encoder with PointNet~\citep{qi2017pointnet}, which could be a better choice depending on the setting, e.g., whether reliable depth perception is available~\citep{simeonovdu2021ndf}.
Incorporating non-visual, like tactile, input would be another exciting direction to explore. \vspace{-.3cm}

\section*{Acknowledgments}
This research has been supported by the German Research Foundation (DFG) under Germany's Excellence Strategy -- EXC 2002/1--390523135 ``Science of Intelligence".

\bibliographystyle{plainnat}
\bibliography{references}

\clearpage
\section*{Appendix}

\subsection{Homography Transformation} \label{app:homography}
The idea of the Homography warping is that two images taken by cameras at the same position but with different orientations and intrinsics can be transformed into each other.
Suppose that we have a source image $\mathcal{I}$ with the camera position $\vt$, rotation matrix $\mR$ and projection matrix $\mK$ and that an object is inside a bounding sphere at $\vp\in\R^3$ with a radius $r\in\R^+$.
An image focusing on the bounding sphere can be taken from a (synthetic) camera at the same position $\vt$ with the view direction as $\vt-\vp$ and the field of view angle as $2\arcsin(||\vt-\vp||_2/r)$, from which we can compute the new camera rotation matrix $\hat{\mR}$ and the intrinsic $\hat{\mK}$.

Given $\hat{\mR}$ and $\hat{\mK}$, the new field warped by the corresponding Homography can be obtained as follows:
First, a pixel in the source image, $p_1 = (u,v,1)$, is reprojected into a ray in the 3D space: $P_1 = \mK^{-1}p_1$. 
Next, the ray is viewed in the new camera coordinate: $P_2 = \hat{\mR}^{T}\mR P_1$.
Lastly, this ray is projected back into a pixel in the new camera: $p_2 = \hat{\mathbf{K}}P_2$.
Putting all together, the Homography warping is given as:
\begin{align}
\mathcal{W}(\hat{\mR},\hat{\mK}): \begin{bmatrix} u \\ v \\ 1 \end{bmatrix} \mapsto w\hat{\mK}\hat{\mR}^{T}\mR\mK^{-1} \begin{bmatrix} u \\ v \\ 1 \end{bmatrix},
\end{align}
where $w$ is the parameter that makes the last element of the output homogeneous coordinate $1$, which results in the warped image $\hat{\mathcal{I}}$ with its camera pose $\hat{\mT}=\begin{bmatrix}\hat{\mR}& \vt \\ \mathbf{0} & 1\end{bmatrix}$ and intrinsic matrix $\hat{\mK}$.

\subsection{Manipulation Constraints} \label{app:constraints}
In this work, we consider two discrete actions, (\textsc{Grasp}, gripper, mug) and (\textsc{Hang}, hook, mug), for grasping and hanging, respectively. Each action imposes three constraints on the path as follows.
\begin{itemize}
	\item \textbf{The action $\ra_k=(\textsc{Grasp}, \text{gripper}, \text{mug})$} first imposes the learned grasping constraint at the end of its phase, $H^i_\text{grasp}(\vx_t, \delta\vq^i_t)=0,~t = kT$, i.e., 
	\begin{align}
	\left(T(\delta\vq^i_t)[\phi^i_\text{grasp}]\circ FK_j\right)(\vx_t)=0,
	\end{align}
	where $i$ and $j$ are indices of the mug and the gripper, respectively.
	It also imposes the zero-impact switching constraint at $t = kT$ for the smooth transition, i.e.,
	\begin{align}
	\hat\vv_t = 0,
	\end{align}
	where $\hat\vv_t$ is a joint velocity computed from $\vx_{t-1}$ and $\vx_t$ via finite difference.
	Lastly, it introduces an equality constraint on the gripper's approaching direction for collision-safe grasping; 
	more precisely, the constraint is imposed at $t\in\{kT-2, kT-1, kT\}$ as:
	\begin{align}
	{}^j\hat\va^i_t = a_\text{approach}\begin{bmatrix}0\\0\\-1\end{bmatrix},
	\end{align}
	where ${}^j\hat\va^i_t$ is the mug's acceleration in the gripper's coordinate computed from ${}^j\vt(\delta\vq^i_{t-2})$, ${}^j\vt(\delta\vq^i_{t-1})$ and ${}^j\vt(\delta\vq^i_t)$ via finite difference, and $a_\text{approach}\in\R^+$ is the predefined approaching acceleration magnitude.
	The gripper's $z$ axis is depicted in Fig.~\ref{fig:keypoints}(a) as a blue arrow.
	Combined with the above zero-impact constraint, this constraint enforces the gripper to approach the mug in the gripper's -$z$ axis direction and to stop moving at the end of the phase.
	\item Similarly, \textbf{the action $\ra_k=(\textsc{Hang}, \text{hook}, \text{mug})$} consists of the learned hanging constraint, the zero-impact and hanging approaching constraints as
	\begin{align}
	&\left(T(\delta\vq^i_{kT})[\phi_\text{hang}]\circ FK_j\right)(\vx_{kT})=0,\\
	&\hat\vv_{kT} = 0, \\
	&{}^j\hat\va^i_t = a_\text{approach}\begin{bmatrix}0\\0\\1\end{bmatrix},~\forall t\in\{kT-2, kT-1, kT\},
	\end{align}
	where $i$ and $j$ are indices of the mug and the hook, respectively, and the hook's $z$ axis is the blue arrow in Fig.~\ref{fig:keypoints}(b) (or outer product of the red and green arrow).
\end{itemize}

The discrete actions above affect the consecutive symbolic states.
While $s_k$ indicates a mug is grasped by a gripper or hung on a hook at the phase $k$, we impose the following path constraint:
$\forall t\in\{(k-1)T+1, \cdots, kT\}$
\begin{align}
\delta\vq^i_{t}-\delta\vq^i_{t-1} = FK_j(\vx_{t})-FK_j(\vx_{t-1}),
\end{align}
where $i$ and $j$ are indices of the mug and the gripper/hook, respectively.
Effectively this introduces a static joint between the two frames~\citep{toussaint2018differentiable} so the mug moves along with its parent frame (the gripper or hook).
The collision constraints are also imposed along the trajectory, where the pair collisions with the mug are computed by the learned SDF feature.
We introduce the collision feature in the following section.

We would like to emphasize that our manipulation planning framework is not limited by the constraints we introduced above, but it can incorporate any existing constraint models and methods for general dexterous manipulation, e.g.,~\citep{toussaint2018differentiable,20-ha-ICRA,20-toussaint-RAL,driess2021learning}.

\subsection{Defining Pair-Collision Constraints with SDFs}\label{app:collision}
For manipulation planning problems written only by convex meshes, the distance or penetration of two objects, which we call pair-collision features, are computed with either Gilbert-Johnson-Keerthi (GJK) for non-penetrating objects or Minkowski Portal Refinement (MPR) for penetrating objects.
In this section, we introduce how to define pair-collision features when one or both objects are given as SDFs.

{\bf SDF vs. Sphere: } Let $\delta\vq_i$, $\vq_j$ and $r_j$ be the rigid transformation of PIFO, the sphere's pose and radius, respectively. Then the pair-collision feature is simply given by:
\begin{align}
d_{ij} = T(\delta\vq_i)[\phi_\text{SDF}](\vt(\vq_j))-r_j.
\end{align}

{\bf SDF vs. Capsule: } Let $\delta\vq_i$, $\vq_j$, $h_j$ and $r_j$ be the rigid transformation of PIFO, the capsule's pose, height and radius, respectively. The pair-collision feature is given by the solution of the following optimization:
\begin{align}
d_{ij} = \min_{-h_j/2\le z\le h_j/2}T(\delta\vq_i)[\phi_\text{SDF}]\left(\mR(\vq_j)\begin{bmatrix}0\\0\\z\end{bmatrix}+\vt(\vq_j)\right)-r_j. \label{eq:coll_capsule}
\end{align}

{\bf SDF vs. Mesh: } Let $\delta\vq_i$ and $\vq_j$ be the rigid transformation of PIFO, the mesh's pose, respectively.
\begin{align}
d_{ij} = \min_{\substack{\vp_1\in\R^3, \vp_2\in\R^3\\T(\delta\vq_i)[\phi_\text{SDF}]\left(\vp_1\right)=0\\d_{j}(\vp_2) = 0}} \vn_1^T(\vp_2 - \vp_1), \label{eq:coll_mesh}
\end{align}
where $\vn_1$ is the normal vector of $\phi_\text{SDF}$ at $\vp_1$ and $d_{j}(\vp_2)$ is the signed distance of $\vp_2$ to the mesh computed by GJK/MPR.

{\bf SDF vs. SDF: } Let $\delta\vq_i$ and $\delta\vq_j$ be the rigid transformations of two PIFOs.
\begin{align}
d_{ij} = &\min_{\substack{\vp_1\in\R^3, \vp_2\in\R^3\\T(\delta\vq_i)[\phi^i_\text{SDF}]\left(\vp_1\right)=0\\T(\delta\vq_j)[\phi^j_\text{SDF}]\left(\vp_2\right)=0}} \vn_1^T(\vp_2 - \vp_1). \label{eq:coll_sdf}
\end{align}

The optimizations in (\ref{eq:coll_capsule})--(\ref{eq:coll_sdf}) should be run multiple times from different initial guesses because the object shape represented as SDF can be non-convex.
In practice, we found approximating the meshes by a number of spheres and computing the collision feature much more efficient because querying the network $\phi_\text{SDF}$ at multiple points can be done in parallel on GPUs.

\subsection{Network Parameters}
{\bf Image encoder} has the U-net architecture~\citep{ronneberger2015u}, especially with the headless ResNet-34~\citep{he2016deep} as its downward path and two residual  $3\times3$ convolutions followed by up-convolution as the upward path. The number of output channels is $64$.

{\bf 3D reprojector} computes the coordinate feature as 32-dimensional vector using one linear+ReLU layer and concatenate it with the local image feature. They are passed to two hidden layers with the width of (256, 128) followed by ReLUs. Therefore, the dimension of the representation vector is 128.

{\bf SDF head} takes as input one representation vector and computes the output through one hidden layer with the width of 128  followed by ReLU.

{\bf Grasp and hang heads} take as input $27$ and $5$ representation vectors at their interaction points (depicted in Fig.~\ref{fig:keypoints}) and predict the feature through two hidden layers with the widths of $(256, 128)$ followed by ReLUs.

As shown in Figure~\ref{fig:baseline}, the network structures for comparison in Section~\ref{sec:learned feature} was kept similar to the above as possible.
Image encoders of the global image feature and vector representation networks are the ResNet-34 returning $64$-dimensional vector.
The feature head structures remain the same, but, because the vector representation scheme doesn't represent objects as implicit functions, the input of their feature head is the frame's pose as 7-dimensional vector (3D translation+4D quaternion).
The grasp and hang heads of the SDF representation scheme take as input $27$- and $5$-dimensional vectors of their interaction points SDF values.

\subsection{Hand-Designed Constraint Models}
Throughout the experiments, objects are represented by meshes, especially with convex-decomposition using \href{https://github.com/kmammou/v-hacd}{the V-HACD library}~\citep{mamou2016volumetric} for non-convex shapes, and thus pair-distance and collision between meshes can be computed via the GJK/MPR algorithm. On top of this mesh representation, the grasping and hanging constraints are defined and optimized as follows.

{\bf The grasping constraint} consists of the aforementioned collision constraints and the so-called oppose constraint. The oppose feature takes as input three meshes, \textsc{finger1}, \textsc{finger2}, and (a set of decomposed) \textsc{object} meshes to grasp. It computes the minimum pair-distances from \textsc{finger1} and \textsc{finger2} to \textsc{object} and returns summation of those two vectors, i.e., $\vv_{\textsc{finger1}\rightarrow\textsc{object}}+\vv_{\textsc{finger2}\rightarrow\textsc{object}}$. 
Making the oppose feature $0$ places the object in the middle of two fingers with proper orientation. 
This geometric heuristic is inspired by the notion of force-closure for two-point grasping~\citep{murray2017mathematical,ten2017grasp} and works very well for simple shapes, such as spheres, capsules, etc.
Because the mug shapes are highly non-convex we ran the optimization from 100 initial seeds and took the best one with the minimum constraint violation.

{\bf The hanging feature}, given the object mesh, iteratively generates a collision-free pose (up to 10,000 iterations) and checks if the hook is kinematically trapped by the mug (as done in data generation). If trapped, it returns the pose difference so that optimizer can output the found pose.

\subsection{Additional Tables and Figures}
The content is in the next pages.

\onecolumn
\begin{table}[t]
	\caption{SDF Feature Evaluation (Training / Test). The SDF errors were also measured at the same grid points as IoU.}
	\label{tab:individual_sdf}
	\begin{center}
		\begin{tabular}{clccc}
			\multicolumn{1}{c}{\# of views} & \multicolumn{1}{c}{Method}  &\multicolumn{1}{c}{ SDF error ($\times10^{-3}$)} &\multicolumn{1}{c}{Volumetric IoU} & \multicolumn{1}{c}{ Chamfer-$L_1$ ($\times10^{-3}$)} 
			\\ \hline \\
			\multirow{3}{*}{ 2 } & PIFO         & 2.91 / 4.63 & 0.760 / 0.577  & 6.14 / 8.84 \\
			& Global Image Feature & 3.58 / 4.96  		 & 0.642 / 0.515 & 8.50 / 10.8  \\
			& Vector Representation & 15.6 / 15.8 & 0.045 / 0.046 & 39.1 / 40.4 \\
			& SDF Representation & \textbf{2.11 / 3.48} & \textbf{0.786 / 0.622} & \textbf{5.78 / 8.13}
			\\ \hline \\
			\multirow{3}{*}{ 4 } & PIFO         & 2.20 / 3.38 & 0.816 / 0.656  & 5.26 / 6.90 \\
			& Global Image Feature & 2.82 / 3.93 & 0.697 / 0.581 & 7.42 / 9.49  \\
			& Vector Representation & 15.0 / 15.2 & 0.036 / 0.014 & 38.6 / 39.7 \\ 
			& SDF Representation & \textbf{1.43 / 2.73} & \textbf{0.845 / 0.667} & \textbf{4.90 / 6.83}
			\\ \hline \\
			\multirow{3}{*}{ 8 } & PIFO         & 1.68 / 2.72 & 0.851 / 0.683  & 4.78 / 6.34 \\
			& Global Image Feature & 2.31 / 3.51 & 0.728 / 0.607 & 6.75 / 8.80  \\
			& Vector Representation & 14.6 / 15.3 & 0.033 / 0.006 & 38.7 / 40.6 \\
			& SDF Representation & \textbf{1.07 / 2.07} & \textbf{0.878 / 0.703} & \textbf{4.51 / 6.06}
		\end{tabular}
	\end{center}
\end{table}
\begin{table}[h]
	\caption{Task Feature Evaluation (Training / Test).}
	\label{tab:individual_opt}
	\begin{center}
		\begin{tabular}{clcccc}
			\multicolumn{1}{c}{\# of views} & \multicolumn{1}{c}{Method}  &\multicolumn{1}{c}{Grasp ($\%$)}&\multicolumn{1}{c}{Grasp+c ($\%$)} &\multicolumn{1}{c}{Hang ($\%$)} &\multicolumn{1}{c}{Hang+c ($\%$)}
			\\ \hline \\
			\multirow{3}{*}{ 2 } & PIFO         & 65.8 / 55.4 & \textbf{82.9 / 77.1} & 87.2 / \textbf{71.4} 		   & \textbf{88.2 / 72.1} \\
			& Global Image Feature & \textbf{67.6 / 63.9} 		   & 80.9 / 70.4  		  & \textbf{88.3} / 70.4   & 86.3 / 71.8  \\
			& Vector Representation & 13.2 / 12.9 & 0.8 / 0.4 			  & 25.6 / 21.8 		   & 0.0 / 0.0 \\
			& SDF Representation & 41.2 / 55.3 & 49.6/ 45.7 			  & 2.6 / 1.1		   & 3.3 / 2.1 
			\\ \hline \\
			\multirow{3}{*}{ 4 } & PIFO         & \textbf{69.0 / 63.9} & \textbf{88.1 / 82.5} & 88.7 / 75.4 		   & \textbf{94.0 / 78.9} \\
			& Global Image Feature & 62.3 / 61.8 		   & 82.7 / 75.7  		  & \textbf{90.3 / 75.7}   & 91.2 / 78.2  \\
			& Vector Representation & 21.2 / 22.5 & 0.5 / 0.4 			  & 55.1 / 46.4 		   & 0.0 / 0.0 \\
			& SDF Representation & 49.1 / 46.1 & 67.9 / 64.3 			  & 3.3 / 2.9 		   & 3.7 / 4.3
			\\ \hline \\
			\multirow{3}{*}{ 8 } & PIFO         & \textbf{71.9 / 69.3} & \textbf{88.7 / 85.0} & \textbf{91.7 / 80.4} 	& \textbf{96.5 / 82.5} \\
			& Global Image Feature & 71.3 / 67.1 		   & 84.0 / 79.3  		  & 91.3 / 77.5   			& 92.9 / 80.4  \\
			& Vector Representation & 29.0 / 23.9 & 0.5 / 0.7 			  & 65.9 / 49.6 		   & 0.0 / 0.0 \\
			& SDF Representation & 51.4 / 52.1 & 75.5 / 70.4 			  & 4.6 / 6.1 		   &  6.3/ 5.7 
		\end{tabular}
	\end{center}
\end{table}

\begin{figure}[t]
	\centering
	\subfigure[]{
		\includegraphics[width=.32\columnwidth]{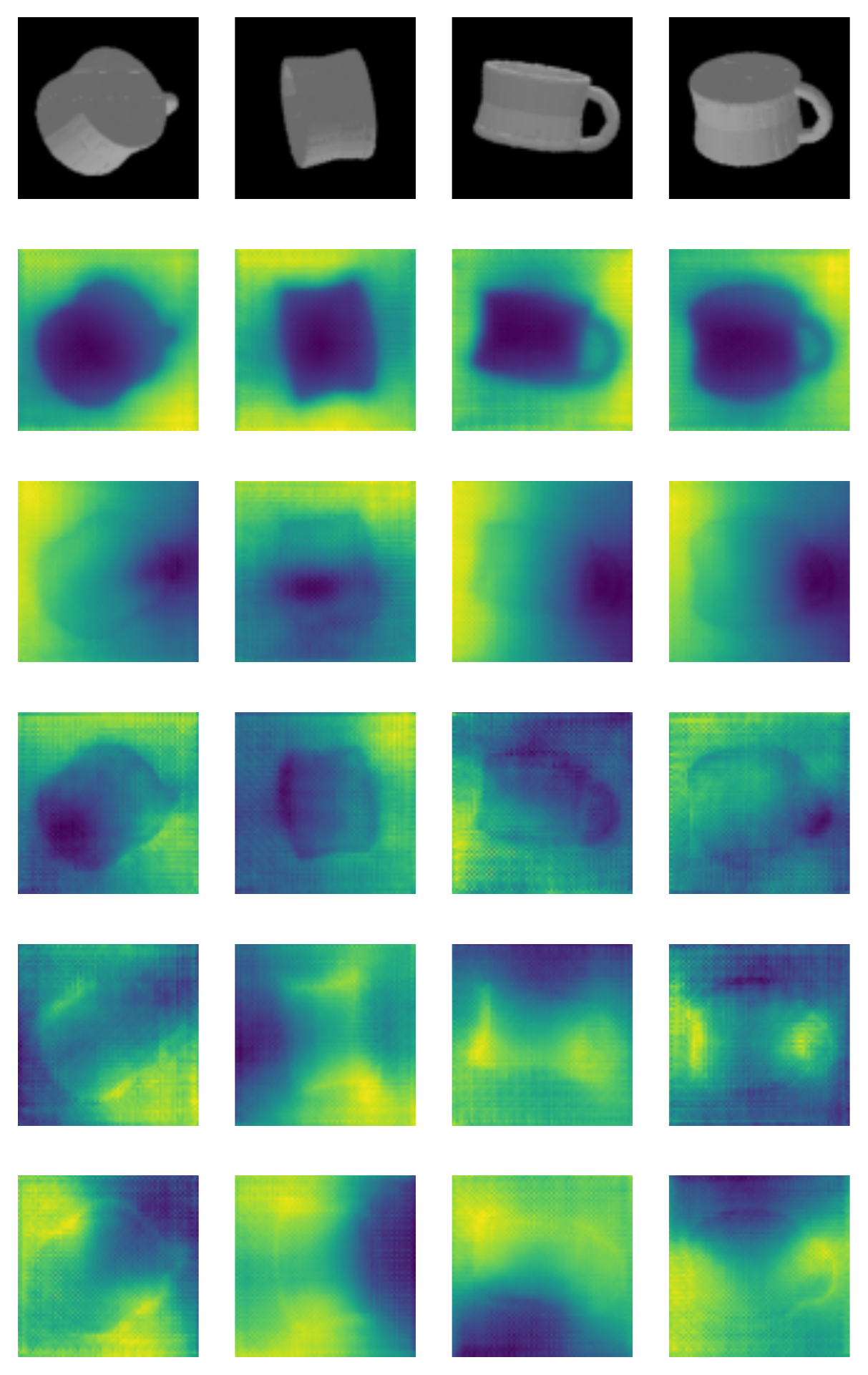}}
	\subfigure[]{
		\includegraphics[width=.32\columnwidth]{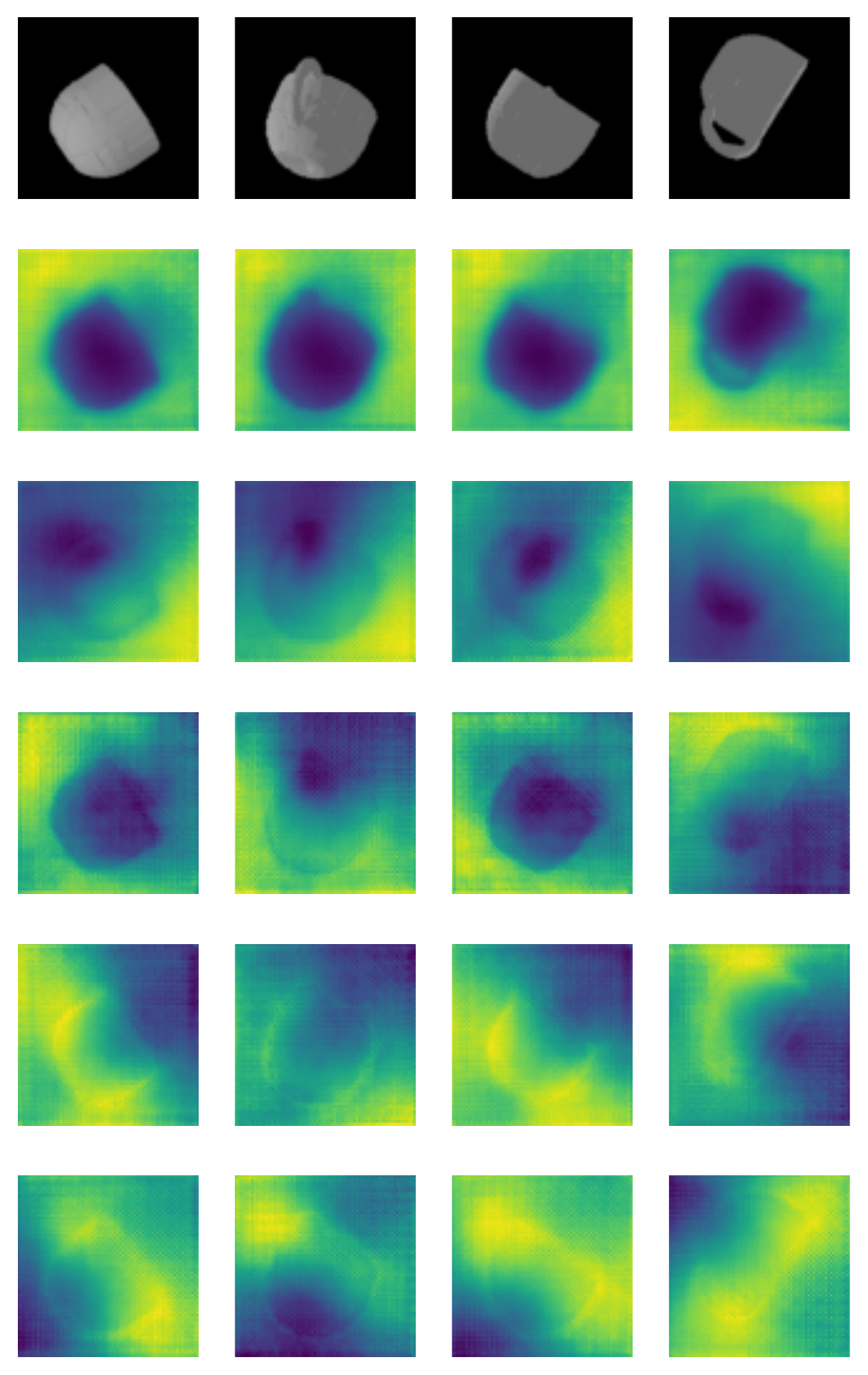}}
	\subfigure[]{
		\includegraphics[width=.32\columnwidth]{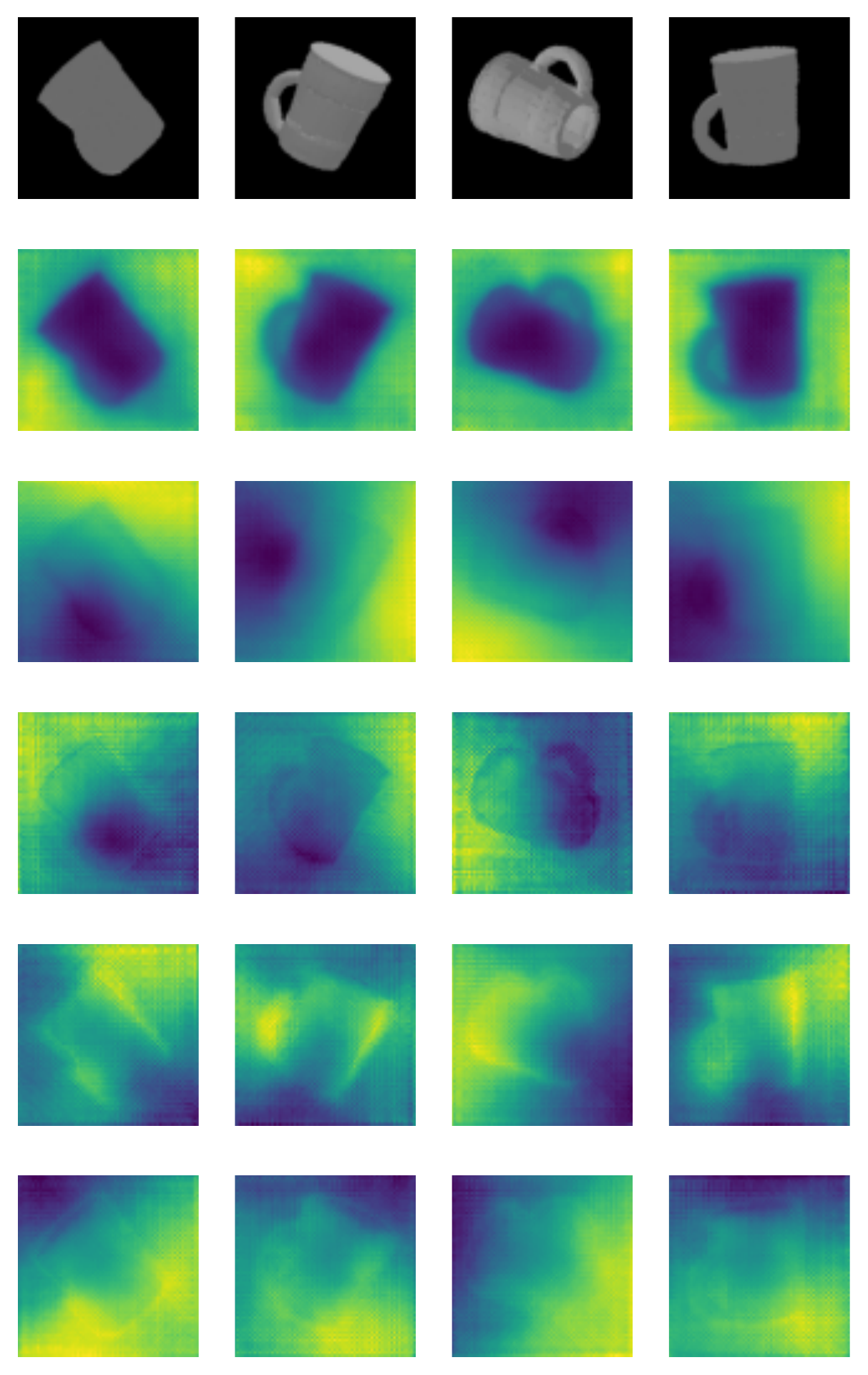}}
	\caption{First 5 principal components from PCA on image features. The first component indicates the object vs. non-object areas, the second component distinguishes the handle parts, and the third one spots the above vs. below of the mugs, etc. Note that the network is trained only via the task feature supervisions.}\label{fig:PCA_img}
\end{figure}

\begin{figure}[t]
	\centering
	\subfigure[Train mugs (1st)]{
		\includegraphics[width=.49\columnwidth]{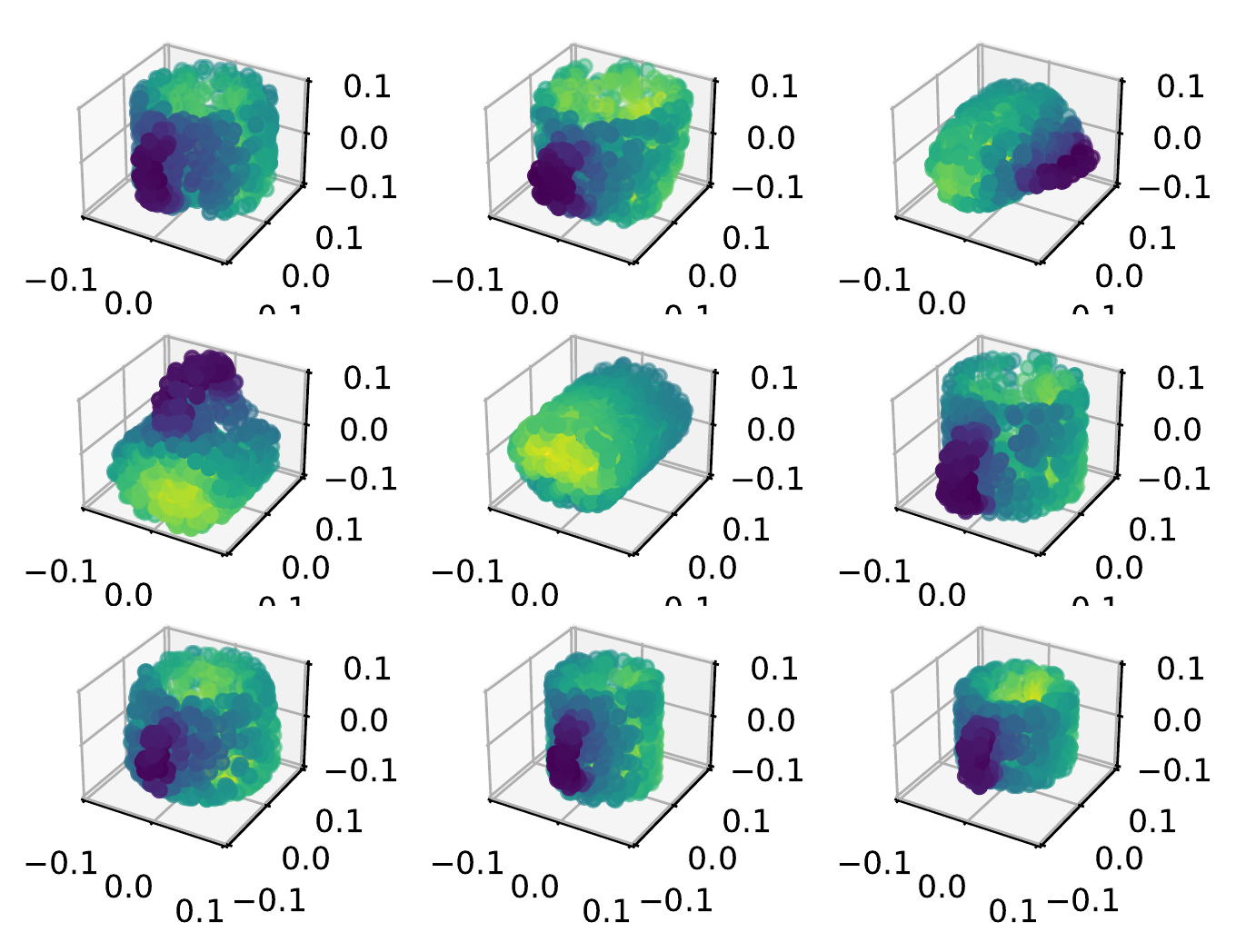}}
	\subfigure[Test mugs (1st)]{
		\includegraphics[width=.49\columnwidth]{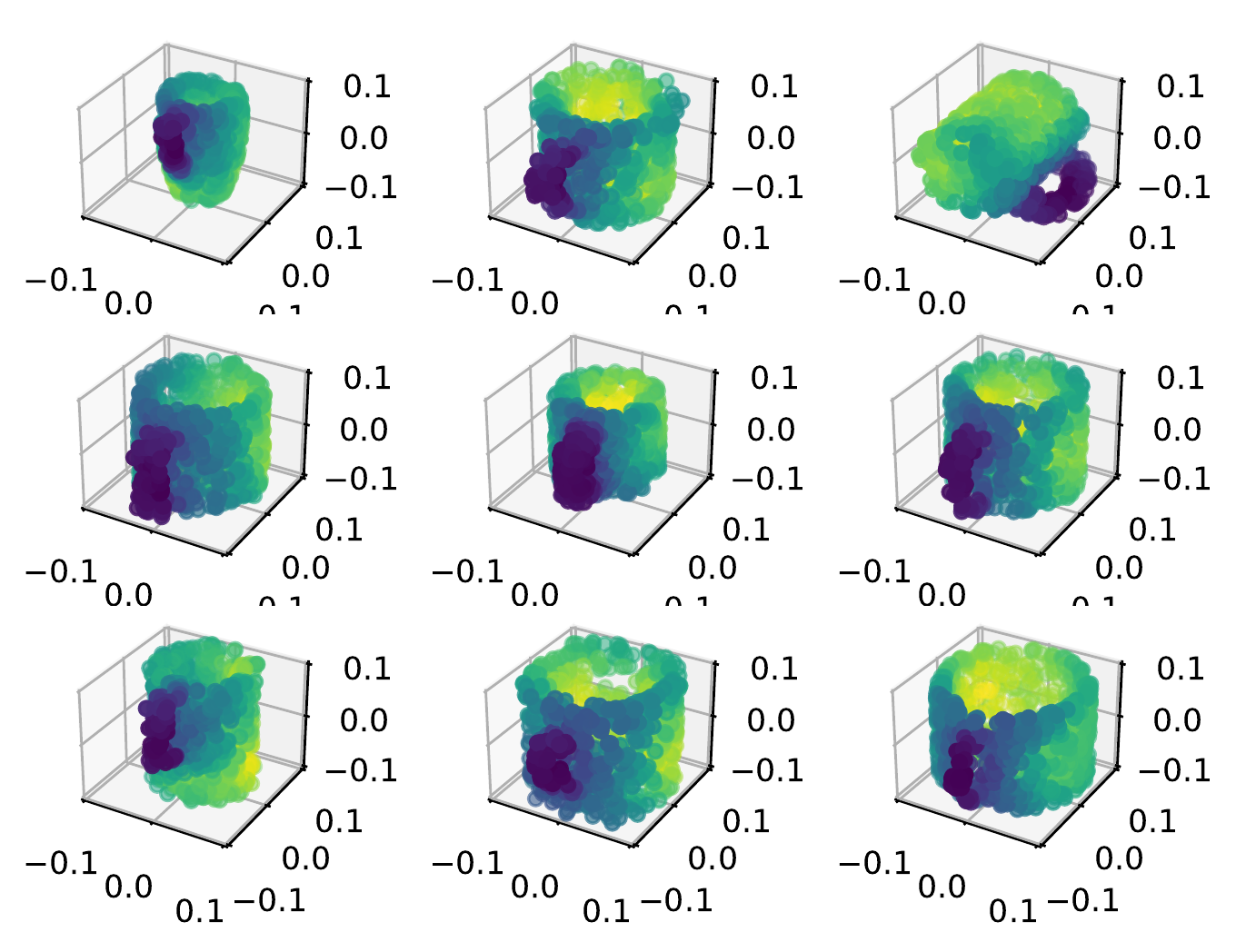}}
	
	\subfigure[Train mugs (2nd)]{
		\includegraphics[width=.49\columnwidth]{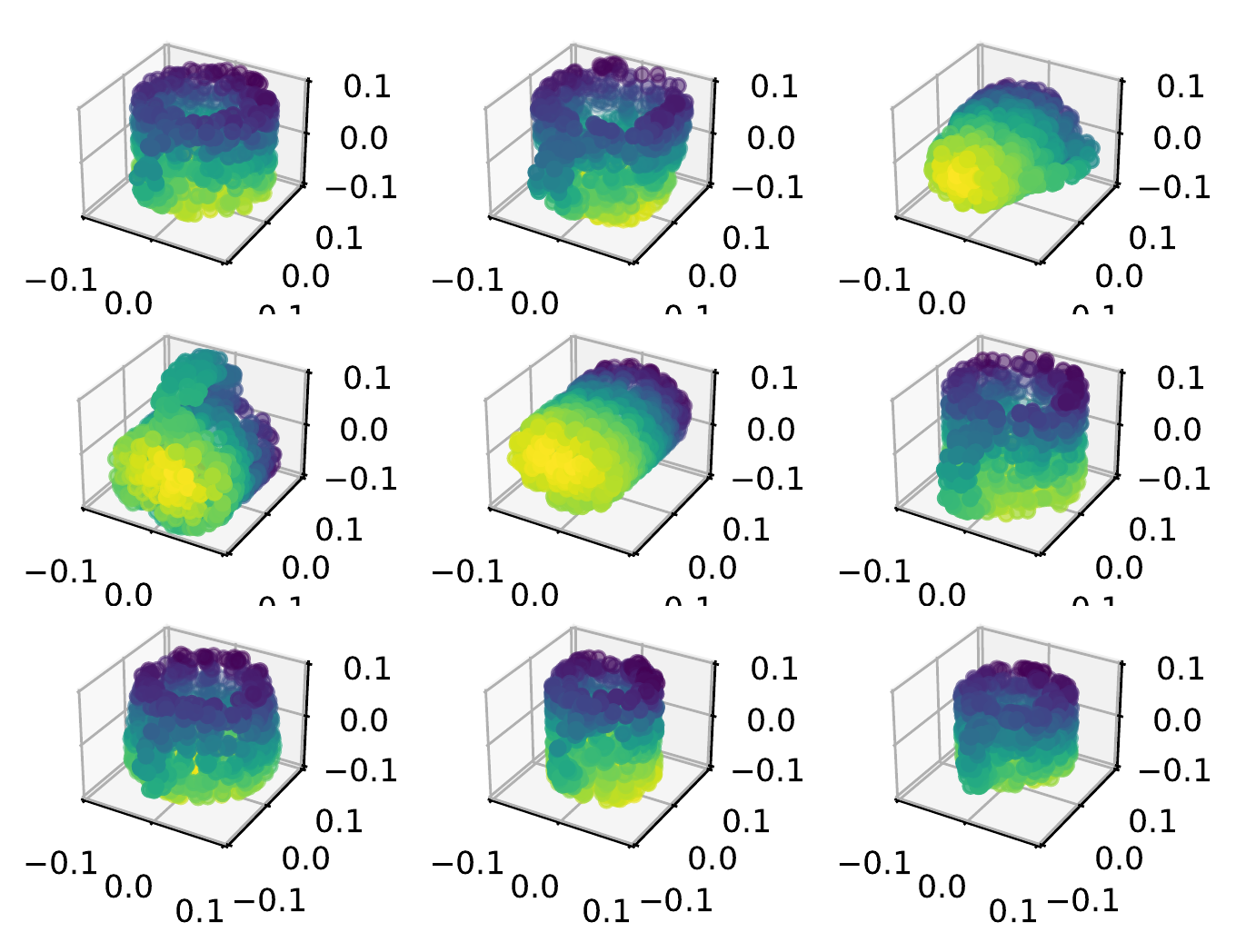}}
	\subfigure[Test mugs (2nd)]{
		\includegraphics[width=.49\columnwidth]{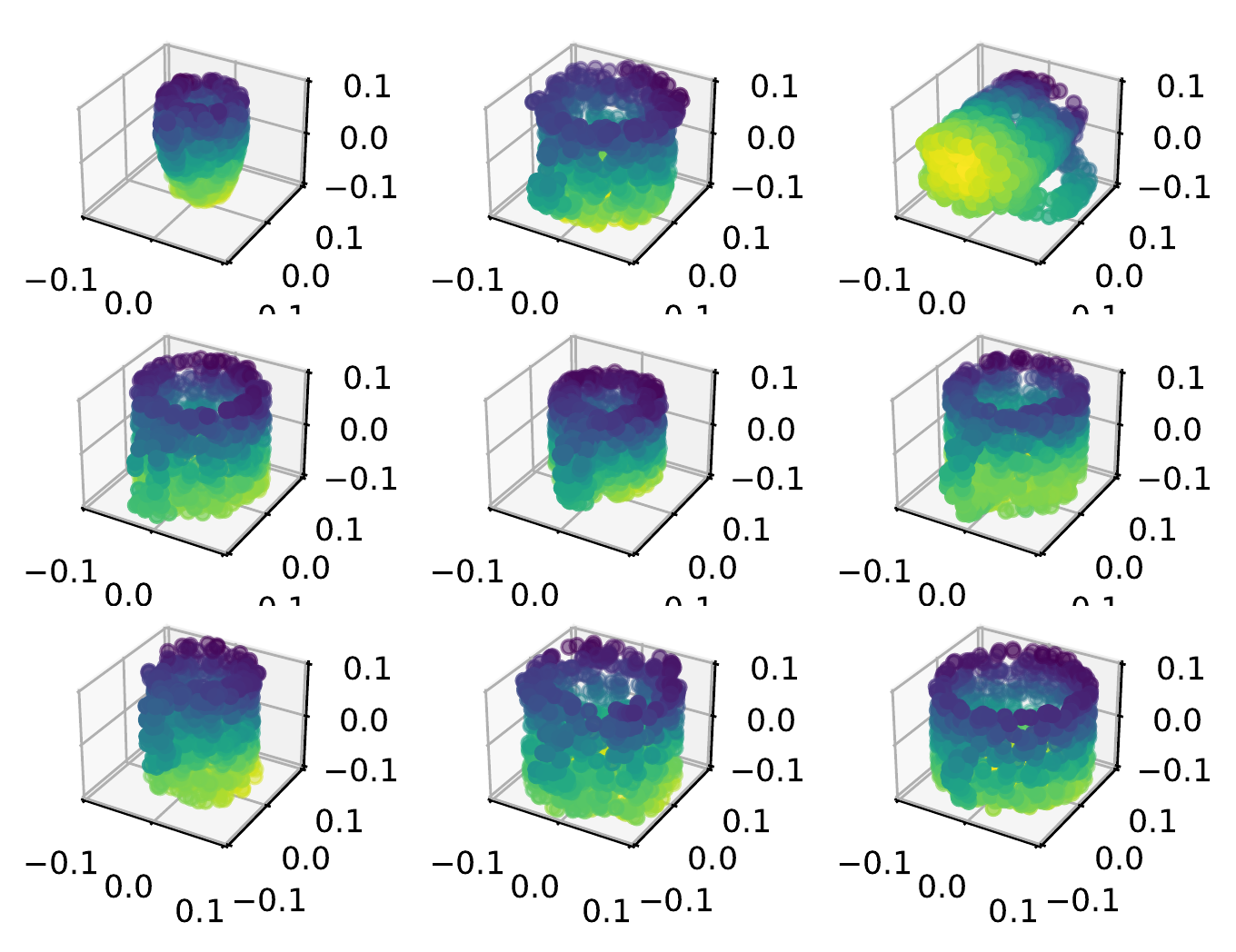}}
	
	\subfigure[Train mugs (3rd)]{
		\includegraphics[width=.49\columnwidth]{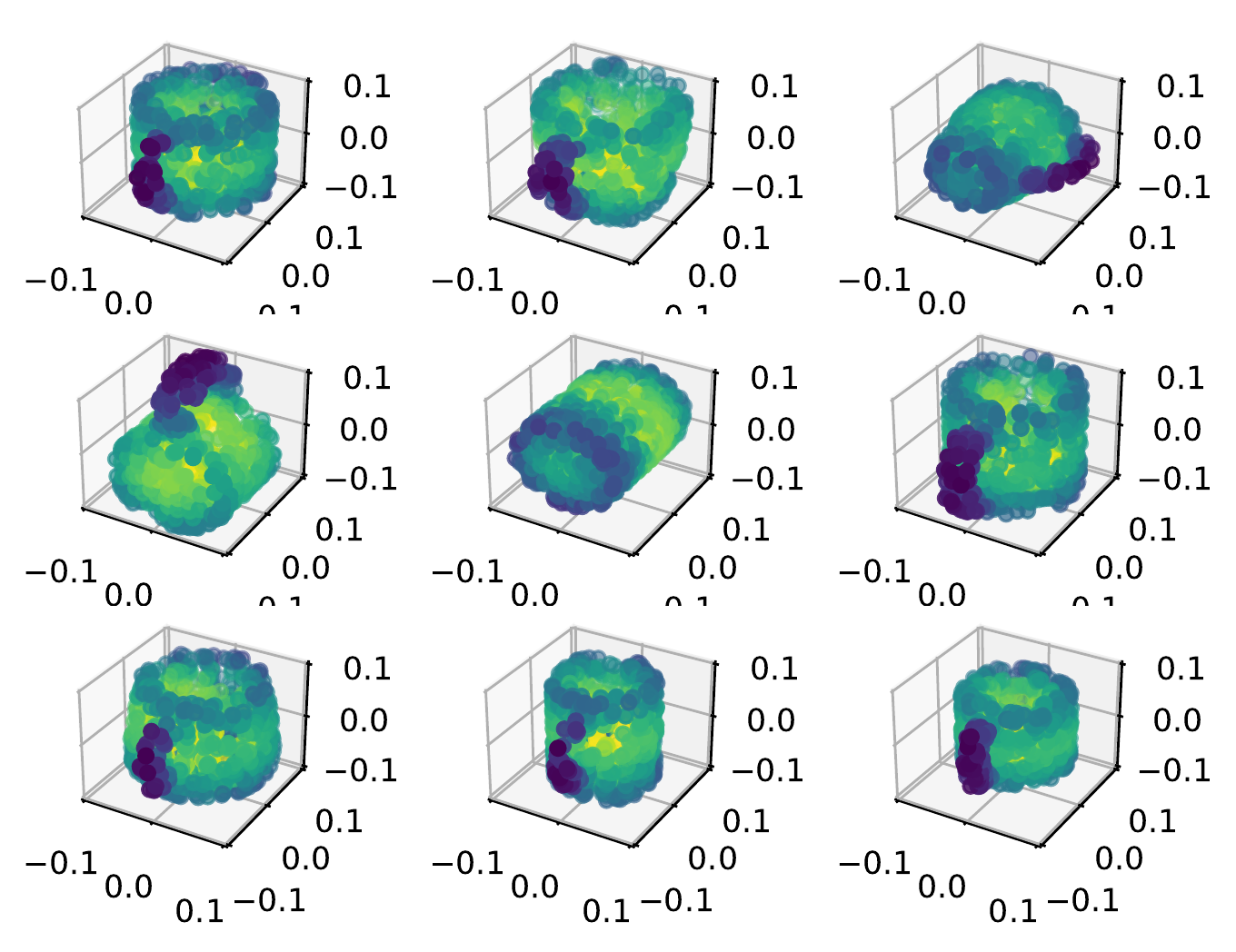}}
	\subfigure[Test mugs (3rd)]{
		\includegraphics[width=.49\columnwidth]{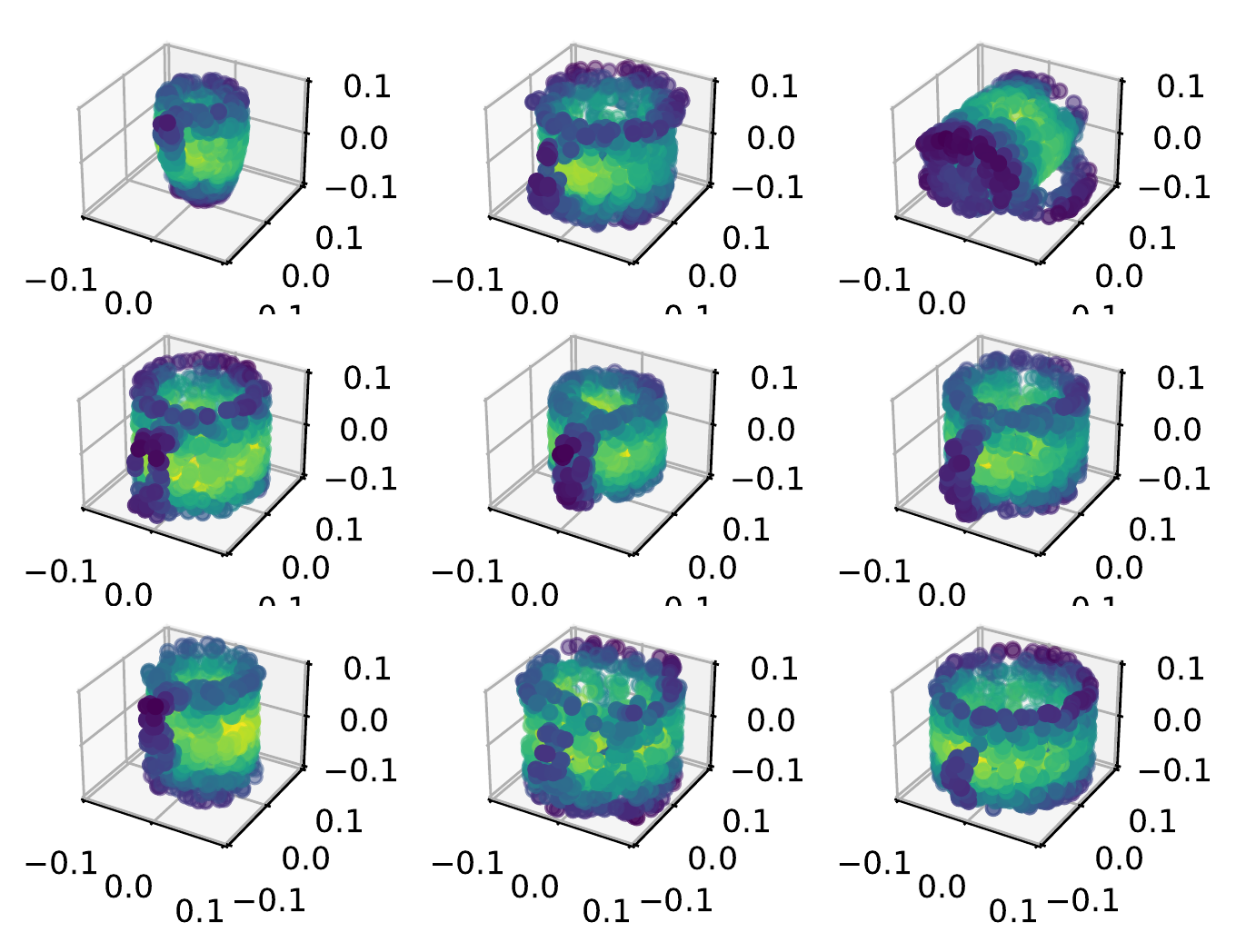}}
	\caption{First 3 principal components from PCA on representation vectors of the 3D surface points. It distinguishes the handles of the mugs from the other parts and is consistent across different mugs.}\label{fig:PCA_3d}
\end{figure}	

\begin{figure}[t]
	\centering
	\subfigure[Model (right) and target (left) mugs]{
		\includegraphics[width=.45\columnwidth, viewport=0 0 800 780, clip]{poseEstim/0}}
	\subfigure[Point clouds for ICP]{
		\includegraphics[width=.45\columnwidth, viewport=0 0 800 780, clip]{poseEstim/1}}
	\subfigure[Point clouds for ICP2 obtained from meshes reconstructed via $\phi_\text{SDF}$]{
		\includegraphics[width=.45\columnwidth, viewport=0 0 800 780, clip]{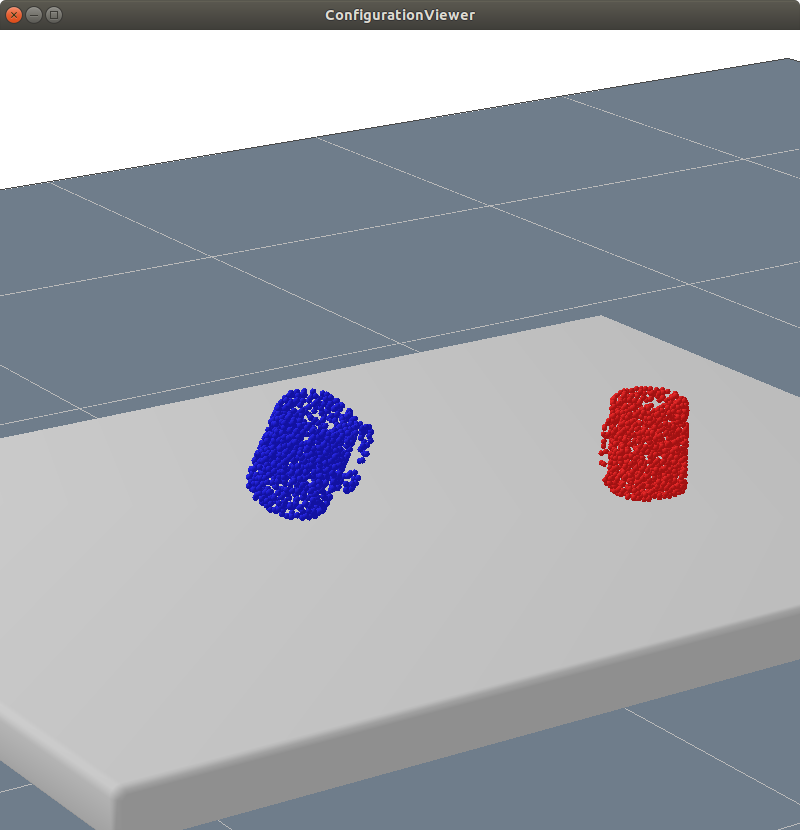}}
	\subfigure[Grid points for FCP]{
		\includegraphics[width=.45\columnwidth, viewport=0 0 800 780, clip]{poseEstim/3}\label{fig:poseEstim_grid}}
	\caption{6D Pose Estimation.  (b) Point clouds for ICP are obtained from depth cameras at the same locations/orientations as the RGB cameras. The size of the point clouds is 1000. (c) Point clouds for ICP are sampled from the surfaces of the meshes reconstructed via the learned $\phi_\text{SDF}$. The size of the point clouds is 1000. (d) FCP uses $10^3$ grid points for the target and $5^3$ grid points (in smaller area) for the model, respectively.}\label{fig:poseEstim1}
\end{figure}

\begin{figure}[t]
	\centering
	\subfigure[ICP]{
		\includegraphics[width=.23\columnwidth, viewport=0 0 800 780, clip]{poseEstim/10}}
	\subfigure[ICP2]{
		\includegraphics[width=.23\columnwidth, viewport=0 0 800 780, clip]{poseEstim/11}}
	\subfigure[FCP]{
		\includegraphics[width=.23\columnwidth, viewport=0 0 800 780, clip]{poseEstim/12}}
	\subfigure[F+ICP2]{
		\includegraphics[width=.23\columnwidth, viewport=0 0 800 780, clip]{poseEstim/13}}
	
	\subfigure[ICP]{
		\includegraphics[width=.23\columnwidth, viewport=0 0 800 780, clip]{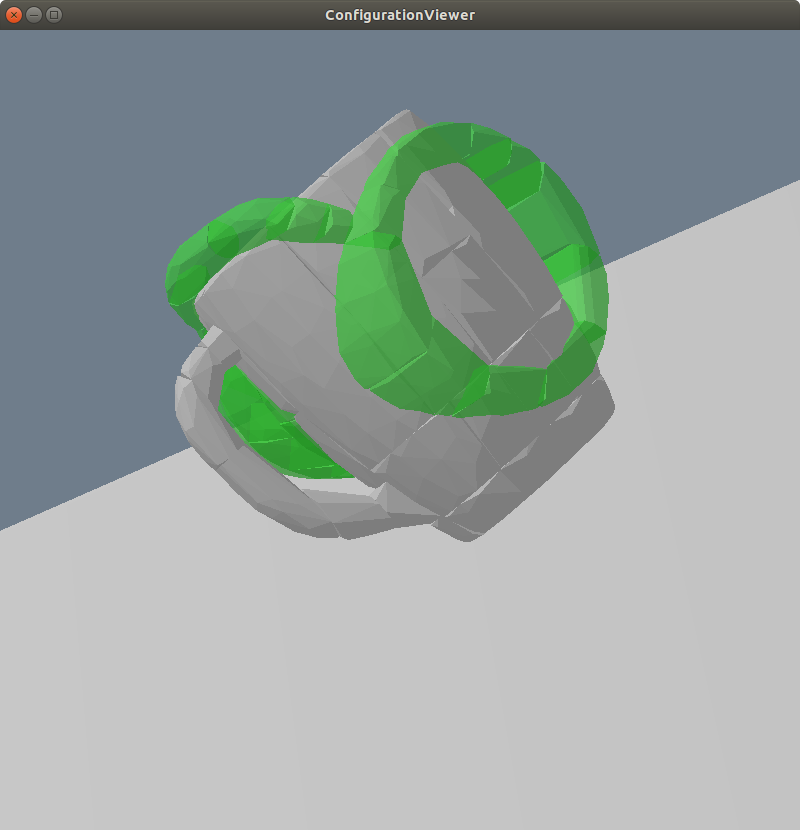}}
	\subfigure[ICP2]{
		\includegraphics[width=.23\columnwidth, viewport=0 0 800 780, clip]{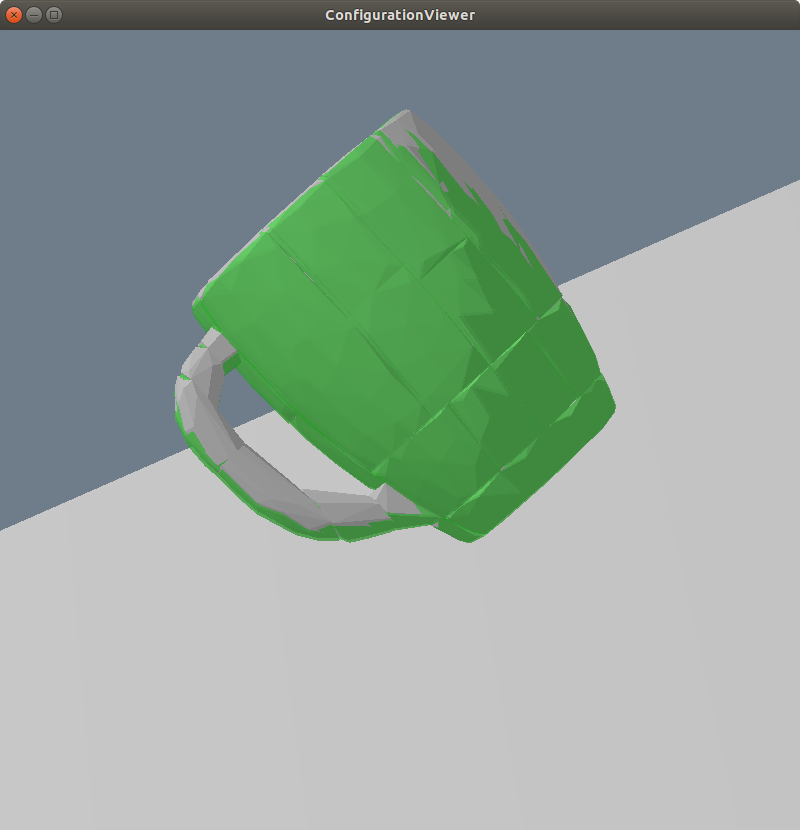}}
	\subfigure[FCP]{
		\includegraphics[width=.23\columnwidth, viewport=0 0 800 780, clip]{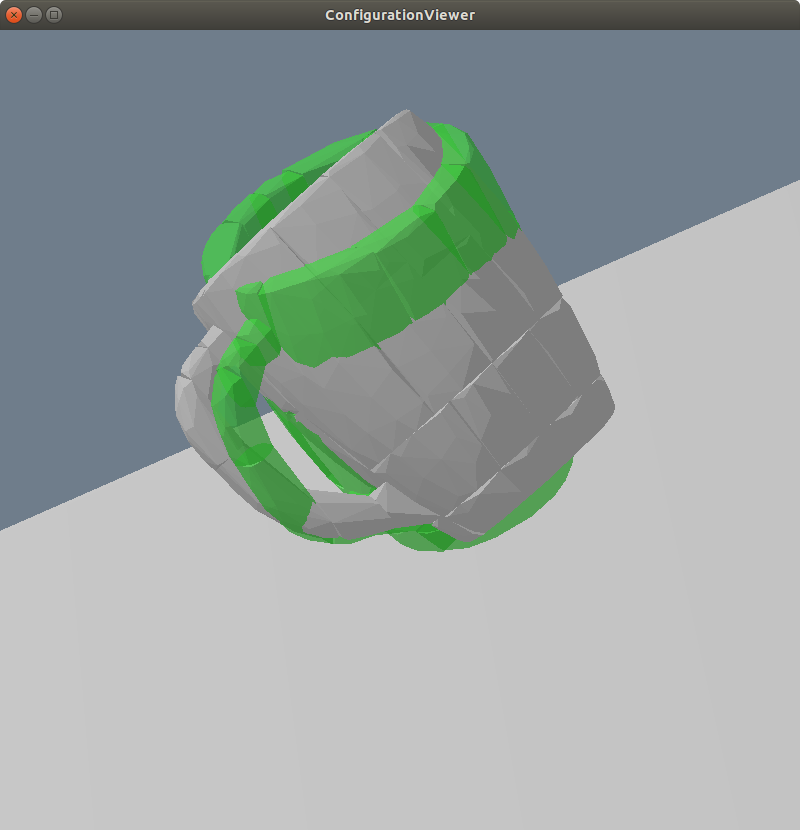}}
	\subfigure[F+ICP2]{
		\includegraphics[width=.23\columnwidth, viewport=0 0 800 780, clip]{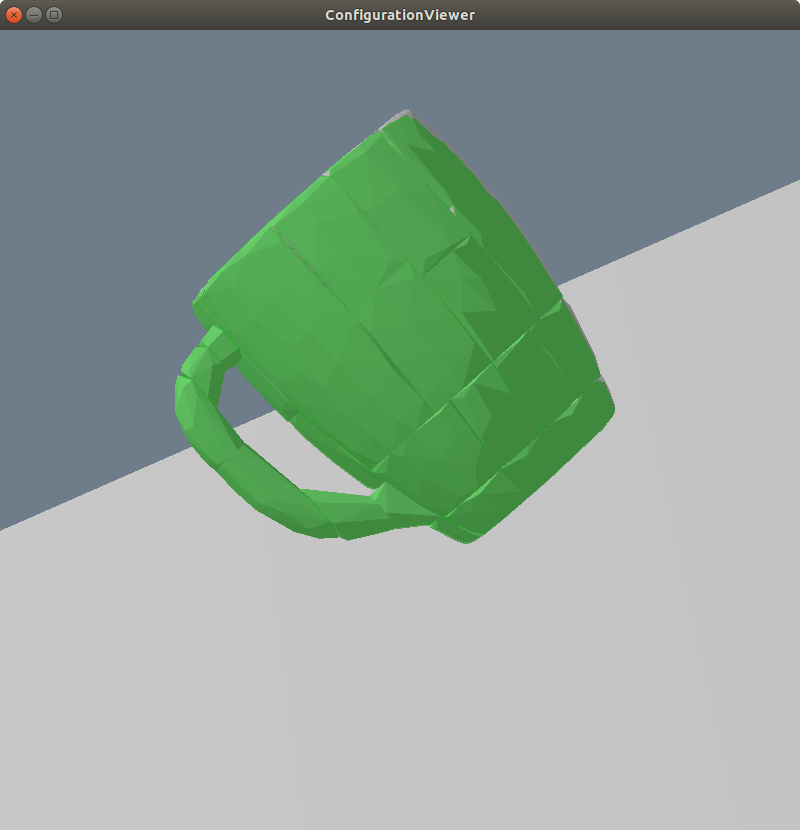}}
	
	\subfigure[ICP]{
		\includegraphics[width=.23\columnwidth, viewport=0 0 800 780, clip]{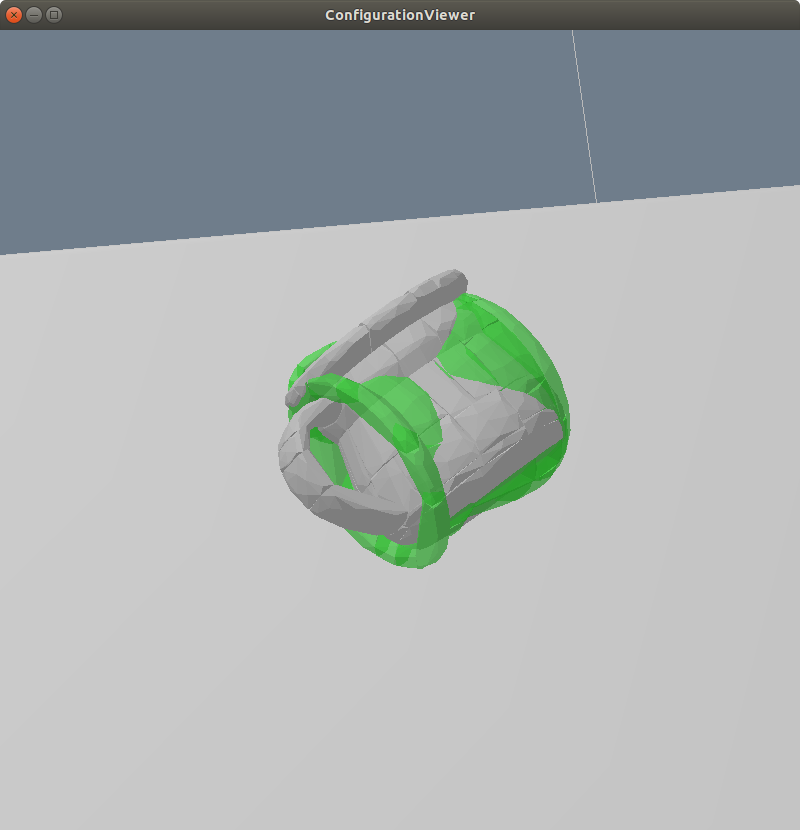}}
	\subfigure[ICP2]{
		\includegraphics[width=.23\columnwidth, viewport=0 0 800 780, clip]{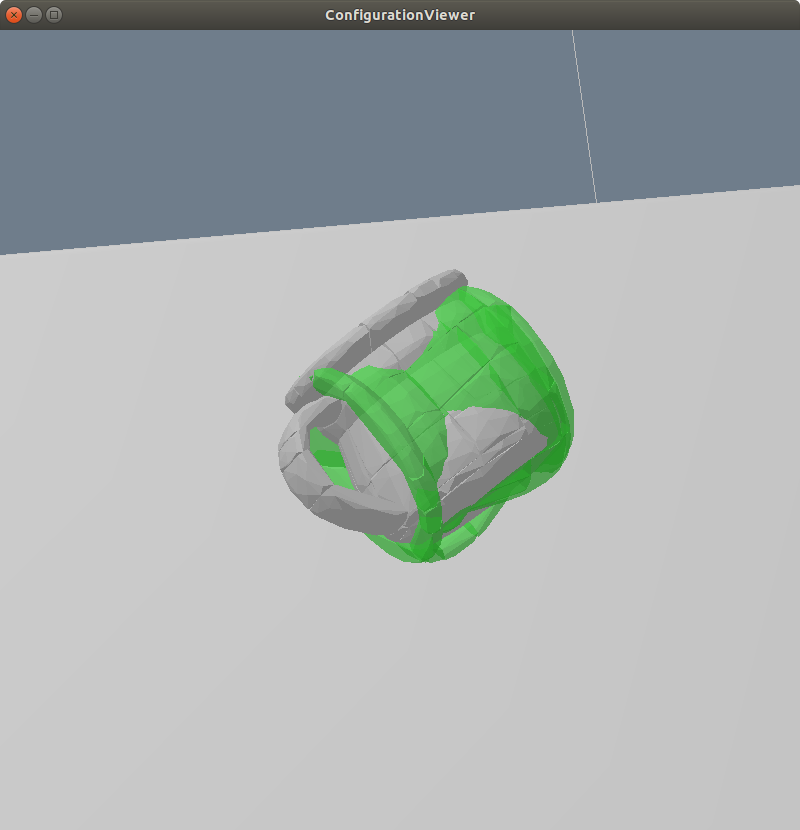}}
	\subfigure[FCP]{
		\includegraphics[width=.23\columnwidth, viewport=0 0 800 780, clip]{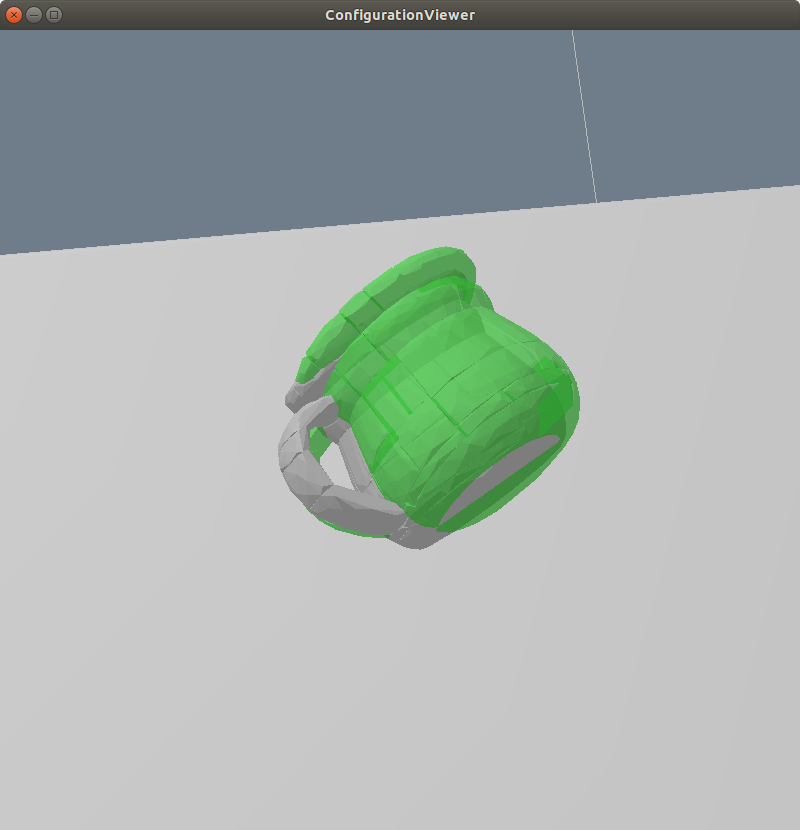}}
	\subfigure[F+ICP2]{
		\includegraphics[width=.23\columnwidth, viewport=0 0 800 780, clip]{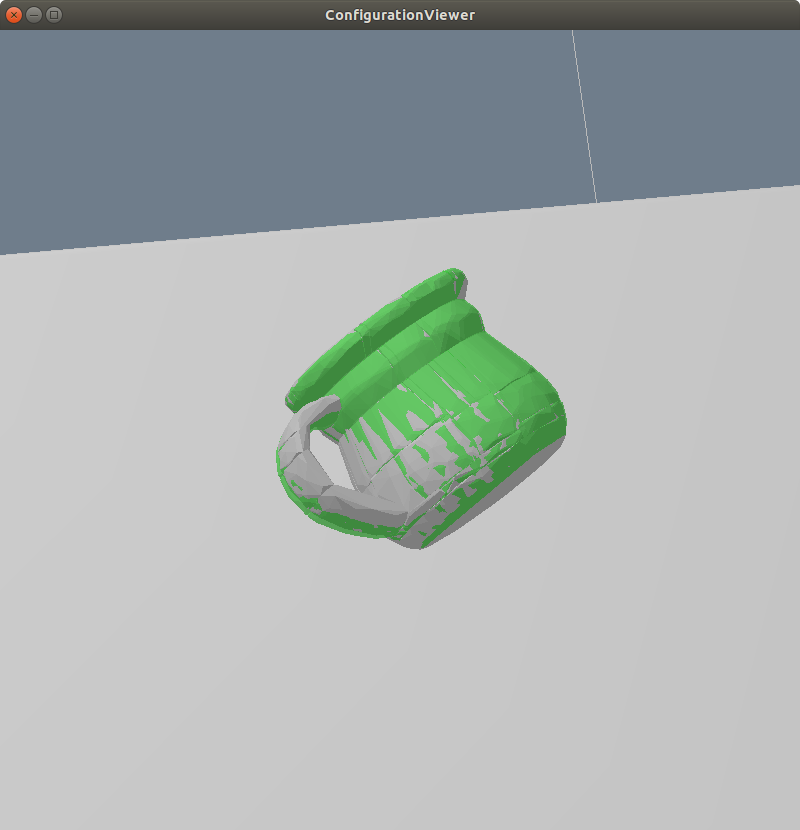}}
	
	\subfigure[ICP]{
		\includegraphics[width=.23\columnwidth, viewport=0 0 800 780, clip]{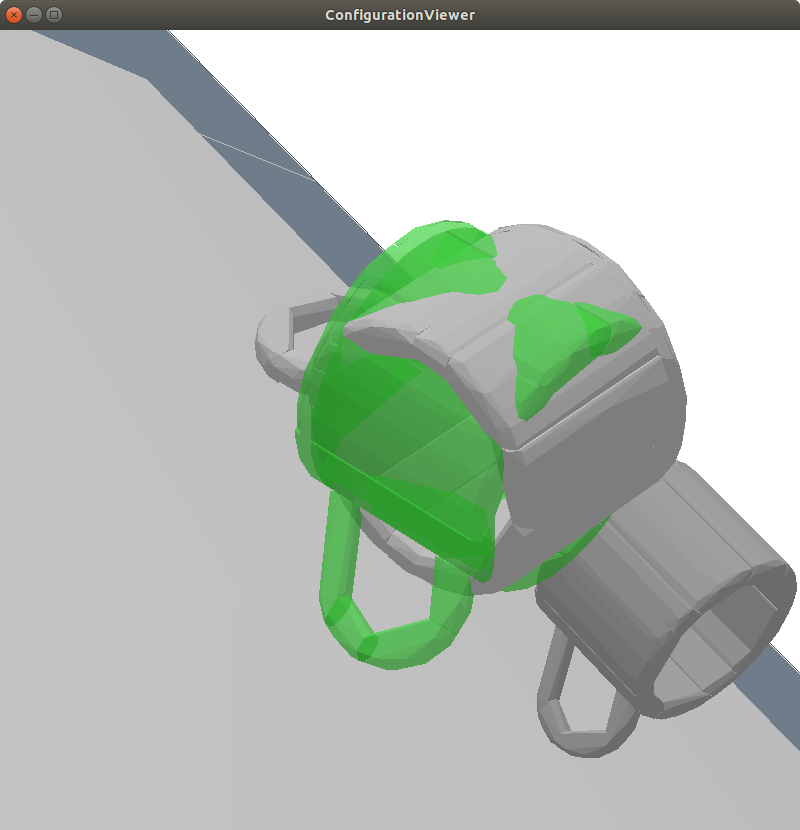}}
	\subfigure[ICP2]{
		\includegraphics[width=.23\columnwidth, viewport=0 0 800 780, clip]{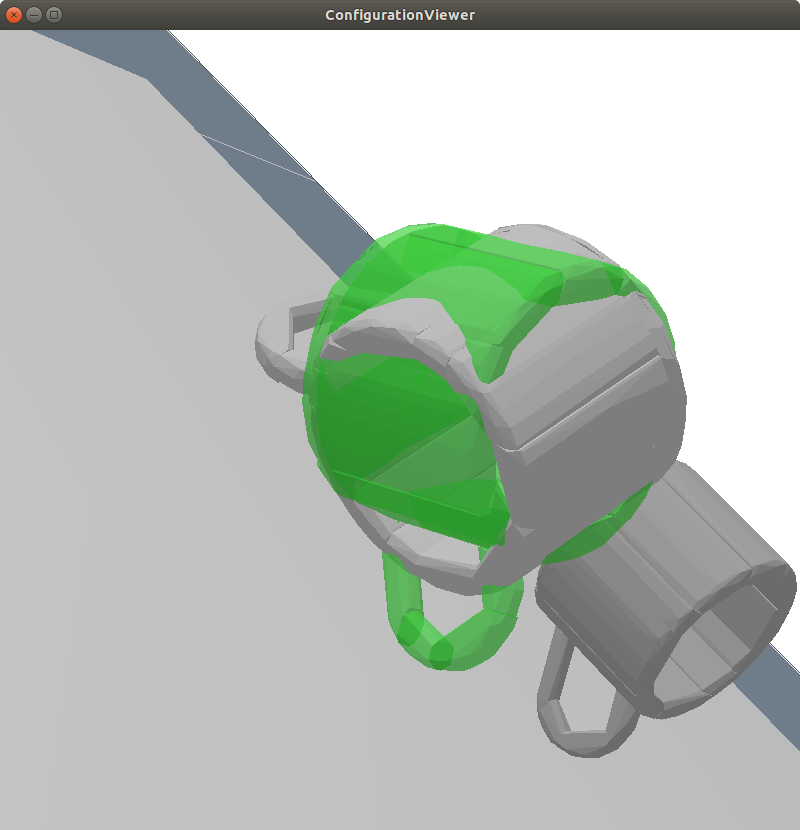}}
	\subfigure[FCP]{
		\includegraphics[width=.23\columnwidth, viewport=0 0 800 780, clip]{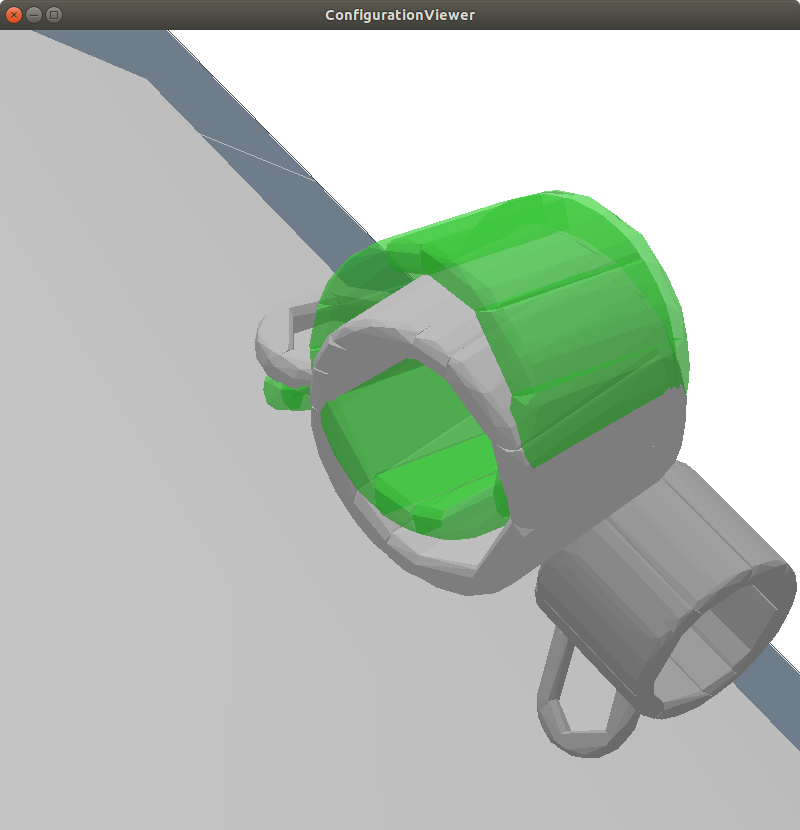}}
	\subfigure[F+ICP2]{
		\includegraphics[width=.23\columnwidth, viewport=0 0 800 780, clip]{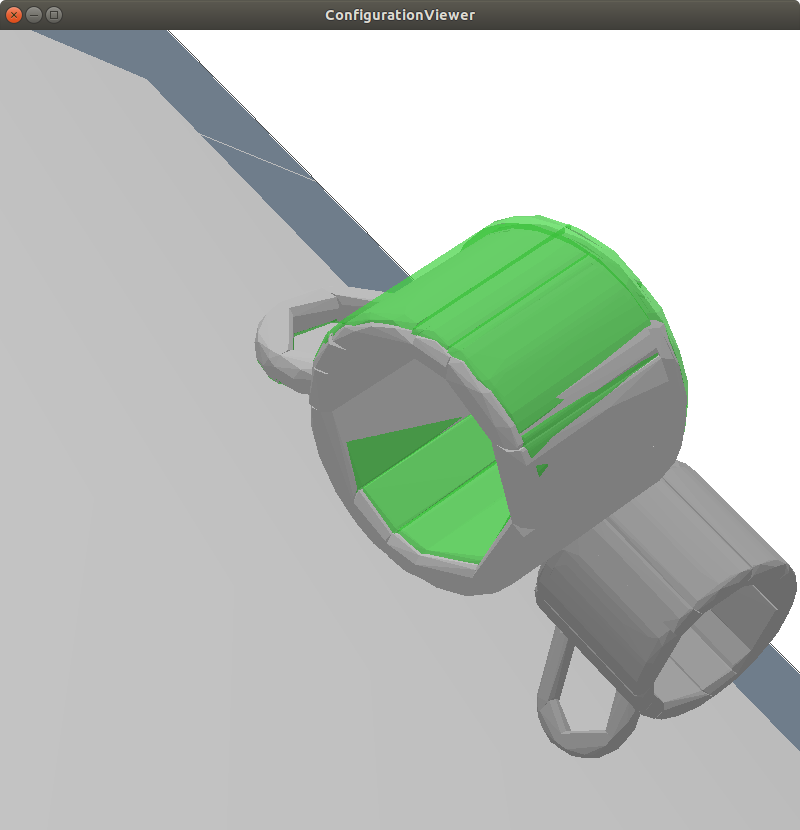}}
	\caption{6D Pose Estimation Results - the estimated poses are applied to the green meshes. ICP easily gets stuck at local optima while FCP produces fairly accurate poses which help F+ICP2 escape the local optima; note that FCP does not iterate to get the results.}\label{fig:poseEstim2}
\end{figure}	

\begin{figure}[t]
	\centering
	\subfigure[t=5]{
		\includegraphics[width=.3\columnwidth]{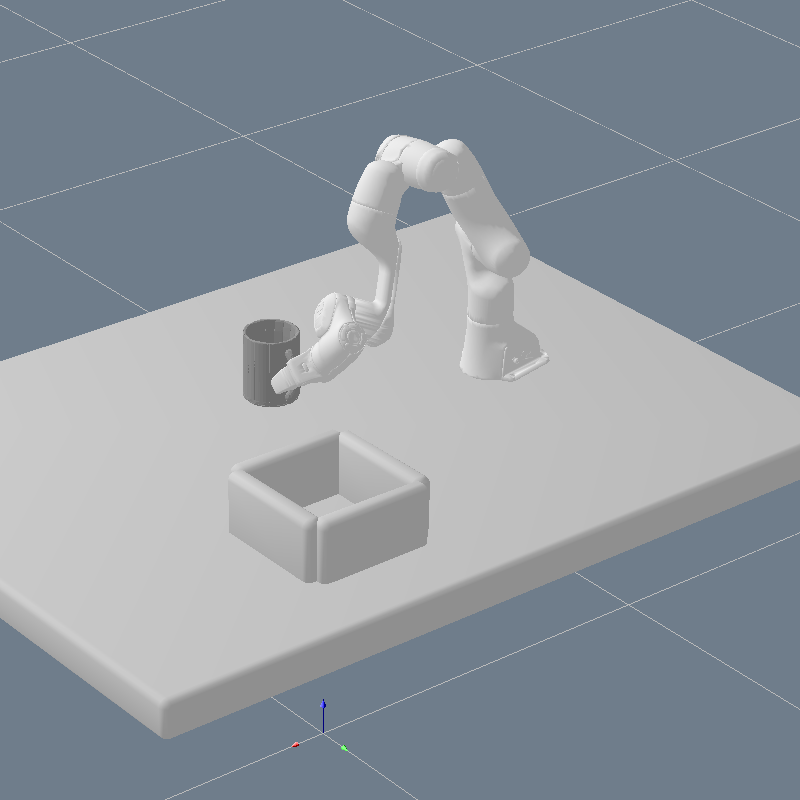}}
	\subfigure[t=10]{
		\includegraphics[width=.3\columnwidth]{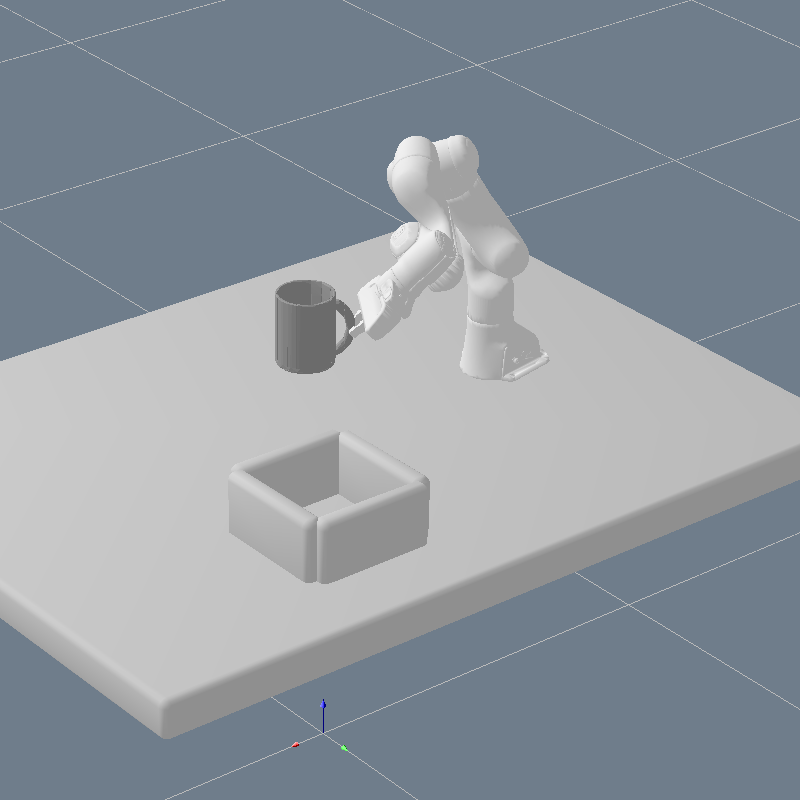}}
	\subfigure[t=15]{
		\includegraphics[width=.3\columnwidth]{im_ref4}}
	\caption{Zero-shot Imitation - reference motion. Two sets of posed images are obtained at $t=10,~15$.} \label{fig:ref}
\end{figure}

\begin{figure}[t]
	\centering
	\subfigure[t=5]{
		\includegraphics[width=.3\columnwidth]{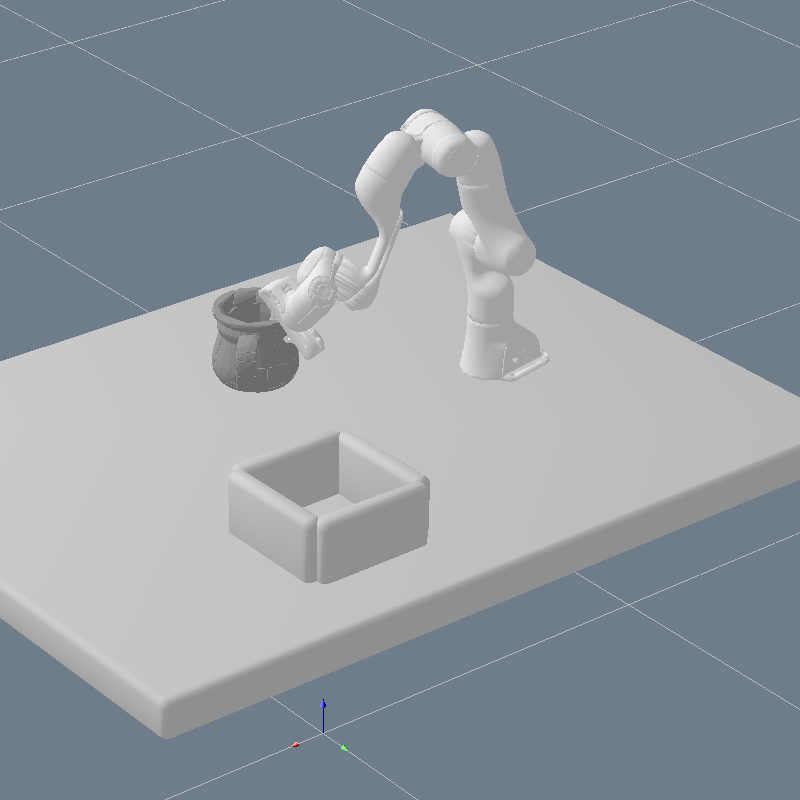}}
	\subfigure[t=10]{
		\includegraphics[width=.3\columnwidth]{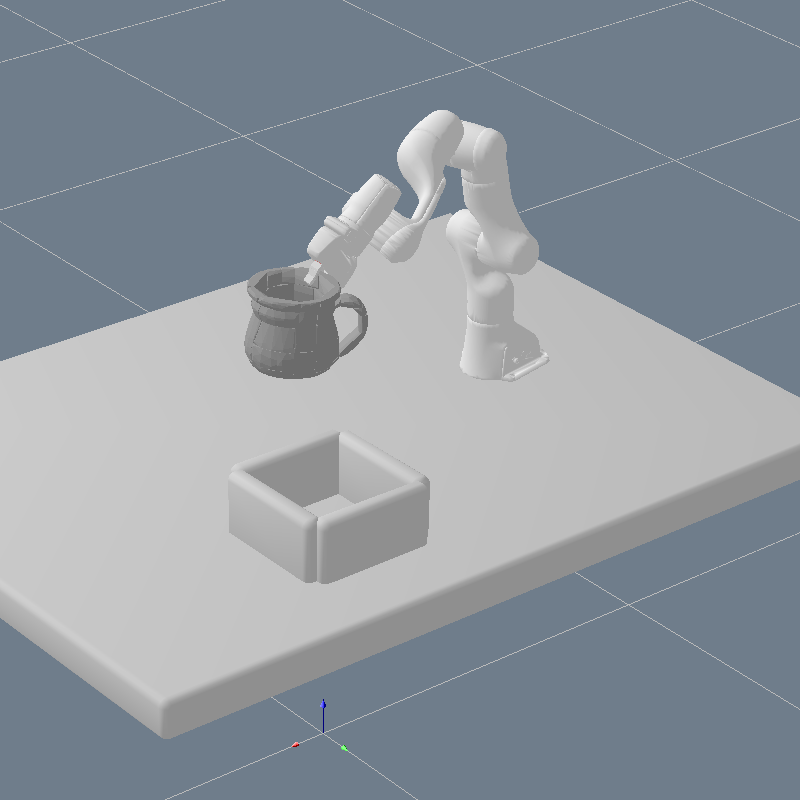}}
	\subfigure[t=15]{
		\includegraphics[width=.3\columnwidth]{im_14}}
	\subfigure[t=5]{
		\includegraphics[width=.3\columnwidth]{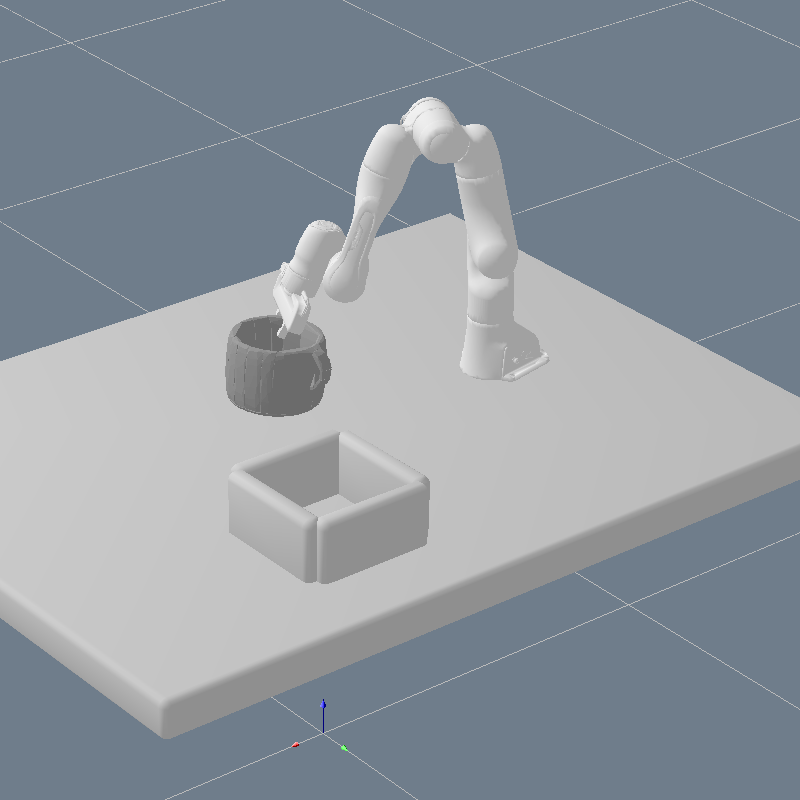}}
	\subfigure[t=10]{
		\includegraphics[width=.3\columnwidth]{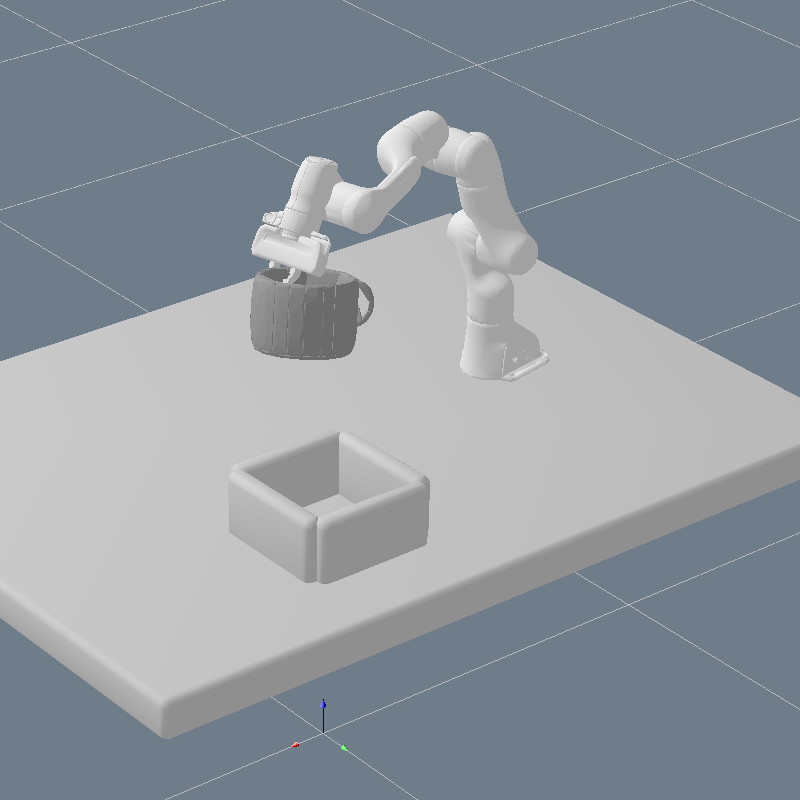}}
	\subfigure[t=15]{
		\includegraphics[width=.3\columnwidth]{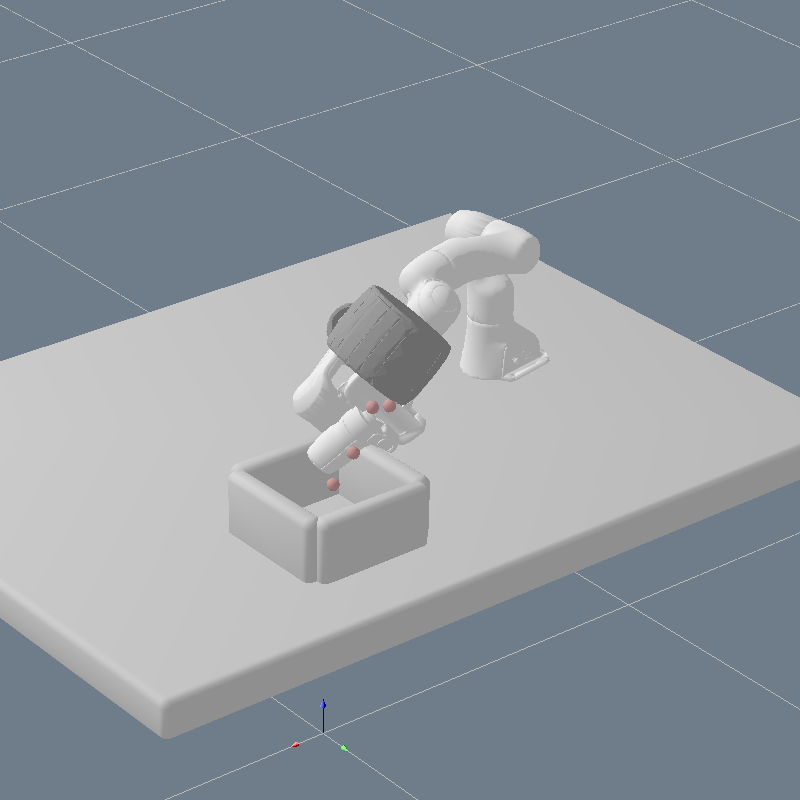}}
	\subfigure[t=5]{
		\includegraphics[width=.3\columnwidth]{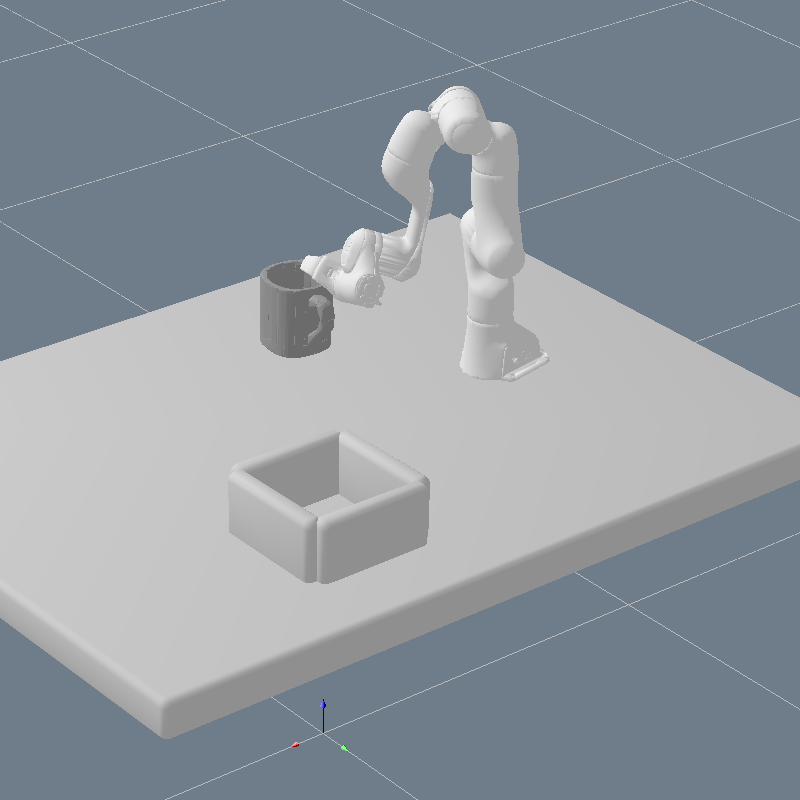}}
	\subfigure[t=10]{
		\includegraphics[width=.3\columnwidth]{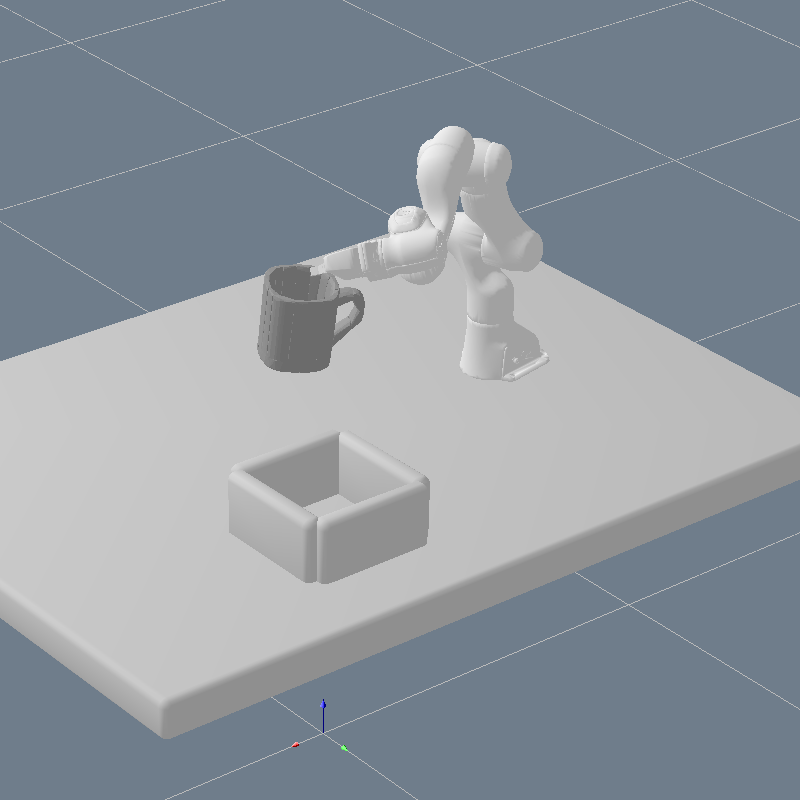}}
	\subfigure[t=15]{
		\includegraphics[width=.3\columnwidth]{im_44}}
	\caption{Zero-shot imitation - optimized motions. The FCP constraints are imposed at $t=10,~15$. The imitations are achieved only from images, without defining the canonical coordinate/pose of the objects.} \label{fig:imitation_full}
\end{figure}

\begin{figure}
	\centering
	\subfigure{\includegraphics[width=.4\columnwidth]{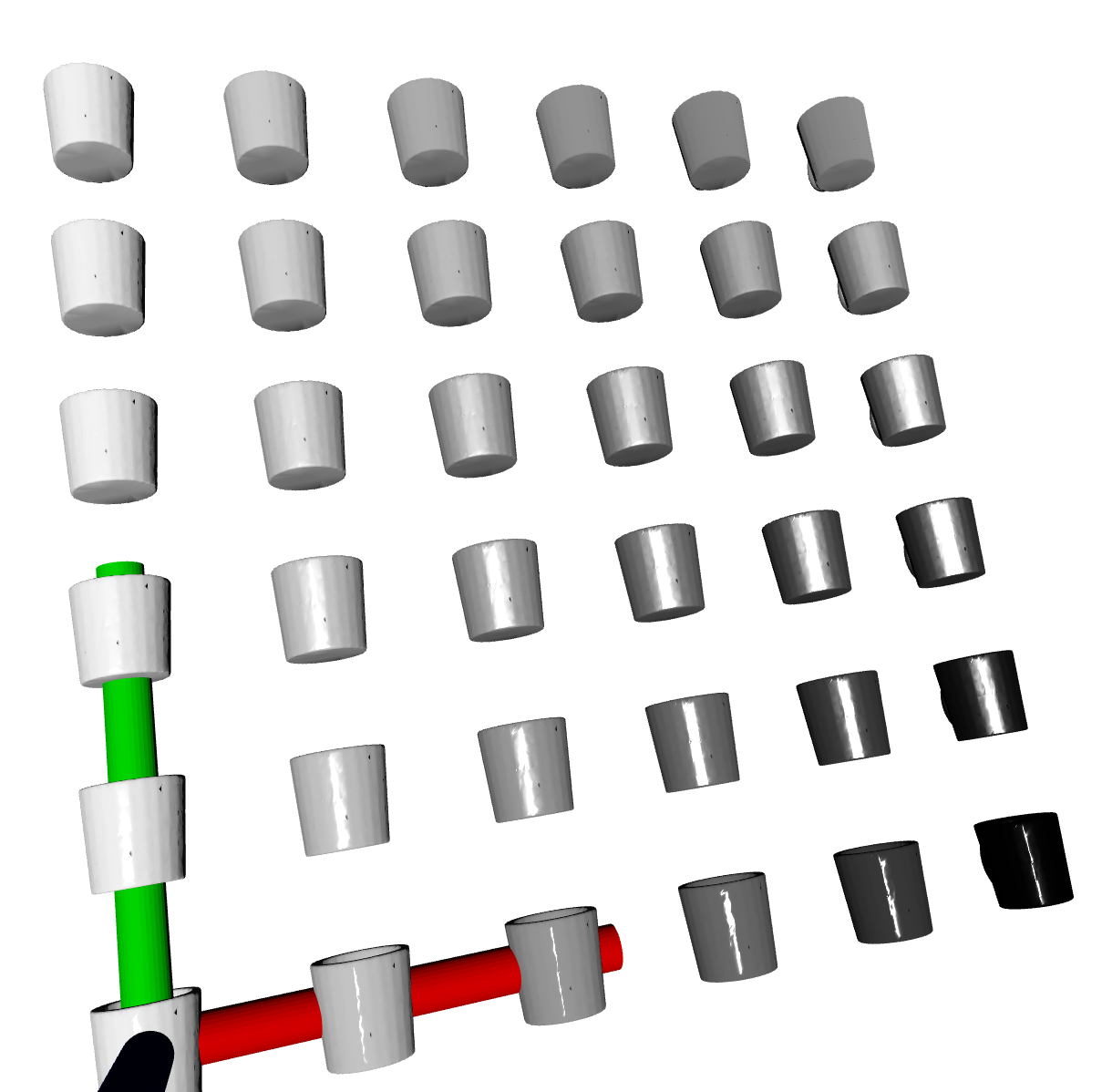}}
	\subfigure{\includegraphics[width=.35\columnwidth]{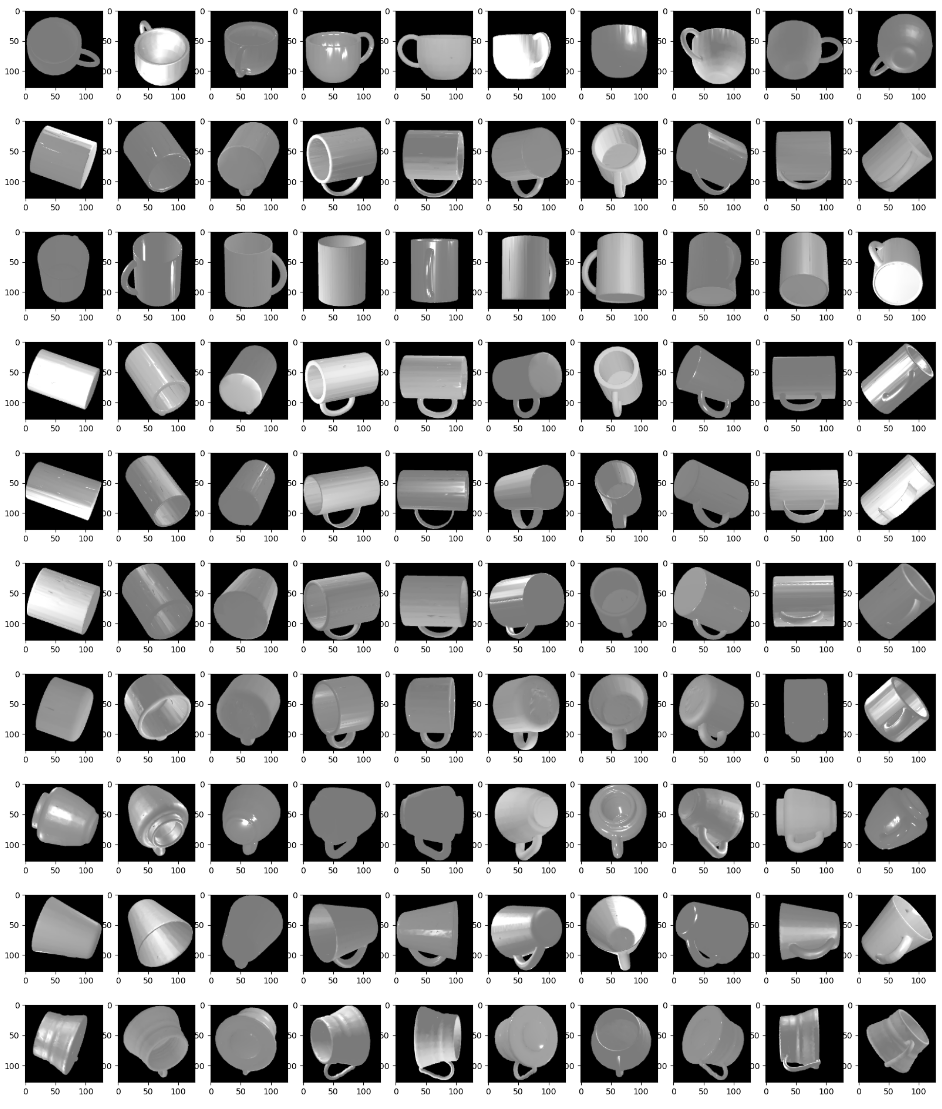}}
	\caption{Real robot transfer: Mug materials used by domain randomization.} \label{fig:real_dr}
\end{figure}

\begin{figure}[t]
	\centering
	\subfigure{
		\includegraphics[width=.475\columnwidth]{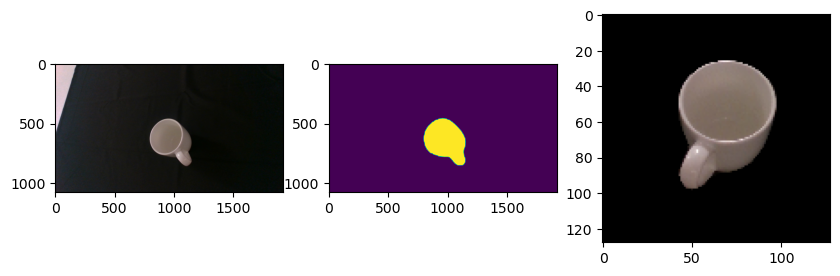}}
	\subfigure{
		\includegraphics[width=.475\columnwidth]{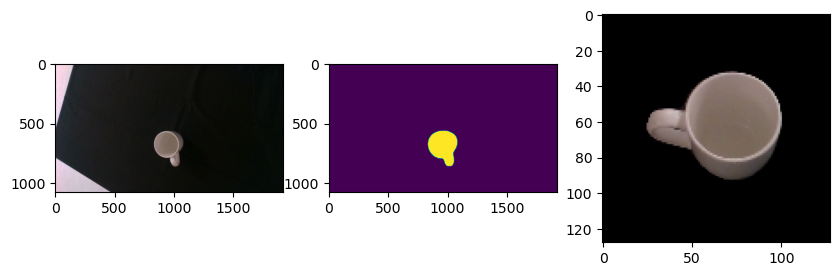}}
	\subfigure{
		\includegraphics[width=.475\columnwidth]{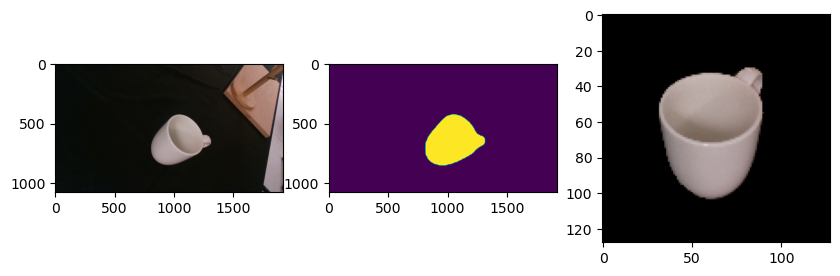}}
	\subfigure{
		\includegraphics[width=.475\columnwidth]{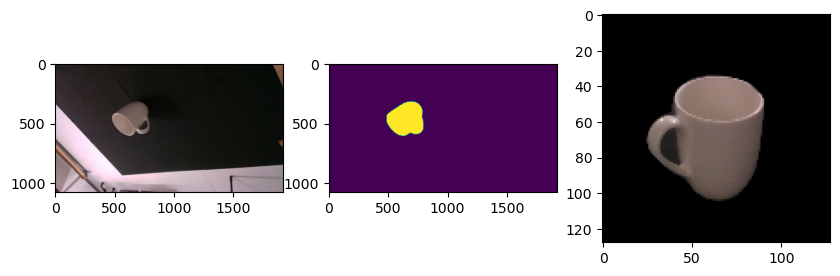}}
	\subfigure{
		\includegraphics[width=.475\columnwidth]{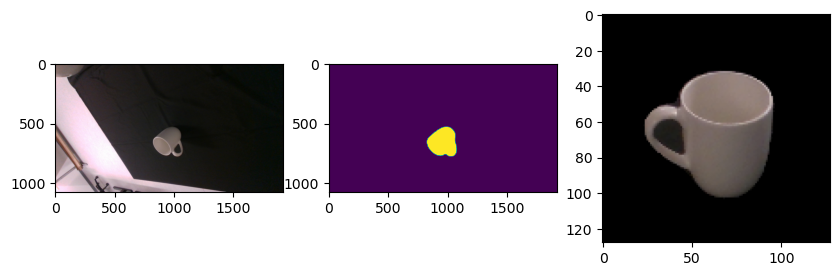}}
	\subfigure{
		\includegraphics[width=.475\columnwidth]{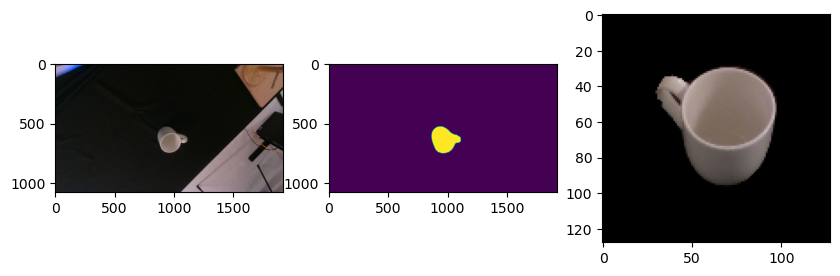}}
	\subfigure{
		\includegraphics[width=.475\columnwidth]{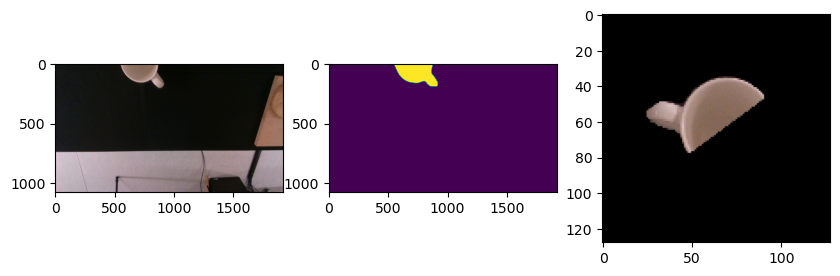}}
	\caption{Real robot transfer: Multi-view processing with real images. The object masks are obtained from Mask R-CNN. 8 images were taken and the mask detection failed in one image.} \label{fig:real_mv}
\end{figure}

\end{document}